\documentclass[10pt,letterpaper]{article}
\usepackage{mvplab}

\usepackage{amsmath}
\usepackage{amssymb}
\usepackage{mathrsfs}
\usepackage[table]{xcolor}
\usepackage{adjustbox} 
\definecolor{rkgold}{HTML}{FFF3BF}
\definecolor{rksilver}{HTML}{E3E8EF}
\definecolor{rkbronze}{HTML}{F7D6C1}
\newcommand{\rkA}[1]{\cellcolor{rkgold}#1}
\newcommand{\rkB}[1]{\cellcolor{rksilver}#1}
\newcommand{\rkC}[1]{\cellcolor{rkbronze}#1}
\DeclareRobustCommand{\rkn}[1]{\,{\tiny\color{black!50}\##1}}

\newcommand{\ours}{SpatialArena\xspace}
\newcommand{\oursdata}{Spatial Arena\xspace} 

\usepackage{listings}
\definecolor{lstkw}{rgb}{0.00,0.33,0.62}
\definecolor{lststr}{rgb}{0.64,0.08,0.08}
\definecolor{lstcom}{rgb}{0.42,0.42,0.42}
\definecolor{lstbg}{rgb}{0.97,0.97,0.97}
\lstdefinestyle{codebase}{
  basicstyle=\ttfamily\scriptsize,
  backgroundcolor=\color{lstbg},
  frame=single, rulecolor=\color{lstbg},
  framesep=2pt, xleftmargin=3pt, xrightmargin=3pt,
  columns=fullflexible, keepspaces=true,
  showstringspaces=false, breaklines=true,
  commentstyle=\color{lstcom}\itshape,
  stringstyle=\color{lststr},
  keywordstyle=\color{lstkw}\bfseries,
  aboveskip=2pt, belowskip=2pt,
}
\newtcolorbox{prompttext}{enhanced,arc=3pt,boxrule=0.6pt,colframe=mvpblue!55,colback=mvpbg,
  left=6pt,right=6pt,top=4pt,bottom=4pt,before skip=4pt,after skip=6pt,
  fontupper=\ttfamily\scriptsize\raggedright,
  before upper={\hyphenpenalty=10000\relax\exhyphenpenalty=10000\relax\frenchspacing}}
\lstdefinestyle{python}{
  style=codebase, language=Python,
  morekeywords={True,False,None},
}
\lstdefinelanguage{yaml}{
  keywords={true,false,null},
  sensitive=false,
  comment=[l]{\#},
  morestring=[b]",
  morestring=[b]',
}
\lstdefinestyle{yaml}{
  style=codebase, language=yaml,
}

\AtEndPreamble{%
  \crefname{algocf}{Alg.}{Algs.}%
  \crefname{appendix}{Sec.}{Secs.}%
  \Crefname{appendix}{Sec.}{Secs.}%
  \crefname{subappendix}{Sec.}{Secs.}%
  \Crefname{subappendix}{Sec.}{Secs.}%
  \crefname{subsubappendix}{Sec.}{Secs.}%
  \Crefname{subsubappendix}{Sec.}{Secs.}%
  \Crefname{algocf}{Algorithm}{Algorithms}%
}

\definecolor{cvprblue}{rgb}{0.21,0.49,0.74}
\usepackage[breaklinks,colorlinks,allcolors=mvpblue]{hyperref}
\usepackage{graphicx}
\usepackage{amsmath}
\usepackage{amssymb}
\usepackage{multirow}
\usepackage{booktabs}
\usepackage{array}
\usepackage{amsfonts}
\usepackage{color}
\usepackage{xcolor}
\usepackage{algorithmicx,algpseudocode}
\usepackage{setspace}
\usepackage{stfloats}

\usepackage[linesnumbered,ruled,vlined]{algorithm2e}

\makeatletter
\newcommand{\suppdef}[3]{%
  \expandafter\def\csname suppnum@#1\endcsname{#2}%
  \expandafter\def\csname supptitle@#1\endcsname{#3}}
\newcommand{\suppletter}[1]{%
  \ifcsname suppnum@#1\endcsname\@Alph{\csname suppnum@#1\endcsname}\else\textbf{??}\fi}
\definecolor{suppcolor}{rgb}{0.85,0.37,0.0}
\newcommand{\suppref}[1]{\hyperref[sec:suppl:#1]{\textcolor{suppcolor}{Sec.~\suppletter{#1}}}}
\newcommand{\Suppref}[1]{\suppref{#1}}
\newcommand{\suppsection}[1]{%
  \FloatBarrier 
  \ifdim\pagetotal=\z@\ifx\@toplist\@empty\global\@colroom\@colht\global\vsize\@colht\fi\fi
  \setcounter{section}{\numexpr\csname suppnum@#1\endcsname-1\relax}%
  \section{\csname supptitle@#1\endcsname}\label{sec:suppl:#1}}
\makeatother

\suppdef{data}{1}{Datasets and Protocol}
\suppdef{results}{2}{Additional Results}
\suppdef{metrics}{3}{Metric Definitions}
\suppdef{impl}{4}{Implementation Details}
\suppdef{design}{5}{Toolkit Design}
\suppdef{depth}{4}{}
\suppdef{traj}{4}{}
\suppdef{models}{4}{}

\usepackage{placeins} 
\usepackage{fontawesome5}
\definecolor{findinggreen}{HTML}{1A7F37}
\newcommand{\findingbox}[2][Key finding]{\par\vspace{3pt}\noindent
  \hbox{\textcolor{findinggreen}{\vrule width 2.5pt}\hspace{7pt}%
  \begin{minipage}{\dimexpr\linewidth-9.5pt\relax}
    \vspace{2pt}{\color{findinggreen}\faLightbulb[regular]\hspace{4pt}\textbf{#1}}\par\vspace{2pt}#2\vspace{2pt}
  \end{minipage}}\par\vspace{5pt}}
\definecolor{noteblue}{HTML}{0969DA}

\colorlet{suppcolor}{mvppink}

\newcommand{\suppfigpage}[1]{\begin{figure*}[!t]#1\end{figure*}}

\title{SpatialArena: Where Do Geometric Foundation Models Stand?}
\shorttitle{SpatialArena: Where Do Geometric Foundation Models Stand?}
\author{Yu~Chen\projectlead \and Haosen~Yang \and Su~Ee~Tan \and Zhensong~Zhang \and Jifei~Song \and Jiankang~Deng\corresponding}
\affiliation{MVP Lab \& Imperial College London}
\mvpdate{November 23, 2026}
\projecturl{https://mvp-ai-lab.github.io/SpatialArena}
\githuburl{https://github.com/mvp-ai-lab/SpatialArena}
\huggingfaceurl{https://huggingface.co/datasets/mvp-lab/SpatialArena-Dataset}
\projectleader{Yu Chen}{hackerdreamer34@gmail.com}
\correspondence{Jiankang Deng}{j.deng16@imperial.ac.uk}

\begin{document}
\begin{abstract}
3D perception is shifting from task-specific pipelines to geometric foundation models: monocular depth networks, feed-forward 3D/4D reconstruction models, and vision--language models that predict geometry. However, their progress cannot be measured reliably. Published numbers are not comparable, because every paper samples frames, sets resolutions, and selects test splits differently. They are not robust because the alignment required for every up-to-scale prediction can outweigh the gap between models: fitting depth in the wrong space nearly triples its error, and standard trajectory alignment understates the error of failing methods by up to $2.6\times$. And they rarely cover long recordings, dynamic scenes, or aerial views. We present \ours, an open arena that scores geometric models with the same metrics and rules, with robust and deterministic alignment. \ours spans five tasks, from monocular depth to SLAM. It covers 23 public datasets with 418,929 images across indoor, outdoor, aerial, object-centric, and dynamic scenes, together with \oursdata, our own captures with accurate poses and depth. Evaluating 52 models, we find that single-image depth is close to saturation on public benchmarks but not on large, uncontrolled scenes, and that progress now comes from consistency across frames and views, where feed-forward 3D/4D models match or surpass monocular specialists and overtake video depth models. The benchmark, toolbox, and leaderboard will be publicly released.

\end{abstract}
\maketitle

\begin{figure}[t]
\centering
\includegraphics[width=\textwidth]{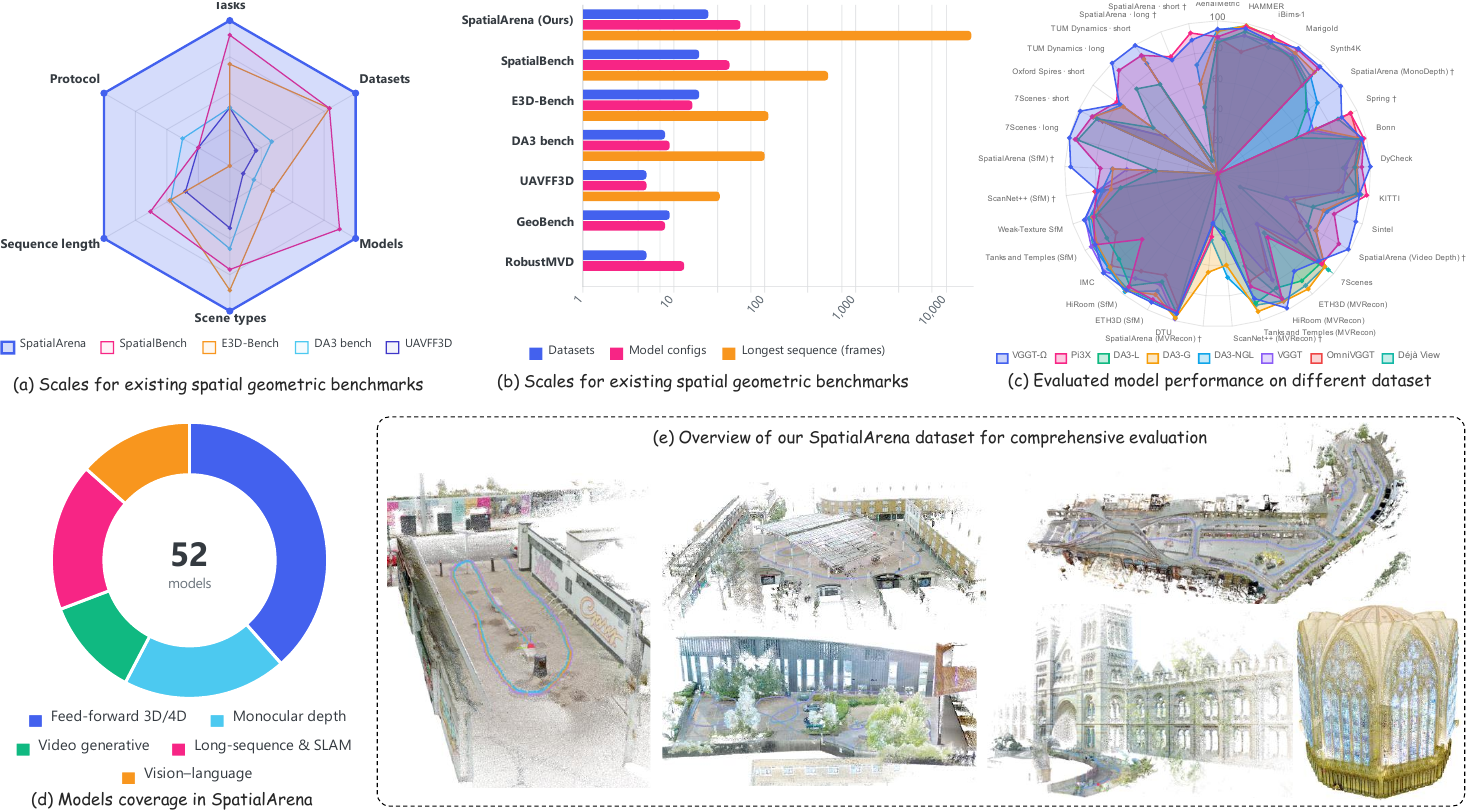}
\caption{\textbf{Overview of the \ours benchmark.} (a) \ours against existing geometry benchmarks along six axes, each normalized by its largest value (sequence length in log frames); E3D-Bench dataset counts are a lower bound. (b) Datasets, models, and longest sequence of each benchmark. (c) Score of eight leading models on each dataset; $\dagger$ marks datasets excluded from the ranking. No model leads on every dataset. (d) Evaluated models by family. (e) Scenes of our \oursdata dataset, which feeds all five tasks.}
\label{fig:teaser}
\end{figure}

\section{Introduction}
\label{sec:intro}

Perceiving the 3D world from images is one of the oldest goals of computer vision, and it is undergoing the shift that natural language processing went through a few years ago. Depth, camera pose, and scene geometry, traditionally estimated by separate hand-engineered pipelines~\cite{DBLP:conf/cvpr/SchonbergerF16,DBLP:conf/eccv/SchonbergerZFP16,DBLP:journals/trob/Mur-ArtalMT15}, are now predicted by large networks trained on massive data: monocular depth models generalize to unseen scenes without further finetuning~\cite{DBLP:conf/nips/YangKH0XFZ24,DBLP:conf/cvpr/KeOHMDS24,DBLP:conf/nips/WangXDDXLSTY25,DBLP:conf/iclr/BochkovskiyDGSZ25}, feed-forward 3D/4D models predict cameras, depth, and point maps for thousands of frames in one pass~\cite{DBLP:conf/cvpr/Wang0CCR24,DBLP:conf/cvpr/WangCKV0N25,DBLP:conf/iclr/WangZZCZLCPSH26,DBLP:conf/iclr/LinCLCLZPGZSFK26,DBLP:conf/cvpr/WangZHEK25,chen2026lingbotmap}, and vision--language models begin to predict metric geometry~\cite{DBLP:conf/cvpr/0003XKISGX24,DBLP:conf/iclr/CaiYXLMLZLCS26,DBLP:journals/corr/abs-2605-15876}. These \emph{geometric foundation models} raise questions beyond any single task. Will general models absorb the specialized ones, as large language models (LLMs) did in NLP? Where do improvements actually come from? And is the field converging toward models that understand 3D geometry, or toward models that fit a few test sets?

Answering them requires measurement. Shared benchmarks have shaped both vision and language: ImageNet~\cite{DBLP:conf/cvpr/DengDSLL009} made progress in recognition visible and comparable, and holistic evaluations and open arenas~\cite{DBLP:journals/tmlr/LiangBLTSYZNWKN23,DBLP:conf/icml/ChiangZ0ALLZ0JG24} track how LLMs evolve across many abilities at once. Geometric foundation models have no such common ground, and three problems make their results hard to compare.
\emph{First, every work evaluates with its own tools.} Frame sampling, input resolution, test splits, and evaluation code differ from paper to paper~\cite{DBLP:conf/cvpr/KeOHMDS24,DBLP:conf/cvpr/WangZHEK25,DBLP:conf/iclr/LinCLCLZPGZSFK26}, so a reported gap between two models may come from evaluation rather than from the models.
\emph{Second, metric computation is neither consistent nor robust.} Most models predict geometry up to scale, so every score passes through an alignment to the ground truth, and that alignment can matter more than the model. Fitting a disparity model in depth space nearly triples its AbsRel, a least-squares fit raises the AbsRel of strong multi-view models by more than 50\% (\suppref{depth}), and the closed-form fit on camera centers used for trajectories~\cite{DBLP:journals/pami/Umeyama91,grupp2017evo} understates the error of failing methods by up to $2.6\times$ (\suppref{traj}).
\emph{Third, coverage is partial.} Most papers report results on a subset of public datasets, rarely on the regimes that matter in practice: dynamic scenes, aerial views, weakly textured objects, or recordings of thousands of frames. Some test scenes are even part of the evaluated models' training data (\suppref{data}).
As a result, it is unclear how far geometric foundation models have actually come, and in which direction.

We argue that the field needs an \emph{arena}: an open platform where geometric models are evaluated with the same metrics, under the same rules. \ours is built on three principles.
\emph{Fairness.} One evaluation suite scores every model from its outputs using the same inputs, alignment, and metrics, and excludes the datasets models were trained on from the ranking.
\emph{Robust measurement.} Depth is aligned by a robust fit in each model's native output space on pixels chosen by the ground truth alone, and trajectories by a pose-aware, outlier-robust similarity estimator.
\emph{Breadth.} \ours spans five tasks, from monocular and video depth to multi-view reconstruction, structure from motion, and SLAM on recordings of nearly 19k frames, on 23 public datasets of indoor, outdoor, aerial, object-centric, static, and dynamic scenes and on \oursdata, 30 new large-scale scenes captured with a handheld LiDAR and three-camera rig, with accurate camera poses and depth.

We evaluate 52 depth, feed-forward 3D/4D, long-sequence SLAM, and open and commercial vision--language models, and three findings stand out (\cref{sec:exp}). First, single-image depth is close to saturation on the public benchmarks: the best monocular AbsRel there improved only from 5.5\% to 4.9\% in two years, whereas on our large, uncontrolled captures the errors are three to four times higher; the remaining headroom lies in fine accuracy and in such scenes. Second, general models are absorbing specialized ones: feed-forward 3D/4D models match the monocular specialists on public data, surpass them once our captures are scored, and, on video depth, halve the error of the best dedicated model. Third, vision--language models improve fastest on single-image depth but remain far behind, lacking metric precision and consistency across frames. No model leads on every dataset, so the field has not yet converged on one model of 3D structure. In summary, our contributions are:
\begin{itemize}[leftmargin=*,itemsep=0pt,topsep=2pt]
  \item \ours, an open arena that evaluates geometric foundation models on five tasks, 23 public datasets, and our captures under one protocol;
  \item robust, deterministic alignment for depth and trajectories, with a comparison of six depth alignments and a controlled study showing that the standard trajectory alignment hides failures;
  \item an evaluation of 52 models across specialist, feed-forward, and vision--language families that traces how geometric foundation models are evolving.
\end{itemize}

\begin{figure*}[h]
\centering
\includegraphics[width=\textwidth]{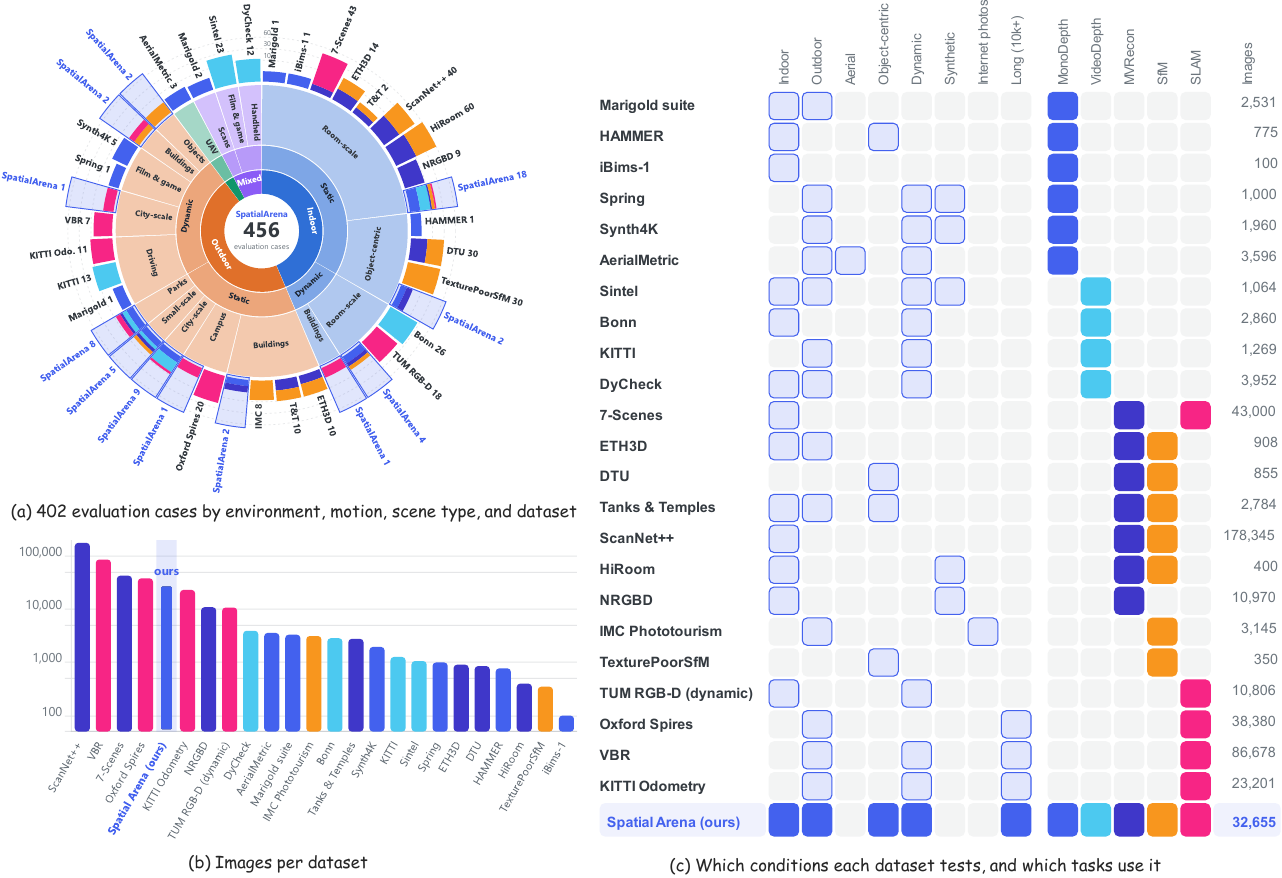}
\caption{\textbf{Datasets of \ours.} (a) Evaluation cases by environment, static or dynamic scenes, and scene type (inner rings), and per dataset, stacked by task
. (b) Unique images per dataset; the 23 public datasets hold 418,929 images. (c) Conditions each dataset tests and the tasks that use it. All five tasks test indoor and outdoor scenes, whereas aerial views, internet photos, and recordings of more than 10k frames are tested by one task each. \oursdata adds large-scale scenes with accurate poses and depth to all five tasks.}
\label{fig:intro_spatial_arena}
\end{figure*}

\section{Related Work}
\label{sec:related}

\paragraph{Depth foundation models.}
Affine-invariant training on mixed data~\cite{DBLP:journals/pami/RanftlLHSK22,DBLP:conf/iccv/RanftlBK21} made monocular depth zero-shot. Later work adds unlabeled and synthetic data~\cite{DBLP:conf/cvpr/YangKHXFZ24,DBLP:conf/nips/YangKH0XFZ24}, diffusion priors~\cite{DBLP:conf/cvpr/KeOHMDS24,DBLP:conf/iclr/HeL0LLZZLC25}, metric depth, cameras, or point maps~\cite{DBLP:journals/pami/HuYZCLCWYSS24,DBLP:conf/cvpr/PiccinelliYSSLG24,DBLP:conf/iclr/BochkovskiyDGSZ25,DBLP:conf/cvpr/WangXDX00Y25,DBLP:conf/nips/WangXDDXLSTY25}, and arbitrary output resolution~\cite{DBLP:conf/cvpr/YuLWLWZWZHP26}, mostly on self-supervised backbones~\cite{DBLP:journals/tmlr/OquabDMVSKFHMEA24}; video models add temporal consistency~\cite{DBLP:conf/cvpr/0002G0ZCZQS25,DBLP:journals/corr/abs-2603-12250,DBLP:conf/cvpr/ChenGZZHFK25}. Each reports results under its own alignment and test split~\cite{DBLP:journals/corr/abs-2406-12671}.

\paragraph{Feed-forward multi-view geometry.}
DUSt3R~\cite{DBLP:conf/cvpr/Wang0CCR24} and MASt3R~\cite{DBLP:conf/eccv/LeroyCR24} regress point maps from uncalibrated image pairs, and successors scale to many views with spatial memory~\cite{DBLP:conf/3dim/WangA25,DBLP:conf/cvpr/WangZHEK25,DBLP:conf/cvpr/CabonSACCR025} or a single forward pass~\cite{DBLP:conf/cvpr/YangSLHTCCMF25}. VGGT~\cite{DBLP:conf/cvpr/WangCKV0N25} predicts cameras, depth, and point maps from hundreds of frames, and was refined for permutation equivariance~\cite{DBLP:conf/iclr/WangZZCZLCPSH26}, depth--ray outputs~\cite{DBLP:conf/iclr/LinCLCLZPGZSFK26}, accuracy and metric scale~\cite{DBLP:conf/cvpr/WangCZKSLBNVR26,DBLP:conf/cvpr/WangA26,DBLP:journals/corr/abs-2605-30215}, efficiency~\cite{DBLP:journals/corr/abs-2509-02560,DBLP:conf/cvpr/ChenQLPS26}, robustness~\cite{DBLP:conf/cvpr/HanHJJAWKF26}, and semantics~\cite{DBLP:conf/iclr/LiZLZHCLZYZL26,DBLP:journals/corr/abs-2607-19228}; others accept cameras or depth as priors~\cite{DBLP:conf/cvpr/JangW0AR25,DBLP:conf/3dim/KeethaMSPZFKZWALLBRRSK26,DBLP:conf/cvpr/PengLDLLQZZZXL26}. They now drive SfM~\cite{DBLP:conf/3dim/DuisterhofZWLCR25} and SLAM~\cite{DBLP:conf/cvpr/MuraiDD25,DBLP:conf/nips/MaggioLC25}, competing with classical pipelines~\cite{DBLP:conf/cvpr/SchonbergerF16,DBLP:conf/eccv/SchonbergerZFP16,DBLP:conf/eccv/PanBPS24,DBLP:journals/trob/Mur-ArtalMT15,DBLP:conf/nips/TeedD21}, and reach thousands of frames through chunk alignment~\cite{DBLP:conf/icra/DengTXYX26,DBLP:conf/cvpr/XieYJCYRZHPGZ26}, test-time training~\cite{DBLP:conf/cvpr/JinWZGBSH26}, compact memories~\cite{DBLP:journals/corr/abs-2603-03269,chen2026lingbotmap,DBLP:journals/corr/abs-2608-27529}, or local pose regression~\cite{DBLP:journals/corr/abs-2605-26519}, each evaluated on its own sequences and trajectory alignment.

\paragraph{Geometry from general-purpose models.}
Vision--language models answer spatial questions~\cite{DBLP:conf/cvpr/0003XKISGX24,DBLP:conf/cvpr/YangYGH0X25} or predict metric depth~\cite{DBLP:conf/iclr/CaiYXLMLZLCS26,DBLP:journals/corr/abs-2605-15876,DBLP:journals/corr/abs-2505-09388}, and video generators are repurposed as geometry estimators~\cite{DBLP:conf/eccv/WangZKUWZCHABZS26}. Probing studies ask how much 3D structure such models encode~\cite{DBLP:conf/cvpr/BananiRMKLRSG0J24,DBLP:conf/eccv/BratulicMBR26}. Commercial VLMs, such as Claude~\cite{anthropic2026opus55}, GPT~\cite{openai2026gpt6astra}, and Gemini~\cite{google2026gemini4argon}, are available only through APIs and do not disclose their training data. Neither open nor commercial models are routinely evaluated alongside dedicated 3D models.

\paragraph{Benchmarks.}
Most papers reuse the protocols of Marigold~\cite{DBLP:conf/cvpr/KeOHMDS24}, $\pi^3$~\cite{DBLP:conf/iclr/WangZZCZLCPSH26}, and DA3-BENCH~\cite{DBLP:conf/iclr/LinCLCLZPGZSFK26}, with differing sampling, resolution, and alignment, and trajectories rely on an Umeyama fit on camera centers~\cite{DBLP:journals/pami/Umeyama91,grupp2017evo}. RobustMVD~\cite{DBLP:conf/3dim/SchroppelBAB22} evaluates multi-view depth
, GeoBench~\cite{DBLP:journals/corr/abs-2406-12671} retrains monocular models under controlled settings, and E3D-Bench~\cite{DBLP:journals/corr/abs-2506-01933} evaluates 16 geometric foundation models. Concurrent SpatialBench~\cite{DBLP:journals/corr/abs-2605-27367} evaluates 41 model variants on 19 datasets, and UAVFF3D~\cite{DBLP:journals/corr/abs-2605-17942} targets UAV imagery. 

\section{Method}
\label{sec:method}

\ours evaluates spatial foundation models on five geometric tasks. 
We first define the tasks and input tracks (\cref{sec:method:tasks}), then introduce the \oursdata dataset (\cref{sec:method:data}) and the depth and trajectory alignment shared by all tasks (\cref{sec:align}). We then summarize the datasets and metrics of each task (\cref{sec:evaltasks}) and the evaluated models (\cref{sec:method:models}). The data leakage audit is described in \suppref{data}.

\subsection{Problem Setting}
\label{sec:method:tasks}

A scene is a set of $N$ images $\mathcal{I}=\{I_i\}_{i=1}^{N}$. Depending on the task, a model $f$ predicts a per-image depth map $\hat{D}_i$, camera intrinsics $\hat{K}_i$, and world-to-camera extrinsics $\hat{T}_i=[\hat{R}_i\,|\,\hat{\mathbf{t}}_i]\in\mathrm{SE}(3)$. These predictions are compared against the ground truth $D_i$, $K_i$, $T_i$, and, where available, a reference surface $\mathcal{S}$. \ours covers five tasks: monocular depth estimation (\emph{MonoDepth}), video depth estimation (\emph{VideoDepth}), multi-view reconstruction (\emph{MVRecon}), structure from motion (\emph{SfM}), and simultaneous localization and mapping (\emph{SLAM}). They cover 23 public datasets and our own \oursdata captures, with 34 task--dataset pairs and 456 evaluation cases; \suppref{data} lists every dataset.

\paragraph{Input tracks.}
Recent models optionally use geometric priors in addition to images~\cite{DBLP:conf/3dim/KeethaMSPZFKZWALLBRRSK26,DBLP:conf/cvpr/PengLDLLQZZZXL26,DBLP:conf/iclr/WangZZCZLCPSH26,DBLP:conf/iclr/LinCLCLZPGZSFK26}. We make such priors an explicit axis of the protocol, the \emph{track}. In \texttt{rgb}, the model sees only images. In \texttt{rgbc}, it also receives the ground-truth COLMAP cameras. In \texttt{rgbd}, it also receives the ground-truth depth maps. A (task, track) pair is evaluated only if the prior does not trivially reveal the prediction target. This gives \texttt{rgb} for MonoDepth; \texttt{rgb} and \texttt{rgbc} for VideoDepth; all three tracks for MVRecon; and \texttt{rgb} and \texttt{rgbd} for SfM and SLAM. 
All methods receive identical input frames and are scored at ground-truth resolution; frame selection and input resolution are detailed in \suppref{data}.

\subsection{\oursdata Dataset}
\label{sec:method:data}

\begin{figure*}[t]
\centering
\includegraphics[width=\textwidth]{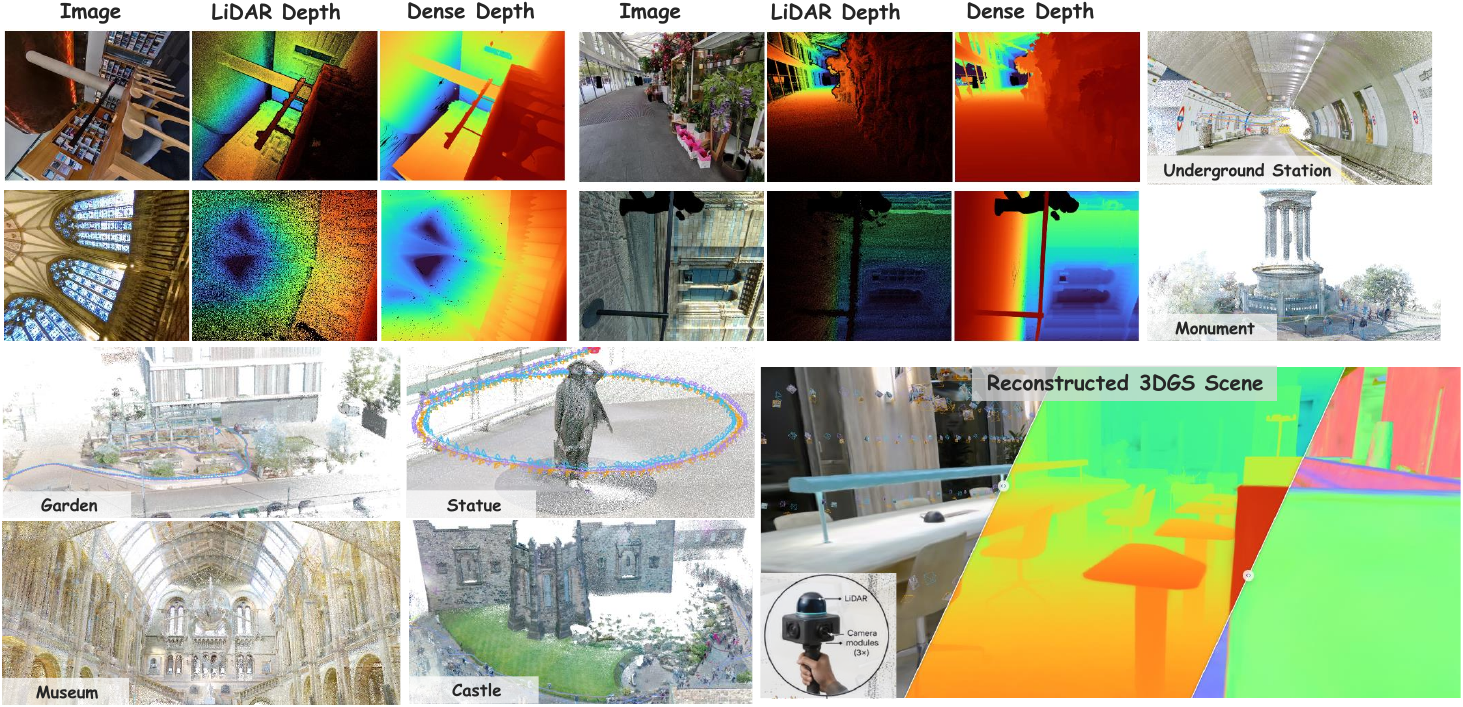}
\caption{\textbf{Overview of the \oursdata dataset.} \emph{Top left:} RGB images with sparse LiDAR depth, 
dense depth rendered from the 3DGS~\cite{DBLP:journals/tog/KerblKLD23} reconstruction, for forward- and upward-looking cameras indoors and outdoors; the dense depth fills the gaps between LiDAR returns, and masked regions such as the sky stay invalid. \emph{Right and bottom left:} LiDAR maps of indoor and outdoor scenes 
with camera trajectories drawn as colored curves. \emph{Bottom right:} a reconstructed 3DGS scene (RGB, depth, and normal) with its trajectory.}
\label{fig:overview_spatial_arena}
\end{figure*}

\paragraph{Data capture.}
Public benchmarks rarely provide metric ground truth for large, uncontrolled scenes captured by handheld devices. We therefore collect \oursdata with a handheld scanner that records LiDAR point clouds and three synchronized cameras
; the front camera faces the direction of walking, and the two side cameras look up and sideways. The benchmark uses 30 scenes, 13 indoor and 17 outdoor: museum halls, station concourses, a library, an office floor, and a stairwell; streets, alleys, gardens, markets, plazas, and monuments (\cref{fig:overview_spatial_arena}). Walks range from 18\,m to over 1\,km, and LiDAR depth from below 1\,m to over 100\,m. In total, \oursdata contributes 32,655 unique images, selected per task as described in \suppref{data}.

\paragraph{Processing.}
The final LiDAR points and camera poses are obtained by LiDAR SLAM and post-processed by iterative global bundle adjustment~\cite{DBLP:conf/cvpr/SchonbergerF16}. Projecting the LiDAR map into every camera with a z-buffer gives sparse metric depth, and SAM~3.1~\cite{DBLP:journals/corr/abs-2511-16719} masks people, vehicles, and sky, which are excluded from the valid pixels. To densify the sparse depth, we fit 3D Gaussian Splats (3DGS)~\cite{DBLP:journals/tog/KerblKLD23} to each scene, supervised by LiDAR depth and regularized by depth--normal and multi-view geometric consistency, and render dense depths for images as pseudo-labels; the projected LiDAR depth remains the sparse metric reference. We validate every rendered depth map against the sparse LiDAR depth: over all MonoDepth and VideoDepth frames, $98.3\%$ and $98.2\%$ of the LiDAR pixels agree within $\pm25\%$ (AbsRel $0.036$ and $0.030$). Since LiDAR has no returns where the rendering fails most often, \eg, near-camera floaters, reflections, and weakly textured surfaces, MonoDepth also selects its images automatically: it keeps only images whose render PSNR exceeds the scene median, so ground truth comes from well-reconstructed views, and samples 10\% of each scene's images spread over time. Finally, every depth frame is checked manually next to its LiDAR and dense depth; frames with depth artifacts are removed from MonoDepth 
or left unscored in VideoDepth. 

\subsection{Geometry Alignment}
\label{sec:align}

Most methods predict geometry only up to an unknown scale, and depth also up to an unknown shift, so every metric is computed after aligning the prediction to the ground truth: one depth alignment for both depth tasks and one trajectory alignment for all pose-based metrics.

\subsubsection{Depth Alignment}
\label{sec:method:depthalign}

Depth alignment estimates an unknown scale and shift between prediction and ground-truth depth.
Models predict in different spaces, \ie, affine-invariant disparity~\cite{DBLP:journals/pami/RanftlLHSK22,DBLP:conf/iclr/HeL0LLZZLC25}, affine-invariant depth~\cite{DBLP:conf/cvpr/KeOHMDS24}, or depth up to scale~\cite{DBLP:conf/cvpr/WangCKV0N25,DBLP:conf/iclr/LinCLCLZPGZSFK26}, and existing protocols align them differently: by least squares in depth~\cite{DBLP:conf/cvpr/KeOHMDS24} or by $\ell_1$ with a fixed number of gradient steps~\cite{DBLP:conf/cvpr/WangZHEK25,DBLP:conf/iclr/WangZZCZLCPSH26}, some only on pixels where the prediction is valid. These choices are not neutral: on identical predictions (\suppref{depth}), least squares in depth raises the AbsRel of VGGT-$\Omega$ and DA3-G by 55--59\% over a robust fit, and a depth-space fit nearly triples the AbsRel of Lotus-D. We therefore align every model with the same robust, deterministic estimator on pixels chosen by the ground truth alone, and fit it in the space in which the model predicts: in depth for depth outputs and in inverse depth for disparity outputs.

\paragraph{Ground-truth-only pixel domain.}
For frame $i$, $\Omega_i$ contains the pixels whose ground-truth depth is finite, inside the dataset's valid range, and inside its mask. The prediction plays no role in this choice. A non-finite prediction on $\Omega_i$ fails the case instead of silently shrinking the set, so every method is fitted and scored on the same pixels.

\paragraph{Native-space parameterization.}
Let $\hat{x}$ be the prediction in its native output space and $x$ the ground truth in the same space: $\hat{x}=\hat{D}$ and $x=D$ for models that predict depth, and $\hat{x}=\hat{q}$ and $x=1/D$ for models that predict disparity $\hat{q}$. A relative depth model is affine-invariant only in the space it was trained in, \eg, disparity for MiDaS-style models~\cite{DBLP:journals/pami/RanftlLHSK22,DBLP:conf/cvpr/YangKHXFZ24} and depth for Marigold~\cite{DBLP:conf/cvpr/KeOHMDS24}, so an affine map matches its ambiguity only in that space. 
We estimate the scale $a$ and shift $b$ that map the prediction onto the ground truth,
\begin{equation}(a^*,b^*)=\arg\min_{a,b}\sum_{i=1}^{N}\sum_{p\in\Omega_i}\rho_\delta\!\left(\frac{a\,\hat{x}_i(p)+b-x_i(p)}{m}\right),
  \label{eq:huber_align}
\end{equation}
where the sum runs over the $N$ frames that share one transform ($N=1$ in MonoDepth), $\rho_\delta$ is the Huber loss~\cite{huber1964robust} with $\delta=0.05$, and $m$ is the median of $x$ over all $\Omega_i$. Dividing by $m$ makes $\delta$ a relative tolerance that does not depend on the units or depth range of a dataset. Residuals below $\delta$ are fitted in the least-squares sense, while larger ones, such as depth discontinuities, thin structures, and moving objects, have only linear influence.

\paragraph{Solver.}
We minimize \cref{eq:huber_align} by iteratively reweighted least squares~\cite{holland1977robust}, starting from the least-squares solution and solving each weighted fit in closed form until convergence. Unlike a fixed number of gradient steps on an $\ell_1$ cost, the result does not depend on a learning rate or on the scale of the prediction. The aligned depth is $\tilde{D}=a^*\hat{D}+b^*$ for depth outputs and $\tilde{D}=1/\max\!\big(a^*\hat{q}+b^*,\,\epsilon\big)$ for disparity outputs, clipped to the dataset's valid range; the floor $\epsilon$ keeps every pixel of $\Omega$ finite and scored. \Suppref{depth} gives the implementation details.

\subsubsection{Trajectory Alignment}
\label{sec:method:align}

SfM, SLAM, and MVRecon need a similarity transform $S=(s, R,\mathbf{t})$, with scale $s>0$, rotation $R\in\mathrm{SO}(3)$, and translation $\mathbf{t}\in\mathbb{R}^3$, that maps the predicted trajectory onto the ground truth: a predicted camera center $\hat{\mathbf{c}}$ is mapped to $sR\,\hat{\mathbf{c}}+\mathbf{t}$. Published results almost always use the closed-form Umeyama fit on camera centers~\cite{DBLP:journals/pami/Umeyama91}, as implemented in evo~\cite{grupp2017evo}. It has two weaknesses: (1) The least-squares fit runs over all frames, so a single tracking-loss jump or a few wrong frames drag the whole alignment. This \emph{hides} error: the reported ATE falls below the true one. (2) The rotation about a straight path cannot be observed from camera centers, so the fit is ill-conditioned on near-collinear trajectories and singular on exactly straight ones. Driving and walking recordings are often nearly straight. We use one estimator for all three tasks and validate it in 
\suppref{traj}.

\paragraph{Pose-aware cost.}
Every prediction contains full poses, so we use the orientations as well as the centers. Let $\hat{R}_i,R_i$ be the predicted and ground-truth world-to-camera rotations of frame $i$, and $\hat{\mathbf{c}}_i,\mathbf{c}_i$ the camera centers. Following GTSAM's trajectory aligner~\cite{dellaert2012gtsam}, $S$ minimizes a weighted sum of per-frame rotation and translation residuals:
\begin{align}
  E(S)&=\sum_{i}w_i\left(\frac{\|\mathbf{r}^{R}_i\|^2}{\sigma_R^2}+\frac{\|\mathbf{r}^{t}_i\|^2}{\sigma_t^2}\right),
  \label{eq:pose_cost}\\
  \mathbf{r}^{R}_i&=\mathrm{Log}\big(R\,\hat{R}_i^{\top}R_i\big),\qquad
  \mathbf{r}^{t}_i=sR\,\hat{\mathbf{c}}_i+\mathbf{t}-\mathbf{c}_i,
  \label{eq:pose_residual}
\end{align}
where $\mathrm{Log}$ maps a rotation to axis-angle vector, $w_i$ is the weight of frame $i$, and $\sigma_R,\sigma_t$ are robust scales of the two residuals ($w_i=\sigma_R=\sigma_t=1$ in a plain fit). 

\paragraph{Solver.}
We minimize \cref{eq:pose_cost} with Levenberg--Marquardt in three steps. An LMedS RANSAC~\cite{rousseeuw1984least} over two-frame minimal samples with a fixed seed, followed by a refit on its inliers, gives an initialization that escapes tracking-loss jumps and gross outliers. Because no subset of frames is exactly right under smooth drift, we then refit on all frames by iteratively reweighted least squares with Cauchy weights on the normalized residuals; gross errors keep a near-zero weight, the result no longer depends on the RANSAC seed, and on clean data it is as precise as a least-squares fit. Finally, if the scale collapses toward zero, which happens only for trajectories that fail outright, we fall back to a center-only fit and flag the case. Details are in \suppref{traj}. 

\subsection{Evaluation Tasks}
\label{sec:evaltasks}

\begin{table*}[t]
\centering
\scriptsize
\setlength{\tabcolsep}{3pt}
\caption{\textbf{Evaluation tasks of \ours.} For each task: the input tracks, the datasets, the alignment applied before scoring (\cref{sec:align}), and the reported metrics. SLAM is additionally split into short, long, and large range tracks by input length. \Suppref{data} lists the scenes and images of every dataset, and \suppref{metrics} defines the metrics. $^\dagger$Kept in the per-dataset tables but excluded from the aggregate ranking (\suppref{data}).}
\label{tab:tasks}
\begin{tabular}{>{\raggedright\arraybackslash}p{0.09\textwidth}>{\raggedright\arraybackslash}p{0.09\textwidth}>{\raggedright\arraybackslash}p{0.365\textwidth}>{\raggedright\arraybackslash}p{0.13\textwidth}>{\raggedright\arraybackslash}p{0.26\textwidth}}
\toprule
Task & Tracks & Datasets & Alignment & Metrics \\
\midrule
MonoDepth & \texttt{rgb} &
NYUv2~\cite{DBLP:conf/eccv/SilbermanHKF12}, KITTI (Eigen)~\cite{DBLP:conf/cvpr/GeigerLU12}, ETH3D~\cite{DBLP:conf/cvpr/SchopsSGSSPG17}, DIODE~\cite{DBLP:journals/corr/abs-1908-00463}, HAMMER~\cite{DBLP:conf/cvpr/JungRZBLVSZAILN23}, iBims-1~\cite{DBLP:conf/eccv/KochLF018}, Spring$^\dagger$~\cite{DBLP:conf/cvpr/MehlSJNB23}, Synth4K~\cite{DBLP:conf/cvpr/YuLWLWZWZHP26}, AerialMetric~\cite{DBLP:conf/eccv/SongCZZFYHCC26}, \oursdata &
affine, per image &
AbsRel, $\delta_\tau$ ($\tau\in\{1.03,1.25,1.25^2,1.25^3\}$), normal angular error \\
VideoDepth & \texttt{rgb}, \texttt{rgbc} &
Sintel~\cite{DBLP:conf/eccv/ButlerWSB12}, Bonn~\cite{DBLP:conf/iros/PalazzoloBLGS19}, KITTI~\cite{DBLP:conf/cvpr/GeigerLU12}, DyCheck~\cite{DBLP:conf/nips/GaoLTRK22}, \oursdata &
affine, per sequence &
AbsRel, $\delta_\tau$ \\
MVRecon & \texttt{rgb}, \texttt{rgbc}, \texttt{rgbd} &
7-Scenes~\cite{DBLP:conf/cvpr/ShottonGZICF13}, ScanNet++$^\dagger$~\cite{DBLP:conf/iccv/YeshwanthLND23}, HiRoom~\cite{DBLP:conf/iclr/LinCLCLZPGZSFK26}, NRGBD~\cite{DBLP:conf/cvpr/AzinovicMGNT22}, ETH3D~\cite{DBLP:conf/cvpr/SchopsSGSSPG17}, Tanks\&Temples~\cite{DBLP:journals/tog/KnapitschPZK17}, DTU~\cite{DBLP:journals/ijcv/AanaesJVTD16}, \oursdata &
$\mathrm{Sim}(3)$ on poses; TSDF fusion (\oursdata: back-projection) &
Acc, Comp, Overall; precision, recall, F-score \\
\bottomrule
\end{tabular}
\end{table*}

\Cref{tab:tasks} summarizes the input tracks, datasets, alignment, and metrics of the five tasks. Dataset details are given in \suppref{data}, and metric definitions in \suppref{metrics}.

\paragraph{Depth.}
MonoDepth scores images after a per-image affine alignment
; besides the usual $\delta_{1.25}$, we report the strict $\delta_{1.03}$, which still separates models that saturate $\delta_{1.25}$, and, where intrinsics are available, the angular error of normals derived from the aligned depth. VideoDepth fits one affine transform per sequence, so per-frame scale drift is penalized.

\paragraph{Reconstruction and poses.}
MVRecon turns depths and poses into a reconstruction exactly as a downstream user would: the poses are aligned with the $\mathrm{Sim}(3)$ estimator of \cref{sec:method:align}, the recovered scale rescales the depths, and all views are fused into a TSDF volume~\cite{DBLP:conf/siggraph/CurlessL96} (on \oursdata, back-projected directly into one point cloud), which is compared with a laser scan, a mesh, or our LiDAR map. SfM tests camera recovery from unordered images, 
including \oursdata scenes that mix three rig cameras looking in different directions. SLAM instead evaluates tracking over continuous video: each case is one complete recording, fed in a single pass at a fixed stride. 
Because sequence length dominates both difficulty and memory, we split SLAM into short, long, and large-range tracks (up to 364, 1,211, and 18,846 input frames), and within each track and dataset, we rank methods by the mean of their ATE and drift ranks.

\subsection{Evaluated Models}
\label{sec:method:models}

\begin{table}[t]
\centering
\scriptsize
\setlength{\tabcolsep}{3pt}
\caption{\textbf{Evaluated models of \ours}. Tasks (union over the family): M = MonoDepth, V = VideoDepth, R = MVRecon, S = SfM, L = SLAM. $^\ast$Also accepts geometric priors (cameras and/or depth). $^\S$Commercial model, queried through its API
. \Suppref{models} gives the tasks, tracks, and preprocessing of every model.}
\label{tab:models}
\begin{tabular}{>{\raggedright\arraybackslash}p{0.18\linewidth}>{\raggedright\arraybackslash}p{0.09\linewidth}>{\raggedright\arraybackslash}p{0.64\linewidth}}
\toprule
Family (\#) & Tasks & Models \\
\midrule
Feed-forward 3D/4D (20) & MVRSL &
DUSt3R~\cite{DBLP:conf/cvpr/Wang0CCR24}, MASt3R~\cite{DBLP:conf/eccv/LeroyCR24}, Pow3R~\cite{DBLP:conf/cvpr/JangW0AR25}, CUT3R~\cite{DBLP:conf/cvpr/WangZHEK25}, Fast3R~\cite{DBLP:conf/cvpr/YangSLHTCCMF25}, VGGT~\cite{DBLP:conf/cvpr/WangCKV0N25}, FastVGGT~\cite{DBLP:journals/corr/abs-2509-02560}, VGGT-$\Omega$~\cite{DBLP:conf/cvpr/WangCZKSLBNVR26}, RobustVGGT~\cite{DBLP:conf/cvpr/HanHJJAWKF26}, OmniVGGT$^\ast$~\cite{DBLP:conf/cvpr/PengLDLLQZZZXL26}, $\pi^3$-X$^\ast$~\cite{DBLP:conf/iclr/WangZZCZLCPSH26}, DA3-L, DA3-G, DA3-NGL$^\ast$~\cite{DBLP:conf/iclr/LinCLCLZPGZSFK26}, AMB3R~\cite{DBLP:conf/cvpr/WangA26}, MapAnything$^\ast$~\cite{DBLP:conf/3dim/KeethaMSPZFKZWALLBRRSK26}, D\'ej\`a View~\cite{DBLP:journals/corr/abs-2605-30215}, Co-Me~\cite{DBLP:conf/cvpr/ChenQLPS26}, IGGT~\cite{DBLP:conf/iclr/LiZLZHCLZYZL26}, IGGT4D~\cite{DBLP:journals/corr/abs-2607-19228} \\
Monocular depth (10) & M &
Marigold v1.0~\cite{DBLP:conf/cvpr/KeOHMDS24}, v1.1~\cite{DBLP:journals/corr/abs-2505-09358}, Marigold V2~\cite{DBLP:journals/corr/abs-2609-08084}, Lotus-D~\cite{DBLP:conf/iclr/HeL0LLZZLC25}, MoGe~\cite{DBLP:conf/cvpr/WangXDX00Y25}, MoGe-2~\cite{DBLP:conf/nips/WangXDDXLSTY25}, MoGe-3~\cite{DBLP:journals/corr/abs-2607-17967}, Depth Pro~\cite{DBLP:conf/iclr/BochkovskiyDGSZ25}, UniDepthV2~\cite{DBLP:conf/cvpr/PiccinelliYSSLG24}, InfiniDepth~\cite{DBLP:conf/cvpr/YuLWLWZWZHP26} \\
Video generative (6) & MVSL &
DVD~\cite{DBLP:journals/corr/abs-2603-12250}, DepthCrafter~\cite{DBLP:conf/cvpr/0002G0ZCZQS25}, ChronoDepth~\cite{DBLP:conf/cvpr/ShaoYZZSGWPL25}, GenCeption (depth, pose)~\cite{DBLP:conf/eccv/WangZKUWZCHABZS26}, GeoNeXt~\cite{DBLP:journals/corr/abs-2608-28549} \\
\bottomrule
\end{tabular}
\end{table}

We evaluate 52 models in five families (\cref{tab:models}). \emph{Feed-forward 3D/4D} models predict cameras, depth, and point maps; they enter the depth, MVRecon, and SfM tasks, and most of them also SLAM; four of them also accept geometric priors and run the \texttt{rgbc} or \texttt{rgbd} tracks. \emph{Monocular depth} models, both affine-invariant and metric, enter MonoDepth. \emph{Video generative} models are built on video diffusion or generation backbones and predict video depth, camera poses, or both. \emph{Long-sequence and SLAM} systems chunk, stream, or compress input to process thousands of frames and enter SLAM only. Finally, \emph{vision--language} models are queried for depth, which places them on the same scale as dedicated geometry models; besides open models, we query three commercial models (Claude Opus 5.5, GPT-6 Astra, and Gemini 4 Argon) through APIs on MonoDepth. 
Variants of models, such as the depth and pose variants of GenCeption, count as separate models. We list the input resolution and preprocessing of every model in \cref{sec:suppl:preprocess}.

%

\section{Experiments}
\label{sec:exp}

\begin{table*}[t]
\centering
\scriptsize
\setlength{\tabcolsep}{2pt}
\setlength{\aboverulesep}{0pt}
\setlength{\belowrulesep}{0pt}
\renewcommand{\arraystretch}{1.12}
\caption{\textbf{Leaderboard of \ours}
. Every (task, dataset, metric) cell min--max normalizes the scores of the models that report it to $[0,100]$. A task score 
is the mean over the cells of that task, and Score is the mean over the five task scores, with a task the model does not support counted as 0; higher is better. AR is the average of the five task ranks (a missing task takes the last rank plus one; lower is better). Cov.\ is the number of scored cells, and Wins is the number of cells a model ranks first in. \colorbox{rkgold}{Gold}, \colorbox{rksilver}{silver}, and \colorbox{rkbronze}{bronze} mark the top three of each column, small numbers give the rank. Only the \texttt{rgb} track enters the ranking, and the datasets marked $\dagger$ in \cref{tab:tasks} are excluded; \oursdata is scored like a public dataset, and a model without an \oursdata result in a task counts as 0 in those cells. \Cref{tab:suppl:leaderboard_public} gives the scores on the public datasets only.}
\label{tab:leaderboard}
\adjustbox{max width=\textwidth}{
\begin{tabular}{@{}rlrrrrrrrrr@{\hspace{5pt}}|@{\hspace{5pt}}rlrrrrrrrrr@{}}
\toprule
\# & Method & Score & AR & Cov. & Wins & Mono & Video & MVR & SfM & SLAM & \# & Method & Score & AR & Cov. & Wins & Mono & Video & MVR & SfM & SLAM \\
\midrule
1 & VGGT-$\Omega$ & \rkA{87.4} & \rkA{2.6} & 415 & 1121 & \rkA{96.7} & \rkA{96.5} & 83.8\rkn{4} & \rkA{94.2} & 65.8\rkn{6} & 27 & LingBot-Map & 17.1 & 32.4 & 100 & 3 & -- & -- & -- & -- & \rkB{85.3} \\
2 & $\pi^3$-X & \rkB{84.3} & \rkB{5.0} & 415 & 612 & \rkC{91.4} & \rkB{91.9} & \rkC{87.7} & 87.9\rkn{9} & 62.7\rkn{8} & 28 & SenseNova-Vision & 16.9 & 35.6 & 122 & 8 & 37.0\rkn{35} & 47.4\rkn{23} & -- & -- & -- \\
3 & D\'ej\`a View & \rkC{81.7} & \rkC{6.2} & 415 & 811 & 81.6\rkn{12} & \rkC{86.7} & 81.9\rkn{7} & \rkB{93.3} & 64.8\rkn{7} & 29 & Marigold V2 & 16.6 & 34.0 & 33 & 14 & 82.8\rkn{10} & -- & -- & -- & -- \\
4 & IGGT4D & 79.0 & 8.0 & 415 & 655 & 81.8\rkn{11} & 84.1\rkn{6} & 82.4\rkn{6} & 88.4\rkn{7} & 58.3\rkn{10} & 30 & Depth Pro & 16.0 & 34.8 & 33 & 4 & 79.9\rkn{14} & -- & -- & -- & -- \\
5 & DA3-G & 77.2 & 7.2 & 415 & 828 & 86.3\rkn{8} & 85.6\rkn{5} & \rkA{89.8} & 90.6\rkn{4} & 33.6\rkn{18} & 31 & VGGT-Long & 15.8 & 32.6 & 100 & 2 & -- & -- & -- & -- & \rkC{78.9} \\
6 & DA3-NGL & 76.8 & 7.0 & 415 & 900 & 85.1\rkn{9} & 86.3\rkn{4} & \rkB{88.1} & \rkC{90.9} & 33.8\rkn{17} & 32 & InfiniDepth & 15.6 & 35.6 & 33 & 3 & 78.0\rkn{18} & -- & -- & -- & -- \\
7 & DA3-L & 76.2 & 10.4 & 415 & 471 & 78.3\rkn{17} & 82.2\rkn{7} & 82.9\rkn{5} & 84.8\rkn{10} & 52.8\rkn{13} & 33 & Lotus-D & 15.1 & 35.8 & 33 & 1 & 75.3\rkn{19} & -- & -- & -- & -- \\
\bottomrule
\end{tabular}}
\end{table*}

\begin{table*}[t]
\centering
\scriptsize
\setlength{\tabcolsep}{3pt}
\setlength{\aboverulesep}{0pt}
\setlength{\belowrulesep}{0pt}
\renewcommand{\arraystretch}{1.12}
\caption{\textbf{MonoDepth, \texttt{rgb} track}, averaged over all datasets, including \oursdata (AbsRel and $\delta$ in \%, normal error in degrees); models are sorted by Score, continuing from the left part to the right. Score is the task score of \cref{tab:leaderboard}; colors and ranks as in \cref{tab:leaderboard}. $^\circ$Partial coverage (\eg, out of memory on some cases): the means cover the evaluated cases and are not ranked. \Cref{tab:suppl:res_mono} lists every metric, and \cref{tab:suppl:mono_public} gives the public datasets only.}
\label{tab:exp:mono}
\adjustbox{max width=\textwidth}{
\begin{tabular}{@{}lrrrrr@{\hspace{5pt}}|@{\hspace{5pt}}lrrrrr@{\hspace{5pt}}|@{\hspace{5pt}}lrrrrr@{}}
\toprule
Method & Score$\uparrow$ & AbsRel$\downarrow$ & $\delta_{1.03}$$\uparrow$ & $\delta_{1.25}$$\uparrow$ & Normal$\downarrow$ & Method & Score$\uparrow$ & AbsRel$\downarrow$ & $\delta_{1.03}$$\uparrow$ & $\delta_{1.25}$$\uparrow$ & Normal$\downarrow$ & Method & Score$\uparrow$ & AbsRel$\downarrow$ & $\delta_{1.03}$$\uparrow$ & $\delta_{1.25}$$\uparrow$ & Normal$\downarrow$ \\
\midrule
VGGT-$\Omega$ & \rkA{96.7} & \rkB{10.76} & \rkA{49.85} & \rkB{89.79} & 26.49 & Depth Pro & 79.9\rkn{14} & 15.50 & 35.18 & 83.59 & 29.60 & FastVGGT & 61.8\rkn{27} & 19.81 & 28.01 & 76.05 & 32.91 \\
MoGe-3 & \rkB{92.1} & 11.57 & \rkC{46.21} & 87.34 & \rkA{24.39} & RobustVGGT & 79.2\rkn{15} & 15.50 & 38.66 & 82.81 & 29.18 & Co-Me & 58.8\rkn{28} & 20.86 & 25.72 & 74.53 & 34.96 \\
$\pi^3$-X & \rkC{91.4} & \rkC{11.38} & 45.02 & 87.45 & 26.16 & VGGT & 79.2\rkn{15} & 15.50 & 38.66 & 82.81 & 29.18 & Gemini 4 Argon & 56.6\rkn{29} & 22.95 & 20.16 & 73.65 & 40.21 \\
AMB3R & 91.3\rkn{4} & 11.93 & 44.32 & 87.43 & \rkC{25.68} & DA3-L & 78.3\rkn{17} & 16.92 & 37.03 & 81.97 & 29.11 & Fast3R & 56.4\rkn{30} & 26.78 & 28.20 & 73.55 & 35.53 \\
UniDepthV2 & 91.0\rkn{5} & 12.73 & 45.97 & \rkC{88.26} & 28.07 & InfiniDepth & 78.0\rkn{18} & 16.47 & 37.19 & 82.96 & 32.65 & IGGT$^\circ$ & 49.7\rkn{31} & 18.42 & 30.34 & 79.11 & 28.67 \\
MoGe & 90.7\rkn{6} & 12.31 & 45.19 & 86.66 & \rkB{24.43} & Lotus-D & 75.3\rkn{19} & 16.53 & 36.54 & 82.29 & 38.03 & Claude Opus 5.5 & 44.4\rkn{32} & 26.66 & 16.10 & 66.48 & 41.60 \\
MoGe-2 & 90.5\rkn{7} & 12.01 & 45.30 & 86.93 & 25.69 & MASt3R & 74.4\rkn{20} & 18.84 & 35.21 & 81.30 & 31.62 & Marigold v1.1 & 41.9\rkn{33} & 27.74 & 20.41 & 66.88 & 36.52 \\
\bottomrule
\end{tabular}}
\end{table*}

\newcommand{\trendpanel}[3][4.9cm]{%
  \begin{tabular}[b]{@{}c@{}}\includegraphics[height=#1]{images/exp/#2}\\[-1pt]\footnotesize #3\end{tabular}}
\begin{figure*}[t]
\centering
\includegraphics[width=\textwidth]{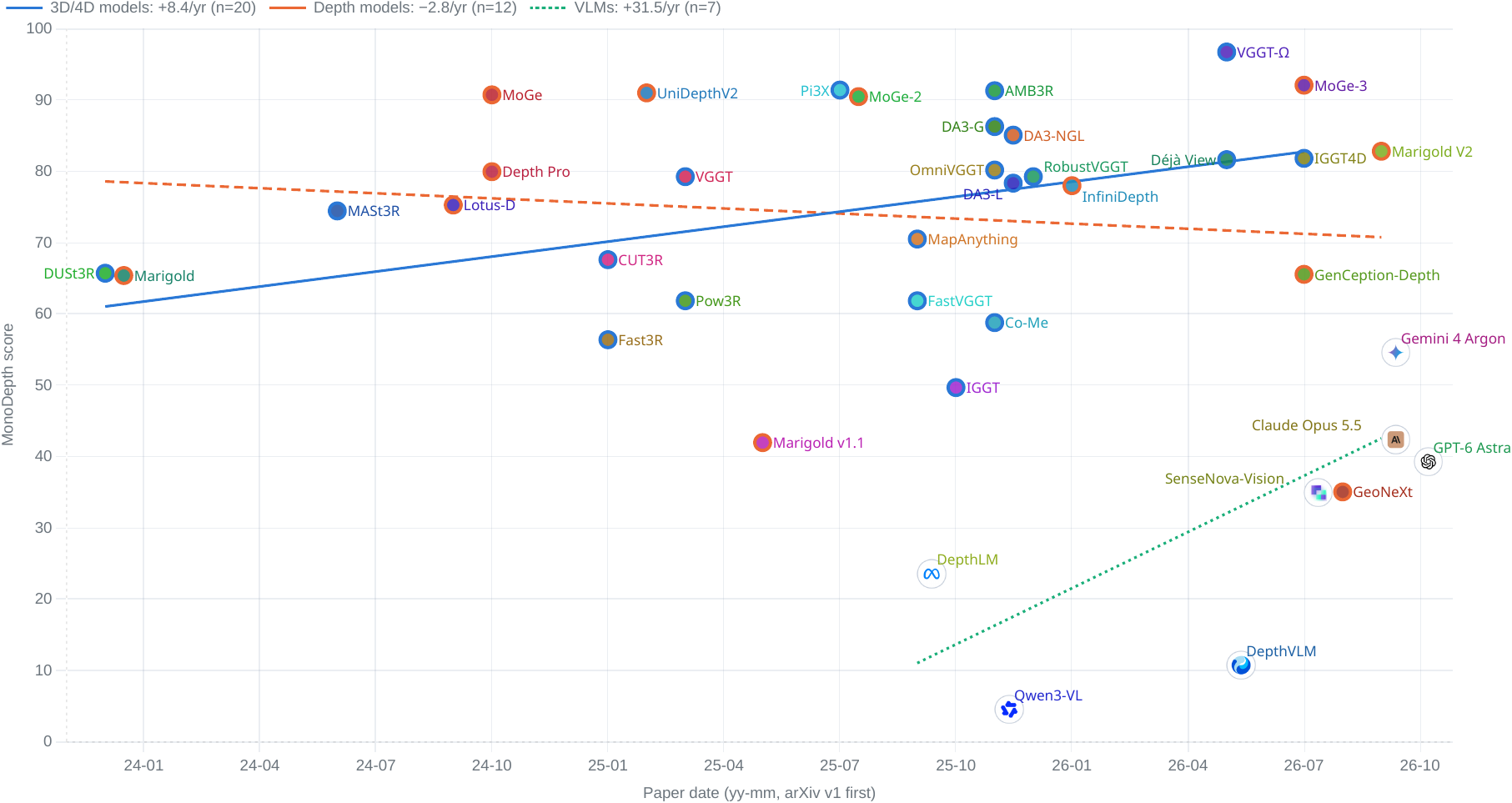}
\caption{\textbf{MonoDepth score vs.\ paper date}. 
Lines: per-family least-squares fits; legend: slope and score. 
Other tasks in \cref{fig:suppl:trends,fig:suppl:trends2}.}
\label{fig:exp:trends}
\end{figure*}

\begin{figure}[t]
\centering
\includegraphics[width=\linewidth]{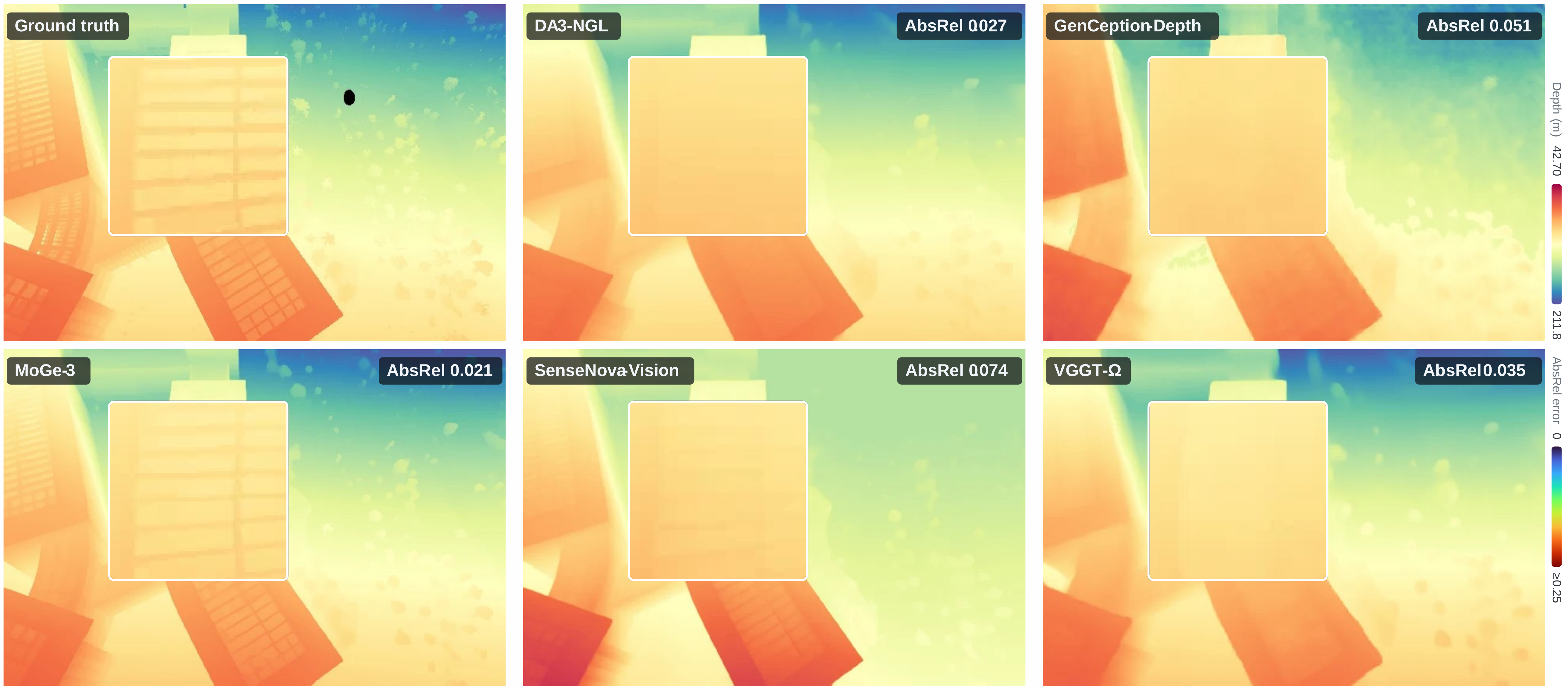}
\caption{\textbf{MonoDepth predictions on AerialMetric}, aligned and colored on the ground-truth scale (black: masked region); each tag gives the frame's AbsRel as a fraction.
}
\label{fig:exp:depth}
\end{figure}

\newcommand{\reconpanel}[2]{%
  \begin{tabular}[b]{@{}c@{}}\includegraphics[width=\textwidth]{images/artifacts/#1}\\[-1pt]\footnotesize #2\end{tabular}}
\begin{figure*}[t]
\centering
\reconpanel{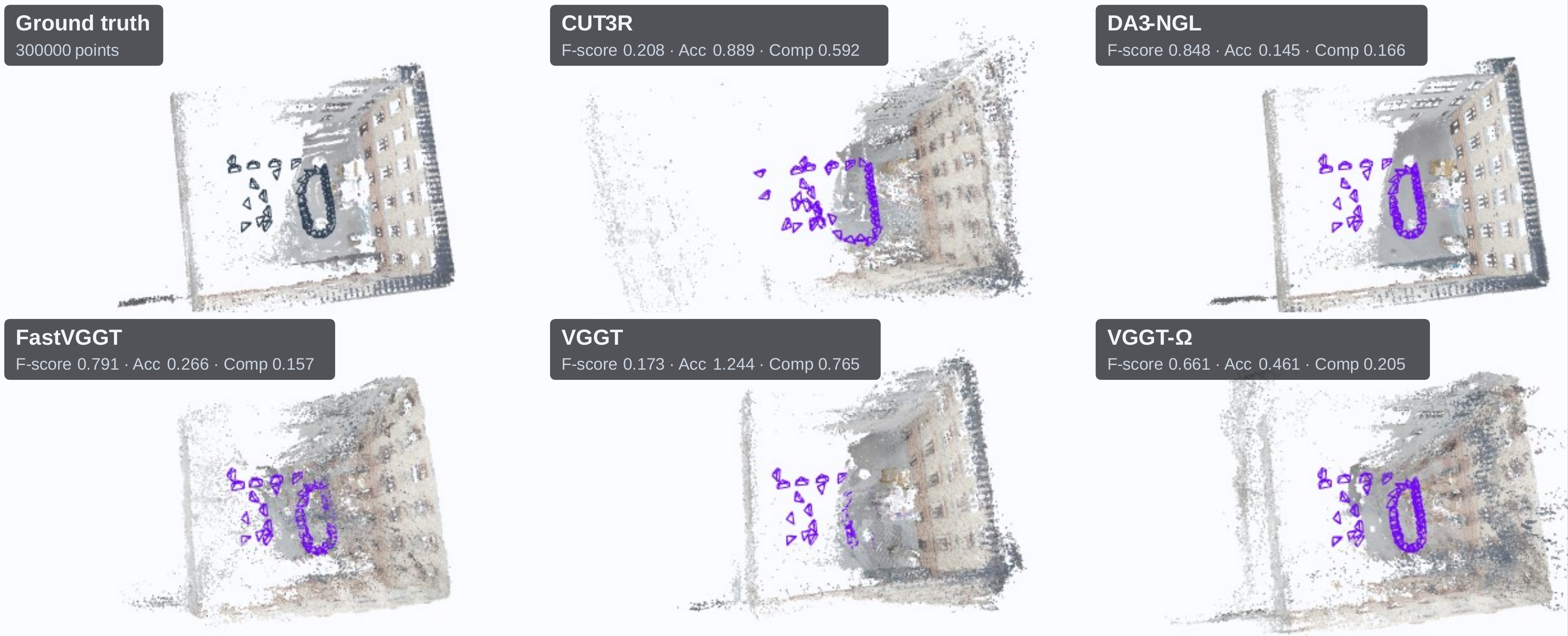}{(a) ETH3D, courtyard}\\[6pt]
\reconpanel{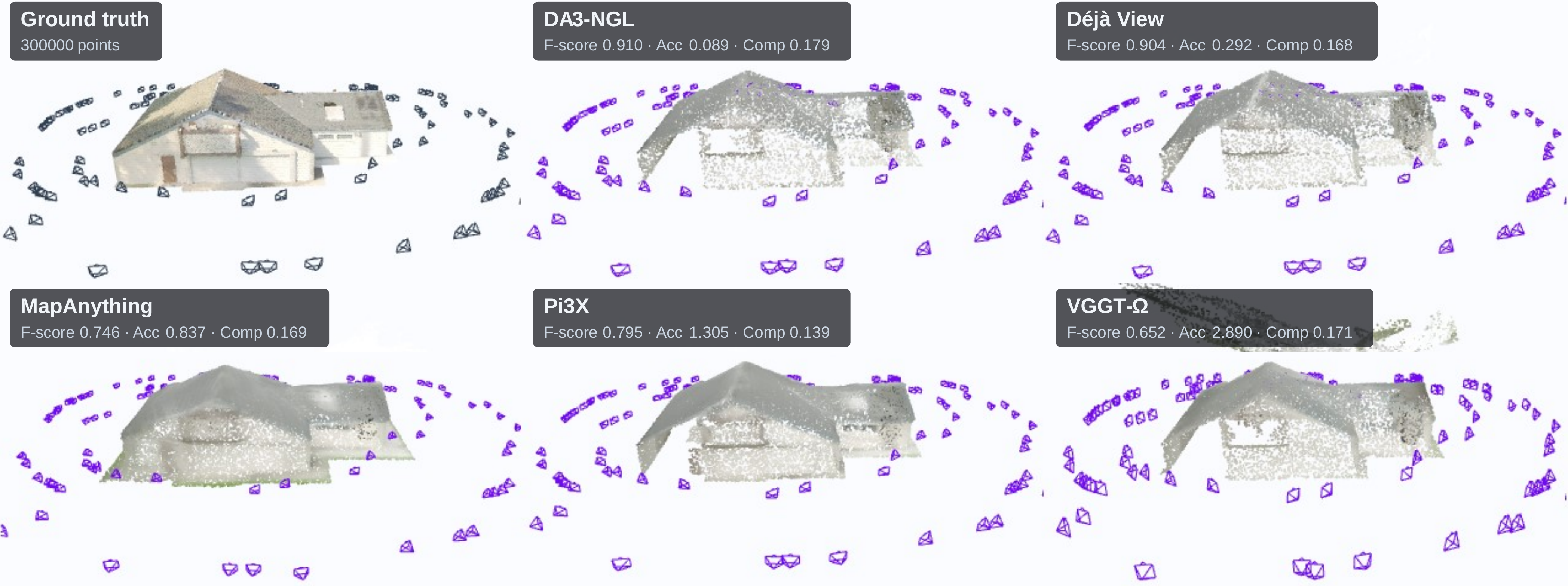}{(b) Tanks\&Temples, Barn}
\caption{\textbf{MVRecon reconstructions} on ETH3D and Tanks\&Temples, aligned to the ground truth as in evaluation. Ground-truth cameras are drawn in black and predicted cameras in purple. The tag on each reconstruction gives its F-score ($\tau=25$\,cm), Acc, and Comp (in m).}
\label{fig:exp:recon}
\end{figure*}

\Cref{tab:leaderboard} ranks every model with results by its overall score, and \cref{tab:exp:mono} gives the headline MonoDepth metrics on the \texttt{rgb} track, averaged over its datasets. Both score and average \oursdata like a public dataset, as do all tables and trends of this paper; \cref{sec:suppl:res_public} gives the leaderboard and the MonoDepth table on the public datasets only. Metrics for the other tasks are provided in \suppref{results}. 
%
\Cref{fig:exp:trends} plots the MonoDepth scores against each model's release date (the other tasks are in \cref{fig:suppl:trends,fig:suppl:trends2}), and \cref{fig:exp:depth,fig:exp:recon} show example depth predictions and reconstructions. More visual results are in \cref{sec:suppl:res_qual}.

\paragraph{Leaderboard.}
VGGT-$\Omega$ ranks first overall, and first in MonoDepth, VideoDepth, and SfM. 
The other task leaders differ: DA3-G leads MVRecon, and MoGe-3, the best depth specialist, ranks second in MonoDepth. In SLAM, VGGT-$\Omega$ leads the short and long ranges, while the large range, which only long-sequence systems enter (\suppref{data}), is led by Pi-Long.

\paragraph{Progress over time.}
Do newer models perform better? \Cref{fig:exp:trends} and the VideoDepth plot in \cref{fig:suppl:trends} answer this for the two depth tasks, and the answer differs between model families. On MonoDepth, the depth specialists have stalled: their fitted trend falls by 2.8 points per year, and the best of them has improved its AbsRel only from 12.3\% (MoGe, 2024) to 11.6\% (MoGe-3, 2026). The 3D/4D models rise by 8.4 points per year, and since mid-2025 the best of them has been more accurate than every depth specialist evaluated on all datasets ($\pi^3$-X, 11.4\%; VGGT-$\Omega$, 10.8\%). Their advantage comes from \oursdata, the only MonoDepth dataset on which the 3D/4D models are clearly more accurate than the specialists (\cref{sec:suppl:disc_general}); on the public datasets alone, the two families rise by only 1.9--2.8 points per year and remain on par (\cref{sec:suppl:res_public}). On VideoDepth, which requires depth that is consistent across frames, the 3D/4D models improve by 28.4 points per year; their best AbsRel fell from 45\% (DUSt3R) to 6.5\% (VGGT-$\Omega$) in two and a half years and has been below that of the best video depth model (GenCeption-Depth, 14.0\%) since mid-2025, while depth models improve by 8.7 points per year. VLMs improve fastest on MonoDepth (31.5 points per year) from a low base, but on VideoDepth more slowly than the 3D/4D models (14.3), and they still trail the other two families on both tasks. Progress comes mainly from multi-view consistency rather than from single-image accuracy; MVRecon, SfM, and SLAM show the same rise for 3D/4D models (\cref{fig:suppl:trends,fig:suppl:trends2}). No model leads on every dataset (\cf \cref{fig:teaser}(c)).

\paragraph{Key findings and outlook.}
We summarize the results along three axes; \cref{sec:discussion} discusses each in depth, with the failure cases we observed.
(1)~\emph{Generalizability.} General models are absorbing specialized ones: the feed-forward 3D/4D models enter all five tasks, lead MonoDepth once \oursdata is scored, lead VideoDepth, and still rise steeply on MVRecon (24.7 points per year) and SLAM (25.5). No model, however, generalizes across all data. The dataset leaders differ even among the top-ranked models (\cref{fig:teaser}(c)); on the seven ranked MVRecon datasets, six different models have the best F-score. All models degrade sharply on the large-scale scenes of \oursdata: the best AbsRel is 14.2\% on MonoDepth, against 2.4--8.0\% on the public datasets and well above the 3.6\% agreement of ground truth with the LiDAR depth, and the best MVRecon F-score of a model that reconstructs all six scenes is 0.20 (IGGT), against 0.29--0.95. Large, uncontrolled scenes, rather than more of the existing test sets, are where generalization still has to be won.
(2)~\emph{Robustness.} 
No feed-forward model runs the large SLAM range on its own, and the long-sequence systems often fail: ZipMap runs out of GPU memory on every Oxford Spires and VBR sequence, and Scal3R runs out of memory on six of the seven VBR sequences. 
The range is led by Pi-Long, which chunks the input of a feed-forward model. 
Bounded-memory inference and failure detection, more than further gains on short clips, are what long-sequence models need next.
(3)~\emph{Accuracy.} Coarse accuracy is saturated on the public datasets, but fine accuracy is not. There, the best depth model reaches a $\delta_{1.25}$ of 96\%, gaining under one point per year, but a $\delta_{1.03}$ of only 62\%; with \oursdata scored, the two fall to 87\% and 46\% (MoGe-3). For VLMs, accuracy is the main bottleneck. Their best MonoDepth AbsRel fell from 33\% to 23\% within one year (DepthLM, 2025, to Gemini 4 Argon, 2026), a factor of 1.4; if this rate holds, VLMs would reach the best depth specialist (11.6\%) only in the second half of 2028. On VideoDepth their error barely fell, from 30\% to 27\% in ten months (DepthLM to SenseNova-Vision), against 6.5\% for VGGT-$\Omega$, so parity is not in sight at this rate. Their MonoDepth $\delta_{1.03}$ is at most 20\%, so metric precision and consistency across frames are what VLMs most need to become geometric foundation models.

\paragraph{Qualitative results.}
\Cref{fig:exp:depth} compares MonoDepth predictions on AerialMetric. All models recover the coarse layout. MoGe-3, 2nd in MonoDepth, has the lowest AbsRel (2.1\%, against 3.5\% for VGGT-$\Omega$, 1st). The per-frame errors do not always follow the leaderboard: on the Spring and \oursdata frames (\cref{fig:suppl:depth}), VGGT-$\Omega$ is best (8.3\% and 13.2\%), most models smooth away thin structures such as the trees in Spring, and on the \oursdata frame VGGT, 15th in MonoDepth, has the highest error (35.6\%). The same holds for MVRecon (\cref{fig:exp:recon}): DA3-NGL, 2nd in MVRecon, has the highest F-score on both scenes (0.848 and 0.910), and FastVGGT is far more accurate than VGGT on the courtyard (0.791 against 0.173), whereas VGGT-$\Omega$, 4th in MVRecon, places a fragment of the Barn above the scene, which raises its Acc to 2.89\,m. On the large SLAM range (\cref{fig:exp:slam_traj}), LingBot-Map stays closest to the ground truth on the sequences shown (ATE 6.9\,m on Oxford Spires, against 13.3\,m for Pi-Long), though Pi-Long has the lower ATE on every dataset but VBR; other long-sequence systems drift or collapse onto a small region.

\section{Discussion}
\label{sec:discussion}\label{sec:suppl:discussion}

The central finding of our evaluation is that models which are accurate on the public benchmarks fail in systematic ways on large, uncontrolled scenes. In this section, we analyze why, along the three axes of \cref{sec:exp}: generality (\cref{sec:suppl:disc_general}), robustness (\cref{sec:suppl:disc_robust}), and accuracy (\cref{sec:suppl:disc_accuracy}). We then summarize the strengths and weaknesses of each model family (\cref{sec:suppl:disc_summary}) and their implications for future work (\cref{sec:suppl:disc_future}). Throughout, $\rho$ denotes Spearman's rank correlation, and ``public'' denotes a model's mean over the ranked public datasets of a task, which exclude ScanNet++ and Spring (\cref{sec:suppl:leakage}).

\begin{figure*}[t]
\centering
\includegraphics[width=\textwidth]{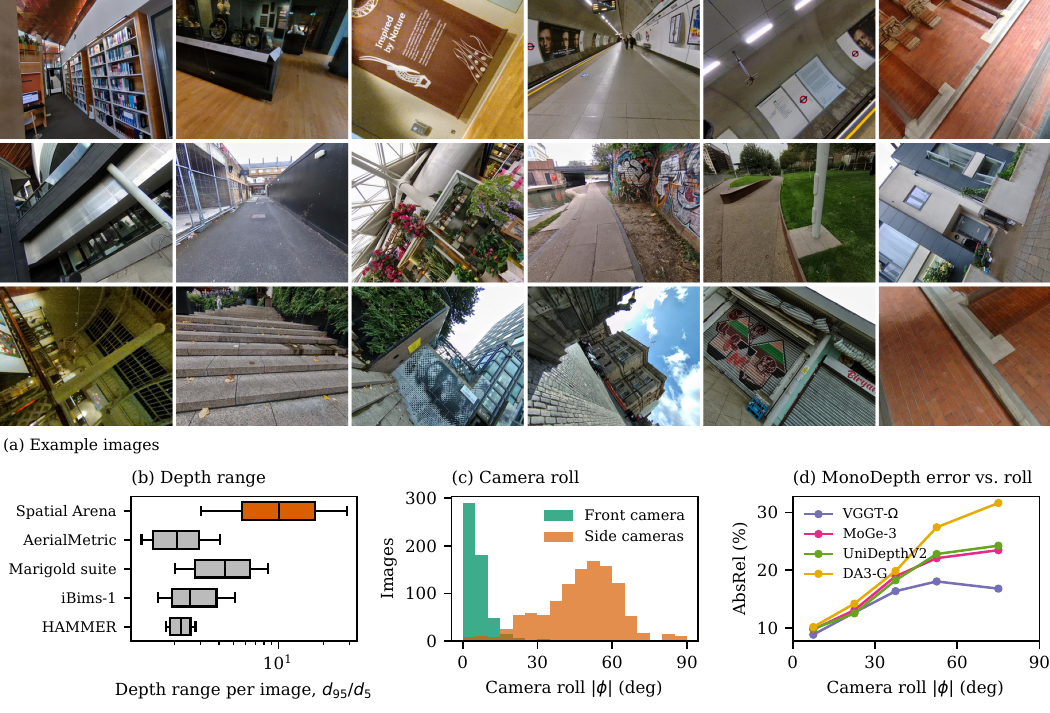}
\caption{\textbf{What makes \oursdata difficult.} (a) Images from 14 scenes, taken by the front camera (upright views along the walking direction) and the two side cameras (rolled views looking up and sideways): dim museum halls with specular floors, a tiled underground platform, close-ups of floors and signs, graffiti, foliage, stairs, and facades seen from below. (b) Depth range within one image ($d_{95}/d_{5}$ of the valid ground truth; box: quartiles, whiskers: 10th and 90th percentiles) on the 1,688 \oursdata MonoDepth images and a sample of the real public MonoDepth datasets (Marigold suite without ETH3D). (c) Absolute camera roll of the same images, from the ground-truth poses (world $z$ is vertical): the front camera is nearly upright (median $5^\circ$), whereas the side cameras are rolled by a median of $50^\circ$. (d) Mean MonoDepth AbsRel of four models in bins of camera roll.}
\label{fig:suppl:sa_diversity}
\end{figure*}

\begin{table*}[t]
\centering
\scriptsize
\setlength{\tabcolsep}{4pt}
\caption{\textbf{Public datasets against \oursdata.} For each model that we evaluate on all ranked public datasets of a task and on \oursdata, we compare its mean over the public datasets with its \oursdata result. We give the best and the median over these $n$ models, the median and range of the per-model ratio \oursdata/public, Spearman's $\rho$ between the two rankings, and the leader on each. The SLAM public datasets are 7-Scenes and TUM RGB-D (short range) and Oxford Spires, KITTI Odometry, and VBR (large range). $^\ast$Without Fast3R, whose public F-score is 0.022 (ratio 2.94).}
\label{tab:suppl:sa_gap}
\begin{tabular}{@{}llr rr rr cc ll@{}}
\toprule
& & & \multicolumn{2}{c}{Public} & \multicolumn{2}{c}{\oursdata} & & & \multicolumn{2}{c}{Leader} \\
\cmidrule(lr){4-5}\cmidrule(lr){6-7}\cmidrule(l){10-11}
Task & Metric & $n$ & Best & Median & Best & Median & Ratio (range) & $\rho$ & Public & \oursdata \\
\midrule
MonoDepth & AbsRel (\%) $\downarrow$ & 37 & 4.03 & 6.80 & 14.18 & 22.96 & 3.28 (1.29--4.23) & 0.82 & MoGe-3 & VGGT-$\Omega$ \\
 & $\delta_{1.03}$ (\%) $\uparrow$ & 37 & 67.6 & 51.7 & 35.9 & 25.3 & 0.45 (0.34--0.84) & 0.89 & VGGT-$\Omega$ & VGGT-$\Omega$ \\
VideoDepth & AbsRel (\%) $\downarrow$ & 26 & 5.29 & 12.65 & 10.96 & 21.69 & 1.84 (0.87--3.54) & 0.73 & VGGT-$\Omega$ & VGGT-$\Omega$ \\
 & $\delta_{1.03}$ (\%) $\uparrow$ & 26 & 60.0 & 34.7 & 34.6 & 16.7 & 0.43 (0.20--0.91) & 0.91 & VGGT-$\Omega$ & VGGT-$\Omega$ \\
MVRecon & F-score $\uparrow$ & 16 & 0.628 & 0.471 & 0.197 & 0.122 & 0.25 (0.18--0.57)$^\ast$ & 0.40 & DA3-G & IGGT \\
SfM & AUC@$5^\circ$ $\uparrow$ & 18 & 0.610 & 0.500 & 0.270 & 0.043 & 0.11 (0.00--0.45) & 0.88 & DA3-NGL & D\'ej\`a View \\
 & AUC@$30^\circ$ $\uparrow$ & 18 & 0.901 & 0.832 & 0.743 & 0.404 & 0.48 (0.01--0.83) & 0.87 & VGGT-$\Omega$ & D\'ej\`a View \\
SLAM, short & Drift (\%) $\downarrow$ & 26 & 12.7 & 21.3 & 13.4 & 121.7 & 5.08 (0.49--10.85) & 0.07 & VGGT-$\Omega$ & ABot-Recon \\
SLAM, large & Drift (\%) $\downarrow$ & 7 & 14.5 & 83.4 & 27.0 & 80.1 & 1.24 (0.27--3.31) & 0.64 & ABot-Recon & Pi-Long \\
\bottomrule
\end{tabular}
\end{table*}

\begin{figure*}[t]
\centering
\includegraphics[width=\textwidth]{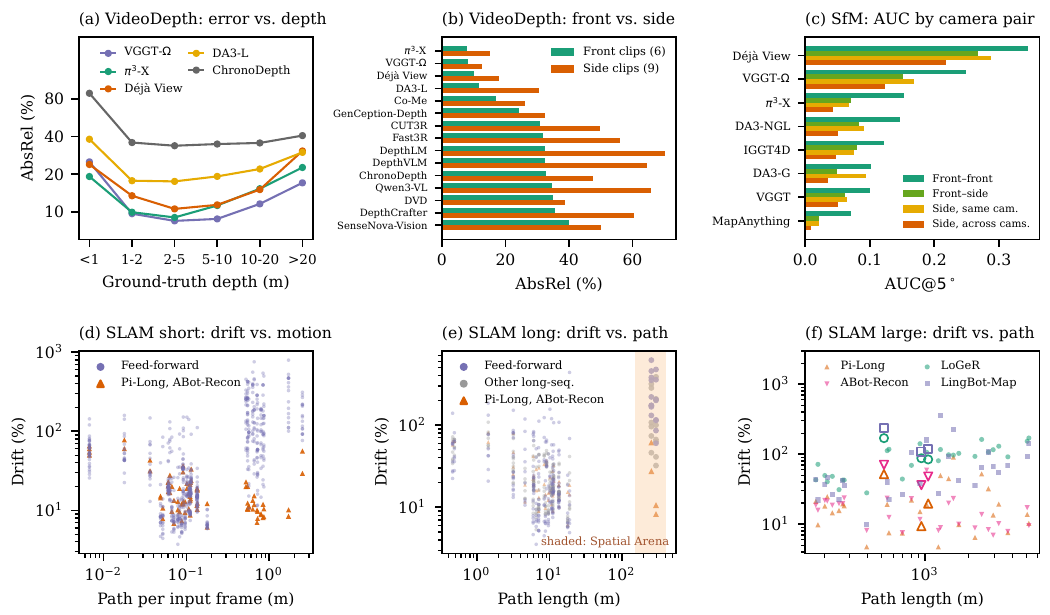}
\caption{\textbf{Why models fail on \oursdata, beyond MonoDepth.} (a) VideoDepth AbsRel of the \oursdata pixels in bins of ground-truth depth (every fourth frame of all 15 clips, with each model's sequence alignment). (b) Mean VideoDepth AbsRel on the six front-camera and nine side-camera \oursdata clips, for the 15 models without auxiliary inputs that cover all clips. (c) SfM AUC@$5^\circ$ on the seven \oursdata scenes, computed separately for image pairs from two front images, from a front and a side image, from the same side camera, and from different side cameras. (d) Drift of every model on every short-range SLAM sequence (7-Scenes, TUM RGB-D, Oxford Spires, and \oursdata) against the ground-truth path length per input frame. (e) Drift on every long-range sequence (7-Scenes, TUM RGB-D, and the two \oursdata walks, shaded) against the path length. (f) Drift of four long-sequence systems on every large-range sequence against the path length; open markers are \oursdata.}
\label{fig:suppl:failure_analysis}
\end{figure*}

\subsection{Generality}
\label{sec:suppl:disc_general}

\paragraph{Across tasks.}
Single-image accuracy is a poor predictor of multi-view performance. The task scores are strongly correlated among VideoDepth and the multi-view tasks ($\rho=0.88$ between VideoDepth and MVRecon, $0.77$ between VideoDepth and SLAM, and $0.74$ between MVRecon and SLAM), but only weakly between MonoDepth and the pose-dependent tasks ($0.52$ with SfM, $0.47$ with SLAM); DA3-G, for instance, leads MVRecon but ranks eighth in MonoDepth. We attribute this to the different cues that the tasks reward. A single image can only be resolved through learned priors about scene layout, whereas consistent depth and poses across views require correspondence, which the multi-view models learn and the monocular ones do not.

\paragraph{What makes \oursdata difficult.}
\Cref{fig:suppl:sa_diversity} shows how \oursdata differs from the public datasets. First, its viewpoints are unusual. The front camera looks along the walking direction and is pitched down by a median of $25^\circ$, but the two side cameras look up and sideways and are rolled by a median of $50^\circ$ (10th to 90th percentile, $26$--$65^\circ$; \cref{fig:suppl:sa_diversity}(c)), so that the horizon runs diagonally through the image and facades, ceilings, and sky fill large parts of the view. The public datasets consist almost entirely of upright images taken by a person, a robot, or a car. Second, a single image spans a wide range of depth: the median ratio between the 95th and 5th percentile of ground-truth depth is $10\times$ on \oursdata and reaches $29\times$ at the 90th percentile, against $2.1$--$4.4\times$ on HAMMER, iBims-1, the Marigold suite, and AerialMetric (\cref{fig:suppl:sa_diversity}(b)). A typical side view contains a wall or floor within a meter and a facade tens of meters away. Third, the camera moves fast relative to the frame rate: on the short SLAM range, it moves 0.55\,m between input frames, against 0.05--0.10\,m on TUM RGB-D and 7-Scenes. Fourth, the scenes vary in content and appearance (\cref{fig:suppl:sa_diversity}(a)): dim museum halls with specular floors and glass cases, a tiled underground platform, graffiti and foliage, stairs, and crowded markets, all captured with a $90^\circ$ horizontal field of view.

\paragraph{Why models fail on \oursdata: MonoDepth.}
These properties, rather than the scale of the scenes alone, explain most of the drop in accuracy (\cref{tab:suppl:sa_gap}). To show this, we analyze the stored predictions of each task per image, per pixel, or per image pair, without re-running any model. In MonoDepth, we analyze ten models (VGGT-$\Omega$, $\pi^3$-X, AMB3R, MoGe-3, UniDepthV2, Depth Pro, DA3-G, VGGT, DUSt3R, and SenseNova-Vision). The clearest evidence is the camera roll. Within the same scenes, the models are about twice as accurate on the upright front camera as on the rolled side cameras: the side-to-front ratio of the AbsRel has a median of 2.14 for VGGT-$\Omega$, 2.16 for MoGe-3, and 2.28 for UniDepthV2, and it exceeds one in 18 or 19 of the 19 scenes. The error grows steadily with roll (\cref{fig:suppl:sa_diversity}(d)): for MoGe-3, from 9.8\% below $15^\circ$ to 22.1\% at $45$--$60^\circ$ and 23.5\% at $60$--$90^\circ$. On the upright front camera, VGGT-$\Omega$ reaches 8.4\%, within $1.8\times$ of its public error, whereas on the side cameras it reaches 16.2--18.0\%. VGGT-$\Omega$ degrades least at large roll (16.8\% at $60$--$90^\circ$, against 23.5--31.6\% for MoGe-3, UniDepthV2, and DA3-G), which suggests that more diverse camera orientations in training reduce this dependence.

\findingbox{Current models rely on a gravity-aligned prior, with the ground at the bottom and the sky at the top of the image, that they acquire from upright training images; they have not learned to predict depth independently of in-plane rotation.}

The wide depth range is a second cause. Binning the pixels of all 1,688 images by ground-truth depth shows that the error is lowest between 1 and 5\,m (11--19\% for the six best models) and rises both below 1\,m (25--35\%) and beyond 20\,m (20--26\%; 31--38\% for Depth Pro and VGGT). Beyond 20\,m, all ten models underestimate depth, by a mean log error of $-0.18$ (UniDepthV2) to $-0.62$ (DUSt3R) and $-1.03$ (SenseNova-Vision), \ie, they compress distant structure towards the camera. Because our alignment fits one scale and shift per image, it cannot undo such a non-linear compression, and the error remains in the score. The compression is consistent with training sets whose dense depth is either indoor (a few meters) or capped at the range of a driving LiDAR, so that near and far structure rarely appear in one training image.

\findingbox{Depth models are calibrated to the narrow depth range of their training images: they are accurate at 1--5\,m but compress distant structure towards the camera, an error that no per-image scale and shift can remove.}

\paragraph{Why models fail on \oursdata: VideoDepth.}
A video provides temporal context and, to the multi-view models, parallax, which could in principle resolve both failures. We test this on the 15 \oursdata clips (\cref{fig:suppl:failure_analysis}(a,b)). It does not. Binning the pixels by depth (every fourth frame, five models) reproduces the MonoDepth profile: the error of VGGT-$\Omega$ is 8.5\% at 2--5\,m and rises to 25.1\% below 1\,m and 17.0\% beyond 20\,m, and beyond 20\,m, all five models underestimate depth, by a mean log error of $-0.14$ (VGGT-$\Omega$) to $-0.59$ (ChronoDepth). ChronoDepth also overestimates depth below 1\,m ($+0.47$); it flattens the whole depth range, which explains its 88.6\% error at close range and its failure on the far opening of the corridor in \cref{fig:suppl:video_depth_sa}(a). The camera dependence persists as well: all 15 models that cover every clip are less accurate on the nine side-camera clips than on the six front-camera clips, by $1.1$--$2.7\times$ (median $1.7\times$). The gap is smallest for DVD ($1.1\times$) and SenseNova-Vision ($1.3\times$), but only because they are inaccurate already on the front clips (35.0\% and 40.2\%, against 8.3\% for VGGT-$\Omega$). Only one walk contributes clips from both kinds of camera, so scene content adds to this gap, but its size and consistency across models agree with the MonoDepth analysis. The multi-view models suffer less from the compression than the video diffusion model (log error $-0.14$ to $-0.45$ against $-0.59$), which indicates that parallax helps, but not enough to recover distant structure.

\findingbox{Temporal context does not remove either failure. Video models inherit the gravity prior and the depth compression of single-image models; parallax from multi-view models reduces the compression but does not remove it, and diffusion-based video depth flattens the whole depth range.}

\paragraph{Why models fail on \oursdata: SfM.}
Each \oursdata SfM scene mixes the upright front camera with the two rolled side cameras, whose views point in different directions and overlap little with each other. We therefore split the image pairs of the seven scenes by camera and compute the AUC@$5^\circ$ separately for each kind of pair (eight models, \cref{fig:suppl:failure_analysis}(c)). For every model, pairs of two front images are the most accurate and pairs from the two different side cameras the least accurate, by $1.6$--$7.4\times$ (VGGT-$\Omega$: 0.250 against 0.124; MapAnything: 0.071 against 0.010). Pairs from the same side camera, which share the roll but view similar content, lie in between. Two factors therefore compound: the roll of the side cameras lowers accuracy even when the views overlap, and the small overlap between differently oriented cameras lowers it further. A third factor limits all pairs alike. The models process the $1600{\times}1296$ images at a width of 504--518 pixels (\cref{tab:suppl:preprocess}), so that one 14-pixel patch spans about $2.5^\circ$ of the $90^\circ$ field of view, half of the $5^\circ$ threshold. The coarse loop is therefore recovered (median AUC@$30^\circ$ 0.404), but fine accuracy is not (median AUC@$5^\circ$ 0.043). Finally, because the models regress poses directly from image tokens, without the geometric verification by which COLMAP rejects matches between similar-looking but distinct places, repeated structure produces spurious jumps (\cref{fig:suppl:sfm_stgiles}).

\findingbox{Feed-forward pose estimation is reliable only for upright views that overlap strongly. Rolled views, small overlap between differently oriented cameras, a working resolution of about $2.5^\circ$ per token, and the absence of geometric verification together explain the scattered cameras and spurious jumps on \oursdata.}

\paragraph{Why models fail on \oursdata: SLAM.}
On the short range, the 14 feed-forward models that drift by at most 18.2\% on 7-Scenes drift by 42--221\% on \oursdata, although the sequences have a similar number of input frames (325 on median). The difference lies in the camera motion: the short \oursdata sequences cover 179\,m, so the camera moves 0.55\,m between input frames, against 0.10\,m on 7-Scenes and 0.05\,m on TUM RGB-D. Over all short-range sequences of all datasets, the drift of every model except Pi-Long and ABot-Recon increases with this motion (\cref{fig:suppl:failure_analysis}(d); median $\rho=0.48$ over the feed-forward models). Below 0.1\,m per frame, their median drift is 11--21\%; at 0.3\,m or more, which only \oursdata and Oxford Spires reach, it is 33--268\%. Large motion reduces the overlap between consecutive frames and produces baselines outside the range of the video clips on which these models are typically trained, so consecutive poses are estimated from little shared content, and their errors accumulate. Only Pi-Long and ABot-Recon behave in the opposite way: they drift less on such sequences (10.4\% and 11.0\%; $\rho=-0.37$ and $-0.59$). The camera roll plays no role here, since five of the eight \oursdata SLAM sequences use the front camera.

The long range separates the number of frames from the length of the walk. On the public datasets, it consists only of indoor sequences (7-Scenes and TUM RGB-D) whose camera paths are 6--10\,m long, and it feeds the same 7-Scenes recordings as the short range at a smaller stride, \ie, with about three times as many frames over the same path. The additional frames change nothing: for seven of the nine feed-forward models that run both \oursdata ranges, the median drift on the long 7-Scenes range is within $0.95$--$1.02\times$ of the short-range drift (CUT3R and Fast3R are the exceptions), \eg, 11.7\% against 12.2\% for VGGT-$\Omega$. The two long \oursdata walks cover 254 and 294\,m in 1,001 and 1,148 frames (\cref{fig:suppl:failure_analysis}(e)). Although the camera moves less per frame than on the short \oursdata range (0.25 against 0.55\,m), the median drift of the feed-forward models doubles (median $1.98\times$), to 96--466\% (VGGT-$\Omega$, 168\%; $\pi^3$-X, 144\%), against 7--123\% on the public long range. Errors thus accumulate along the path once the walk extends far beyond the local context of the model, which the indoor sequences of the public long range never test. The long-sequence systems split into two groups. Pi-Long (17.9\%), ABot-Recon (36.1\%), LoGeR (43.6\%), and VGGT-Long (53.0\%) keep the drift low, whereas ZipMap (352\%) and R3 (246\%) drift as much as the feed-forward models. On the large range, where the public datasets do contain outdoor paths (median 1.7\,km on KITTI Odometry and 1.6\,km on VBR), the same split appears within systems (\cref{fig:suppl:failure_analysis}(f)): the drift of LoGeR and LingBot-Map grows with the path length ($\rho=0.67$ and $0.51$ over all large-range sequences), whereas that of Pi-Long and ABot-Recon does not ($\rho=-0.04$ and $-0.42$), and Pi-Long drifts by 19.7\% on \oursdata and 16.9\% on the public large range.

\findingbox{Drift grows with the camera motion between frames and, once the walk leaves the local context of the model, with the length of the walk; the number of frames alone does not matter. Feed-forward models track well only on short, slow, indoor paths. Only long-sequence systems that bound the error per chunk, most consistently Pi-Long, keep the drift low on long outdoor walks, which \oursdata provides on every range.}

\paragraph{Why models fail on \oursdata: MVRecon.}
MVRecon combines the two previous mechanisms. It aligns the predicted cameras to the ground truth, recovers the scale from the camera centers, and fuses the depth maps with it. Depth that is compressed at long range then places distant surfaces too close to the cameras, and pose errors among the rolled side views misplace whole depth maps; both move points beyond the 25\,cm threshold. This is consistent with the reconstructions of \cref{fig:suppl:recon_fail,fig:suppl:recon_fail_roof}, whose shapes are recognizable but whose scale and placement are wrong. We did not measure the error of the fused points per distance, so this explanation rests on the depth and pose analyses above rather than on a direct measurement.

\findingbox[Key finding across tasks]{The failures on \oursdata trace to three properties that the public benchmarks lack: rolled views, which break the gravity prior of depth and pose models; a wide depth range within one image, which every model compresses at long range; and large camera motion between frames over long walks, along which feed-forward models accumulate drift. None of these properties depends on the scale of a scene alone.}

\paragraph{Across data: rankings change.}
Because these failure modes affect models to different degrees, public rankings transfer only partly. The rank correlation between public and \oursdata results is high for depth ($\rho=0.73$--$0.91$) and SfM ($0.87$--$0.88$), low for MVRecon (0.40), and absent for short-range SLAM (0.07), and the leader changes in every task except VideoDepth. In MonoDepth, the three most accurate public models are monocular specialists (MoGe-3, MoGe-2, MoGe), whereas on \oursdata the three most accurate are 3D/4D models (VGGT-$\Omega$, AMB3R, $\pi^3$-X); in MVRecon, IGGT ranks 13th of 16 on the public datasets (F-score 0.348) but first on \oursdata (0.197), whereas the public leader DA3-G falls to eighth (0.122); in SfM, DA3-G falls from second to ninth in AUC@$5^\circ$; and in short-range SLAM, VGGT-$\Omega$ has the lowest public drift (12.7\%) but drifts by 66.2\% on \oursdata. A model that is selected on public benchmarks is therefore not necessarily the best choice for large, uncontrolled scenes.

\paragraph{Across inputs.}
\label{sec:suppl:disc_priors}
Additional inputs should not degrade a model, but the \texttt{rgbc} and \texttt{rgbd} tracks show that for current models they sometimes do (\cref{tab:exp:aux,tab:suppl:res_aux}). Ground-truth cameras help weaker reconstructions: the MVRecon F-score of MapAnything increases from 0.264 to 0.746 on ETH3D. In contrast, DA3-NGL degrades severely when conditioned on cameras (mean F-score 0.667 to 0.222; 0.870 to 0.086 on HiRoom), and depth priors halve the SfM AUC@$5^\circ$ of $\pi^3$-X on DTU (0.743 to 0.336) and raise its MVRecon Acc on 7-Scenes from 0.24\,m to 20.5\,m. A prior that conflicts with the image evidence, in convention, scale, or density, is evidently not reconciled with it but followed. This points to conditioning that was trained on clean, consistent priors and is not robust to the conventions and noise of real ones.

\subsection{Robustness}
\label{sec:suppl:disc_robust}

\paragraph{Failure cases on \oursdata: depth.}
The qualitative results show the failures analyzed in \cref{sec:suppl:disc_general}. On the \oursdata frame of \cref{fig:suppl:depth}, VGGT-$\Omega$ has an AbsRel of 13.2\% and VGGT of 35.6\%, although the two models differ by only 1.4 points on the public datasets. In VideoDepth (\cref{fig:suppl:video_depth_sa}), ChronoDepth misses the far opening of a corridor (0.171, against 0.044 for D\'ej\`a View and $\pi^3$-X), an instance of the compressed far range. On a close-up of a nearly planar museum surface, every model fails to some degree (0.117--0.375), and VGGT-$\Omega$ and DA3-NGL, first and third on the public VideoDepth datasets, are among the least accurate. Such a view contains no horizon, ground plane, or familiar object that would anchor a learned layout prior. The VLMs fail even on the coarse layout: in the library corridor (\cref{fig:suppl:depth_vlm}), Claude Opus 5.5 and GPT-6 Astra return nearly planar depth (0.399 and 0.392).

\paragraph{Failure cases on \oursdata: reconstruction and poses.}
In MVRecon, no model exceeds an F-score of 0.35 on the scenes of \cref{fig:suppl:recon_fail}, and on the roof (\cref{fig:suppl:recon_fail_roof}), most models recover the shape but at the wrong scale and position. In SfM, all models recover the overall loop around the Dugald Stewart Monument (\cref{fig:suppl:sfm_monument}), but their cameras scatter around it (best AUC@$5^\circ$ 0.230, D\'ej\`a View; MapAnything 0.005). Around the St Giles' statue (\cref{fig:suppl:sfm_stgiles}), whose sides look alike, DA3-NGL and VGGT add spurious jumps across the loop. These are the effects of camera roll, small overlap, low working resolution, and missing geometric verification analyzed in \cref{sec:suppl:disc_general}.

\paragraph{Failure cases on \oursdata: long recordings.}
The 14 feed-forward models that drift by at most 18.2\% on the short 7-Scenes sequences drift by 42--221\% on those of \oursdata. As analyzed in \cref{sec:suppl:disc_general}, the cause is the camera motion between frames. On the large range (\cref{fig:suppl:slam_traj}), only Pi-Long keeps the drift below 50\% (27.0\%, ATE 13.1\,m); the other systems drift by 51--155\% or collapse onto a small region, and VGGT-SLAM returns poses for only 10\% of the frames. Pi-Long chunks the input of a strong feed-forward model and aligns overlapping chunks, which bounds both memory and the error per chunk. Unlike the other tasks, the large range of \oursdata is about as difficult as the public long recordings (median drift 80.1\% against 83.4\%); for SLAM, \oursdata adds most on the short and long ranges.

\begin{table*}[t]
\centering
\scriptsize
\setlength{\tabcolsep}{3pt}
\renewcommand{\arraystretch}{1.1}
\caption{\textbf{Failure modes observed in \ours}, with examples and the tables and figures that show them. Values are per-dataset means unless stated otherwise; ``per-dataset results'' are those of the leaderboard, which are not listed in the tables; the last column gives what a future model or benchmark should do about each mode.}
\label{tab:suppl:failures}
\begin{tabular}{@{}p{2.45cm}p{5.75cm}p{2.35cm}p{5.25cm}@{}}
\toprule
Failure mode & Examples & Evidence & Implication \\
\midrule
\multicolumn{4}{@{}l}{\emph{Robustness: the model does not return a usable prediction}} \\
Out of memory & DUSt3R, Pow3R: all short/long SLAM sequences but two; ZipMap: all large Oxford Spires and VBR; Scal3R: 6/7 VBR & \cref{tab:exp:slam} & Bounded-memory inference (streaming, chunking) \\
Input-length limit & GenCeption (pose): at most 81 frames, 2/18 7-Scenes and 3/9 TUM short sequences & \cref{tab:exp:slam} & Declare limits; chunk longer inputs \\
Incomplete output & VGGT-SLAM: completeness 0.28 (7-Scenes, short) to 0.02 (Oxford Spires, large) & \cref{tab:suppl:res_slam} & Report completeness next to ATE \\
Degenerate output & Qwen3-VL: constant depth on an AerialMetric scene & Per-dataset results & Self-assessed confidence; refuse rather than guess \\
Far outliers & MapAnything: MVRecon Acc 10.4\,m (66.6\,m on ETH3D); $\pi^3$-X with sparse depth: Acc 20.5\,m on 7-Scenes & \cref{tab:exp:mvrecon,tab:exp:aux} & Outlier-sensitive metrics; confidence-based filtering \\
\midrule
\multicolumn{4}{@{}l}{\emph{Generality: the prediction degrades outside the familiar data or inputs}} \\
Large outdoor scenes & Every model degrades: median MonoDepth AbsRel $\times$3.3, MVRecon F-score $\times$0.25; SfM AUC@$5^\circ$ median 0.043 (0.500 on public data); scattered cameras, spurious jumps & \cref{tab:suppl:sa_gap,fig:suppl:sfm_monument,fig:suppl:sfm_stgiles} & Train and test on large, uncontrolled scenes \\
Compressed far range & All analyzed models underestimate depth beyond 20\,m (mean log error $-0.18$ to $-1.03$); MVRecon on \oursdata: right shape, wrong scale and placement, F-score at most 0.35 & \cref{fig:suppl:sa_diversity,fig:suppl:recon_fail,fig:suppl:recon_fail_roof} & Training data with near and far structure in one image \\
Drift over long paths & Short-range median drift 21.3\% (7-Scenes, TUM) vs.\ 121.7\% (\oursdata); VGGT-$\Omega$ 12.7\% vs.\ 66.2\%; large-range best ATE 5.5--30\,m & \cref{tab:suppl:sa_gap,fig:exp:slam_traj,fig:suppl:slam_traj} & Loop closure or global optimization in the model \\
Harmful priors & DA3-NGL with cameras: MVRecon F-score 0.667 $\to$ 0.222 (0.870 $\to$ 0.086 on HiRoom); $\pi^3$-X with depth: DTU AUC@$5^\circ$ 0.743 $\to$ 0.336 & \cref{tab:exp:aux} & Train conditioning on noisy, sparse priors \\
Unusual viewpoints & Rolled side cameras: MonoDepth AbsRel $\times$2.1--2.3 that of the upright front camera in the same scenes; close-up of a planar museum surface: VideoDepth AbsRel 0.117--0.375 & \cref{fig:suppl:sa_diversity,fig:suppl:video_depth_sa} & Rotation-robust training; gravity conditioning \\
\midrule
\multicolumn{4}{@{}l}{\emph{Accuracy: the prediction is usable but imprecise}} \\
Thin structures & Spring trees smoothed away; iBims-1 leaves and railings: AbsRel $\times$2.5--4.6 & \cref{fig:suppl:depth,fig:suppl:depth_ibims} & Higher output resolution; edge-aware losses \\
Far regions & ChronoDepth misses the far opening of a corridor (AbsRel 0.171 vs.\ 0.044) & \cref{fig:suppl:video_depth_sa} & Depth range beyond the training distribution \\
Texture-poor objects & Best SfM AUC@$5^\circ$ on TexturePoorSfM: 0.10 & Per-dataset results & Priors beyond appearance matching \\
Isolated fragments & VGGT-$\Omega$ places a fragment of the Barn above the scene (Acc 2.89\,m) & \cref{fig:exp:recon} & Per-point confidence \\
Coarse VLM depth & Nearly flat depth (AbsRel 0.39--0.40) on a \oursdata frame, even at the coarse layout that a $16{\times}16$ query grid can represent & \cref{fig:suppl:depth_vlm,fig:suppl:depth_vlm_public} & Dense, metric output heads for VLMs \\
\bottomrule
\end{tabular}
\end{table*}

\paragraph{Input length and memory.}
On the public datasets, most failures stem from input length rather than from prediction errors. Global attention over all frames requires memory that grows quadratically with the number of tokens, and pairwise models must process a number of pairs that grows quadratically with the number of frames. Accordingly, DUSt3R and Pow3R exhaust memory on all but two short-range 7-Scenes sequences and on every other short- and long-range SLAM sequence, ZipMap on every large-range Oxford Spires and VBR sequence, and Scal3R on six of the seven VBR sequences (\cref{tab:exp:slam}). GenCeption (pose) accepts at most 81 frames, so we can score it on only 2 of the 18 short-range 7-Scenes sequences and 3 of the 9 TUM sequences.

\paragraph{Unreported failures and scene content.}
Some failures go unreported by the model. VGGT-SLAM returns poses for only part of the input (completeness 0.02 on the large range of Oxford Spires), and a few distant points raise the MVRecon Acc of MapAnything to 66.6\,m on ETH3D. None of these models estimates its own reliability, so a downstream system cannot distinguish such outputs from valid ones. Local failures follow the content: thin structures such as the trees in Spring are smoothed away (\cref{fig:suppl:depth}), foliage and railings raise the error of every depth specialist on iBims-1 by $2.5$--$4.6\times$ (\cref{fig:suppl:depth_ibims}), and weak texture limits pose accuracy (best AUC@$5^\circ$ on TexturePoorSfM, 0.10).

\paragraph{Robustness of the measurement.}
An evaluation protocol must itself be robust to such failures; otherwise it measures the alignment rather than the model. A least-squares depth fit increases the AbsRel of VGGT-$\Omega$ from 6.56\% to 10.40\%, and the Umeyama fit underestimates the ATE by up to $2.6\times$ under outliers or tracking loss (\cref{tab:depth_align,tab:align_real}). Results obtained with a non-robust alignment or on leaked test data (\cref{tab:suppl:leakage}) are therefore not directly comparable with ours.

\subsection{Accuracy}
\label{sec:suppl:disc_accuracy}

\paragraph{Coarse versus fine accuracy.}
On the public datasets, coarse accuracy is close to saturation, but fine accuracy is not. The best $\delta_{1.25}$ is 98.2--99.5\% on the Marigold suite, HAMMER, iBims-1, and AerialMetric, and the best SfM AUC@$30^\circ$ is 0.89--0.98 on DTU, ETH3D, HiRoom, IMC, and Tanks and Temples, whereas the best $\delta_{1.03}$ is only 70.7--82.1\% and the median AUC@$5^\circ$ is 0.36--0.74. Fine thresholds also separate the models better: on HAMMER and iBims-1, the median model is within 2.1 points of the best $\delta_{1.25}$ but 14--23 points behind the best $\delta_{1.03}$. We see two limits on fine accuracy. The models predict at a fraction of the input resolution, and their training objectives are dominated by the coarse structure of a scene, which a $\delta_{1.25}$-level prediction already captures. On \oursdata, fine accuracy is the first to break, with a best $\delta_{1.03}$ of 35.9\% (\cref{tab:suppl:sa_gap}).

\paragraph{Drift and global consistency.}
Errors grow rapidly with trajectory length: the best short-range ATE is 1.0\,cm on TUM and 4.0\,cm on 7-Scenes, whereas the best large-range drift is 9.8--27\% of the path length (\cref{tab:exp:slam}). Low drift does not imply global consistency, nor does a low ATE imply small local errors. ABot-Recon has the lowest drift on KITTI, VBR, and the short \oursdata range, but on the large range about three times the ATE of Pi-Long on Oxford Spires and \oursdata (16.0 and 36.8\,m against 5.5 and 13.1\,m), although it has the lowest ATE on VBR (16.5\,m), while LingBot-Map shows the opposite pattern. Locally accurate tracking and globally consistent maps are thus different capabilities, and only Pi-Long ranks among the three best systems in both drift and ATE on every large-range dataset.

\subsection{Strengths and weaknesses of the evaluated models}
\label{sec:suppl:disc_summary}

\begin{table*}[t]
\centering
\scriptsize
\setlength{\tabcolsep}{3pt}
\renewcommand{\arraystretch}{1.12}
\caption{\textbf{Strengths and weaknesses of the evaluated model families.} ``Public'' is the mean over the ranked public datasets of a task, as in \cref{tab:suppl:sa_gap}; errors are AbsRel unless stated otherwise.}
\label{tab:suppl:strengths}
\begin{tabular}{@{}p{2.2cm}p{5.9cm}p{6.2cm}p{1.5cm}@{}}
\toprule
Family & Strengths & Weaknesses & Use for \\
\midrule
Monocular depth (MoGe-3, MoGe-2, UniDepthV2, Marigold V2, Depth Pro) &
Lowest AbsRel on every ranked public MonoDepth dataset; MoGe-3 ranks second in MonoDepth (first without \oursdata; 4.0\% public) and has the best $\delta_{1.03}$ on iBims-1 (82.1\%). &
Single image only; no progress once \oursdata is scored ($-2.8$ points per year; $+1.9$ without); four of the five largest increases on \oursdata (MoGe-3, InfiniDepth, MoGe-2, MoGe; $4.1$--$4.2\times$; MoGe-3 17.0\%, 4th), where three 3D/4D models are more accurate (14.2--16.0\%); error more than doubles under camera roll (MoGe-3: 9.8\% below $15^\circ$, 23.5\% at $60$--$90^\circ$); smooth away thin structures. &
Single images of indoor and object-scale scenes \\
\addlinespace
Video depth (DVD, DepthCrafter, ChronoDepth) &
Temporally consistent depth without cameras; DVD is the best video depth model on public data (11\%). &
Less accurate than the best 3D/4D model on every VideoDepth dataset; collapse on \oursdata (DVD 37.3\%, ChronoDepth 41.6\%, DepthCrafter 50.7\%, against 11.0\% for VGGT-$\Omega$); miss distant regions (\cref{fig:suppl:video_depth_sa}). &
Short clips where cameras are not needed \\
\addlinespace
Pairwise 3D (DUSt3R, MASt3R, Pow3R) &
Enter all five tasks with a simple pairwise design, on which most later 3D/4D models build. &
Memory grows with the number of pairs: DUSt3R and Pow3R run out of memory on every \oursdata SLAM sequence and on almost all other short- and long-range sequences; \oursdata MonoDepth 33.7--33.8\% for DUSt3R and Pow3R (8.1--8.7\% public); errors over entire walls on HiRoom. &
Few-view reconstruction \\
\addlinespace
Multi-view 3D/4D (VGGT-$\Omega$, $\pi^3$-X, D\'ej\`a View, DA3, IGGT4D, VGGT) &
Hold the first 20 overall places; lead MonoDepth, VideoDepth, MVRecon, and SfM; on \oursdata, lead MonoDepth (VGGT-$\Omega$, 14.2\%), VideoDepth (11.0\%), and SfM (D\'ej\`a View, AUC@$5^\circ$ 0.270); fastest progress on multi-view tasks (24.7 points per year in MVRecon). &
Cannot run the large SLAM range; drift on short \oursdata sequences (VGGT-$\Omega$ 66.2\%, against 11.9\% on 7-Scenes; 42--221\% for the 14 models that drift by at most 18.2\% on 7-Scenes); MVRecon F-score at most 0.20 and median SfM AUC@$5^\circ$ 0.062 on \oursdata; isolated fragments and far outliers; inconsistent use of priors (DA3-NGL with cameras: F-score 0.667 to 0.222). DA3 models transfer least (DA3-G: SfM AUC@$5^\circ$ rank 2 to 9, MVRecon F-score rank 1 to 8; DA3-NGL: VideoDepth rank 3 to 12). &
Depth, reconstruction, and poses up to a few hundred images \\
\addlinespace
Long-sequence systems (Pi-Long, LingBot-Map, VGGT-Long, ABot-Recon, LoGeR) &
The only models that run the large SLAM range; lead the short and large \oursdata ranges (Pi-Long: ATE 1.20\,m and 13.1\,m, drift 13.8\% and 27.0\%; ABot-Recon: drift 13.4\% on the short range). &
Frequent memory failures (ZipMap on every large Oxford Spires and VBR sequence, Scal3R on 6 of 7 VBR sequences); incomplete trajectories (VGGT-SLAM: completeness 0.10 on the large \oursdata range); large-range drift above 50\% on \oursdata for all but Pi-Long (27.0\%) and ABot-Recon (48.1\%); low drift and low ATE do not coincide (ABot-Recon, LingBot-Map). &
Long recordings, city-scale paths \\
\addlinespace
VLMs (Gemini 4 Argon, GPT-6 Astra, Claude Opus 5.5, SenseNova-Vision, DepthLM) &
Fastest progress on MonoDepth (31.5 points per year); close to the median model on AerialMetric (Gemini 4 Argon, 6.0\%; median 4.9\%); the commercial VLMs answer metric depth queries in text, without a depth head. &
Least accurate family: on \oursdata, 28.2\% (Gemini 4 Argon) to 45.4\% (Qwen3-VL), and $\delta_{1.03}$ at most 13.9\%; nearly planar depth on a \oursdata frame (Claude Opus 5.5, GPT-6 Astra, AbsRel 0.39--0.40); VideoDepth 27--36\% (\cref{tab:exp:video}) and 46.2--55.1\% on \oursdata; commercial VLMs are queried only at $16{\times}16$ points. &
Coarse metric layout from language queries \\
\bottomrule
\end{tabular}
\end{table*}

\Cref{tab:suppl:strengths} summarizes the strengths and weaknesses of each model family. We draw five conclusions from it.

\emph{(1) Multi-view 3D/4D models are the strongest general-purpose geometric models today, but they are not yet reliable.} They enter all five tasks, lead VideoDepth, MVRecon, and SfM, and are the most accurate single-image depth models on \oursdata, where VGGT-$\Omega$ is also the least sensitive to camera roll. Their weaknesses are long inputs, on which window-wise alignment accumulates drift (42--221\% even on short \oursdata sequences), fine pose accuracy, which their low working resolution and the absence of geometric verification limit, and priors, which they follow even when the priors conflict with the image.

\emph{(2) Monocular specialists remain the most accurate single-image models on upright views of controlled scenes, but they depend on that setting.} They have the lowest AbsRel on every public MonoDepth dataset, yet their error rises by $2.7$--$4.2\times$ on \oursdata and more than doubles under strong camera roll, which indicates a learned gravity prior rather than an understanding of geometry that is independent of orientation.

\emph{(3) Dedicated video depth models have been overtaken.} The best 3D/4D model is more accurate on every VideoDepth dataset, and on \oursdata the video depth models have $3.4$--$4.6\times$ the error of VGGT-$\Omega$; temporal smoothness without multi-view geometry does not resolve the depth of distant structure.

\emph{(4) Long recordings require dedicated long-sequence machinery, and its robustness is the bottleneck.} Only the long-sequence systems run the large range, and they also lead the short \oursdata range, on which the feed-forward models drift. Chunking a strong feed-forward model with overlapping alignment, as Pi-Long does, is currently the most reliable design; the other systems fail through memory exhaustion, incomplete trajectories, or drift above 50\%.

\emph{(5) VLMs recover the coarse metric layout but are not yet geometric models.} They improve fastest on MonoDepth and approach the median model on some public datasets, but on \oursdata their best AbsRel is 28.2\% and their $\delta_{1.03}$ is at most 13.9\%. We query the commercial VLMs at only $16{\times}16$ points (\cref{sec:suppl:vlm}), so their scores bound only their coarse accuracy.

Overall, no family combines breadth, fine accuracy, and robustness to viewpoint, depth range, and input length. The public benchmarks share the upright viewpoints, narrow depth ranges, and short inputs of the training data, and therefore do not expose these weaknesses; \oursdata does.

\subsection{Implications for future work}
\label{sec:suppl:disc_future}

These observations suggest six directions for the design and evaluation of geometric foundation models.
\begin{enumerate}[leftmargin=*,itemsep=1pt,topsep=2pt]
\item \emph{Remove the dependence on upright views.} The error more than doubles under camera roll. Training with in-plane rotation and with views that look up or down, or conditioning on the gravity direction where a device provides it, should close much of this gap; evaluation should include such views.
\item \emph{Cover near and far structure in one image.} All models compress depth beyond 20\,m. Training data in which near and far structure appear together, and losses that weight relative error uniformly over depth, address the failure that a single scale and shift cannot correct.
\item \emph{Verify geometry and raise working resolution.} Spurious jumps on repeated structure and the gap between AUC@$30^\circ$ and AUC@$5^\circ$ call for geometric consistency checks inside or after the network and for pose estimation at higher resolution.
\item \emph{Handle large motion, bound memory, and correct drift.} Training sequences with large camera motion between frames should accompany the small-motion video clips that dominate current training data. Models should run within bounded memory, through streaming, chunking, or a compressed state, and correct accumulated scale drift, \eg, through loop closure or global optimization.
\item \emph{Detect failures and handle priors.} Incomplete trajectories, constant predictions, and far outliers produce plausible-looking outputs; per-frame and per-point confidence is needed to expose them. Prior conditioning should be trained on noisy, sparse, and inconsistently scaled inputs, so that a model given cameras or depth is never less accurate than without them.
\item \emph{Evaluate fine accuracy on uncontrolled data.} $\delta_{1.25}$ and AUC@$30^\circ$ are saturated on most public datasets, whereas $\delta_{1.03}$ and AUC@$5^\circ$ on \oursdata still separate the models, under a common robust alignment on audited test data.
\end{enumerate}

\section{Conclusion}
\label{sec:conclusion}

We presented \ours, an open arena that evaluates geometric foundation models on five tasks, 23 public datasets, and our \oursdata captures under one protocol, with robust and deterministic alignment of depth and trajectories, and with leaked test data excluded from the ranking. Evaluating 52 models shows that single-image depth is close to saturation on public benchmarks but not on large, uncontrolled scenes, that progress now comes from consistency across frames and views, where feed-forward 3D/4D models match or surpass the monocular specialists and overtake dedicated video depth models, and that vision--language models improve fastest on single-image depth but still lack metric precision. No model leads on every dataset, and long recordings remain out of reach for feed-forward models on their own.

\paragraph{Limitations.}
We measure the effect of camera roll by comparing images, not by rotating them, so content that differs between front and side views, such as more sky and facades in the side views, contributes to the gap; the same-scene comparison and the monotonic trend with roll make roll the most likely main factor, but not the only one. The trends rest on one to three years of releases per model family and are indicative rather than forecasts. Several comparisons involve few models (four accept auxiliary inputs, eight run the large SLAM range), some models cover \oursdata only partially (marked $^\circ$), and we evaluate the commercial VLMs only on MonoDepth and through a $16{\times}16$ query grid. We will keep the leaderboard open to new models and datasets.

\section*{LLM Usage Statement}
We used Claude Code with Claude Opus 5.5 as a writing and debugging assistant. For the text, it checked grammar and consistency; the authors revised and verified all text. It assisted with the code of the leaderboard webpage. No LLM was used to design the benchmark, produce ground truth, or compute any metric. Separately, Claude Opus 5.5, GPT-6 Astra, Gemini 4 Argon, and four open vision--language models are \emph{evaluated} as subjects of the benchmark (\suppref{models}); their predictions are scored like those of any other model. The authors take full responsibility for the content of the paper.

{
    \small
    \bibliographystyle{ieeenat_fullname}
    \bibliography{main}
}

\appendixpage
\renewcommand{\topfraction}{0.95}\renewcommand{\textfraction}{0.05}
\renewcommand{\floatpagefraction}{0.35}
\setcounter{topnumber}{4}\setcounter{totalnumber}{6}
\makeatletter
\setlength{\@fptop}{0pt}\setlength{\@fpbot}{0pt plus 1fil}\setlength{\@fpsep}{8pt}
\makeatother
\appendix

In \cref{sec:suppl:data}, we detail the datasets, the protocol of each task, and our data leakage audit. In \cref{sec:suppl:results}, we report the complete quantitative and qualitative results, including the scores on the public datasets alone (\cref{sec:suppl:res_public}). We define the metrics in \cref{sec:suppl:metrics}, and describe our depth and trajectory alignment, the input preprocessing of each model, and our query protocol for the commercial VLMs in \cref{sec:suppl:impl}. As in the main text, every table and trend scores and averages \oursdata like a public dataset unless stated otherwise.

\suppsection{data}

We list every dataset of \ours with its scenes, images, ground truth, and input tracks in \cref{tab:datasets}.

\paragraph{Evaluation environment.}
We run all evaluations on a cluster of NVIDIA H200 GPUs (141\,GB of memory each) with 512\,GB of host RAM per node. Each job runs one method on one GPU, so the memory limits that we report are those of a single H200 and of the host RAM of its node. We run the commercial VLMs through their APIs (\cref{sec:suppl:vlm}).

\begin{table*}[t]
\centering
\footnotesize
\setlength{\tabcolsep}{4pt}
\caption{\textbf{Dataset checklist of \ours.} MonoDepth scores every image. VideoDepth, MVRecon, and SfM cap every scene at 100 evaluated frames (fixed-seed sampling, temporal order preserved), except on \oursdata, where we evaluate the VideoDepth clips whole and cap the MVRecon and SfM scenes at 300 frames. We feed the SLAM sequences whole at a fixed stride; their \#Evaluated lists the frames fed per range track (short\,/\,long, short\,/\,large, or short\,/\,long\,/\,large). For the public datasets, these are the same sequences at different strides; each \oursdata sequence runs in one track only, so its three numbers count disjoint sequences (3\,/\,2\,/\,3). \oursdata (\cref{sec:method:data}) is our own capture. Scene type: I = indoor, O = outdoor, A = aerial, Obj = object-centric; S/D = static/dynamic; R/Syn = real/synthetic. Datasets used by several tasks share one physical copy on disk. $^\dagger$We exclude it from the aggregate ranking because several evaluated methods report training on it (\cref{sec:suppl:leakage}).}
\label{tab:datasets}
\begin{adjustbox}{max width=\textwidth}
\begin{tabular}{llllrrrl}
\toprule
Task & Dataset & Scene type & Ground truth & \#Scenes & \#Images & \#Evaluated & Tracks \\
\midrule
\multirow{7}{*}{MonoDepth}
 & Marigold suite~\cite{DBLP:conf/cvpr/KeOHMDS24}$^{\ddagger}$ & I+O, S, R & depth & 4 & 2,531 & 2,531 & \multirow{7}{*}{\texttt{rgb}} \\
 & HAMMER~\cite{DBLP:conf/cvpr/JungRZBLVSZAILN23} & I+Obj, S, R & depth & 1 & 775 & 775 & \\
 & iBims-1~\cite{DBLP:conf/eccv/KochLF018} & I, S, R & depth & 1 & 100 & 100 & \\
 & Spring~\cite{DBLP:conf/cvpr/MehlSJNB23}$^\dagger$ & O, D, Syn & depth (relative) & 1 & 1,000 & 1,000 & \\
 & Synth4K~\cite{DBLP:conf/cvpr/YuLWLWZWZHP26} & O, D, Syn & depth & 5 & 1,960 & 1,960 & \\
 & AerialMetric~\cite{DBLP:conf/eccv/SongCZZFYHCC26} & A, D, R & depth & 3 & 3,596 & 3,596 & \\
 & \oursdata (ours) & I+O, S, R & depth & 19 & 1,688 & 1,688 & \\
\midrule
\multirow{5}{*}{VideoDepth}
 & Sintel~\cite{DBLP:conf/eccv/ButlerWSB12} & I+O, D, Syn & depth & 23 & 1,064 & 1,064 & \multirow{5}{*}{\texttt{rgb}, \texttt{rgbc}} \\
 & Bonn~\cite{DBLP:conf/iros/PalazzoloBLGS19} & I, D, R & depth & 26 & 2,860 & 2,600 & \\
 & KITTI~\cite{DBLP:conf/cvpr/GeigerLU12} & O, D, R & depth & 13 & 1,269 & 1,189 & \\
 & DyCheck~\cite{DBLP:conf/nips/GaoLTRK22} & I+O, D, R & depth & 12 & 3,952 & 1,200 & \\
 & \oursdata (ours) & I+O, S, R & depth & 15 & 4,510 & 4,484 & \\
\midrule
\multirow{8}{*}{MVRecon}
 & 7-Scenes~\cite{DBLP:conf/cvpr/ShottonGZICF13} & I, S, R & depth, pose, mesh & 7 & 43,000 & 700 & \texttt{rgb}, \texttt{rgbc}, \texttt{rgbd} \\
 & ETH3D~\cite{DBLP:conf/cvpr/SchopsSGSSPG17} & I+O, S, R & depth, pose, scan & 11 & 908 & 856 & \texttt{rgb}, \texttt{rgbc}, \texttt{rgbd} \\
 & DTU~\cite{DBLP:journals/ijcv/AanaesJVTD16} & Obj, S, R & depth, pose, scan & 15 & 855 & 855 & \texttt{rgb}, \texttt{rgbc}, \texttt{rgbd} \\
 & Tanks\&Temples~\cite{DBLP:journals/tog/KnapitschPZK17} & I+O+Obj, S, R & pose, scan & 6 & 2,784 & 600 & \multirow{5}{*}{\texttt{rgb}, \texttt{rgbc}} \\
 & ScanNet++~\cite{DBLP:conf/iccv/YeshwanthLND23}$^\dagger$ & I, S, R & depth, pose, mesh & 20 & 178,345 & 2,000 & \\
 & HiRoom~\cite{DBLP:conf/iclr/LinCLCLZPGZSFK26} & I, S, Syn & depth, pose, points & 30 & 400 & 400 & \\
 & NRGBD~\cite{DBLP:conf/cvpr/AzinovicMGNT22} & I, S, Syn & depth, pose, mesh & 9 & 10,970 & 900 & \\
 & \oursdata (ours) & I+O, S, R & depth, pose, LiDAR & 6 & 5,295 & 1,800 & \\
\midrule
\multirow{8}{*}{SfM}
 & ETH3D~\cite{DBLP:conf/cvpr/SchopsSGSSPG17} & I+O, S, R & pose & 13 & 908 & 856 & \multirow{8}{*}{\texttt{rgb}, \texttt{rgbd}} \\
 & Tanks\&Temples~\cite{DBLP:journals/tog/KnapitschPZK17} & I+O+Obj, S, R & pose & 6 & 2,784 & 600 & \\
 & IMC Phototourism~\cite{DBLP:journals/ijcv/JinMMMFYT21} & O, S, R & pose & 8 & 3,145 & 800 & \\
 & DTU~\cite{DBLP:journals/ijcv/AanaesJVTD16} & Obj, S, R & pose & 15 & 855 & 855 & \\
 & ScanNet++~\cite{DBLP:conf/iccv/YeshwanthLND23}$^\dagger$ & I, S, R & pose & 20 & 178,345 & 2,000 & \\
 & HiRoom~\cite{DBLP:conf/iclr/LinCLCLZPGZSFK26} & I, S, Syn & pose & 30 & 400 & 400 & \\
 & TexturePoorSfM~\cite{DBLP:conf/cvpr/HeSWPHBZ24} & Obj, S, R & pose & 30 & 350 & 350 & \\
 & \oursdata (ours) & I+O, D, R & pose & 7 & 7,680 & 2,100 & \\
\midrule
\multirow{6}{*}{SLAM}
 & 7-Scenes~\cite{DBLP:conf/cvpr/ShottonGZICF13} & I, S, R & pose & 18 & 17,000 & 1,700\,/\,5,678 & \multirow{3}{*}{\texttt{rgb}, \texttt{rgbd}} \\
 & TUM RGB-D (dynamic)~\cite{DBLP:conf/iros/SturmEEBC12} & I, D, R & pose & 9 & 10,806 & 1,085\,/\,3,606 & \\
 & \oursdata (ours) & I+O, D, R & pose & 8 & 15,821 & 918\,/\,2,149\,/\,10,919 & \\
 & Oxford Spires~\cite{DBLP:journals/ijrr/TaoMZWFF26} & O, S, R & pose & 10 & 38,380 & 3,199\,/\,38,380 & \multirow{3}{*}{\texttt{rgb}} \\
 & VBR~\cite{DBLP:conf/icra/BriziGGFSRG24} & O, D, R & pose & 7 & 86,678 & --\,/\,86,678 & \\
 & KITTI Odometry~\cite{DBLP:conf/cvpr/GeigerLU12} & O, D, R & pose & 11 & 23,201 & --\,/\,23,201 & \\
\midrule
\multicolumn{8}{l}{Total: 24 unique datasets (23 public and \oursdata), 34 task--dataset pairs, 456 evaluation cases, 451,584 unique images.} \\
\bottomrule
\end{tabular}
\end{adjustbox}\\[2pt]
\raggedright\scriptsize $^{\ddagger}$NYUv2~\cite{DBLP:conf/eccv/SilbermanHKF12}, KITTI~\cite{DBLP:conf/cvpr/GeigerLU12}, ETH3D~\cite{DBLP:conf/cvpr/SchopsSGSSPG17}, and DIODE~\cite{DBLP:journals/corr/abs-1908-00463}, as packaged by Marigold; each subset is one scene. MVRecon ETH3D omits the two scenes without a released reference surface. SLAM 7-Scenes uses the 18 official test sequences, which are part of the MVRecon images. A SLAM case is one sequence in one range track. Each \oursdata SLAM sequence runs only in the range track it was selected for, and \texttt{rgbd} runs only on its three static or lightly populated sequences. \oursdata MonoDepth images are drawn from all 19 depth scenes, including the VideoDepth and MVRecon ones. Totals count shared images once.
\end{table*}

\paragraph{Frame selection and input resolution.}
To give all feed-forward models identical inputs, we cap each scene of VideoDepth, MVRecon, and SfM at 100 frames, following DA3-BENCH~\cite{DBLP:conf/iclr/LinCLCLZPGZSFK26}; for \oursdata, we feed the VideoDepth clips whole and cap the MVRecon and SfM scenes at 300 frames. For larger scenes, we draw a fixed-seed random subset in temporal order and process it in one forward pass, so that all predictions of a scene share one coordinate frame.
We run each method at its own native input resolution and map its outputs back to the source image, so we always compute the metrics at ground-truth resolution. We list the resolution, resizing, padding, and output masking of every model in every task in \cref{sec:suppl:preprocess}.

\paragraph{MonoDepth.}
We take NYUv2, KITTI (Eigen split), ETH3D, and DIODE from the Marigold suite~\cite{DBLP:conf/cvpr/KeOHMDS24}. Following the zero-shot suite of MoGe~\cite{DBLP:conf/cvpr/WangXDX00Y25}, we add HAMMER (indoor tabletop scenes with high-quality depth), iBims-1 (indoor scans with dense laser depth), and Spring (rendered film scenes with relative depth). We further add Synth4K ($3840{\times}2160$ frames from five video games), AerialMetric (UAV viewpoints in three scenes), and 1,688 \oursdata images from 19 indoor and outdoor scenes. We treat each test set of the Marigold suite as one scene, and we use the frames and validity masks of MoGe's zero-shot suite~\cite{DBLP:conf/cvpr/WangXDX00Y25} for HAMMER, iBims-1, and Spring. Since Synth4K publishes no evaluation list, we keep half of the frames of each game with ground truth in $(3,150)$\,m, spaced uniformly. AerialMetric covers controlled pitch, altitude, and field-of-view sweeps, oblique views, and online videos; we sample it in the same way and mask out moving objects.

\paragraph{VideoDepth.}
Sintel has 23 synthetic, highly dynamic sequences; Bonn 26 indoor RGB-D sequences with moving people and objects; KITTI 13 driving sequences with projected LiDAR depth; DyCheck 12 casual handheld phone captures of people, pets, and objects in motion; and \oursdata 15 single-camera clips of 132 to 1,021 frames from seven walks. For DyCheck, we use the first camera of the evaluation release distributed by MegaSaM~\cite{DBLP:conf/cvpr/Li0C0JYKHS25}, with its depth maps at their native resolution. We pair the Bonn depth maps with RGB frames by nearest timestamp (within $0.25$\,s). In \oursdata, 26 frames failed our manual review of the ground truth; we keep them as model inputs but do not score them.

\paragraph{MVRecon.}
For ScanNet++ and HiRoom, we follow the scene lists of DA3-BENCH~\cite{DBLP:conf/iclr/LinCLCLZPGZSFK26}. We run the \texttt{rgbd} track only where the ground-truth depth is sparse or incomplete (7-Scenes, ETH3D, and DTU), so that the prior does not already contain the reference surface. Two ETH3D scenes (\emph{meadow} and \emph{terrace}) have no released reference surface; we leave them out, which leaves 11 ETH3D scenes. For each \oursdata scene, we feed a method a fixed-seed subset of 300 of its 363 to 1,806 images and use the LiDAR map as the reference, cropped to the points that some view sees in front of its dense depth, so that completeness does not penalize unobserved regions. Because \oursdata has dense ground-truth depth, we exclude it from the \texttt{rgbd} track.

\paragraph{SfM.}
The \oursdata scenes include crowded landmarks, a repetitively tiled underground platform, and fast turns. Each method exports a COLMAP sparse model, and we match predicted and ground-truth images by filename. For the \texttt{rgbd} track of \oursdata, we use the projected LiDAR depth as the prior.

\begin{table*}[t]
\centering
\scriptsize
\setlength{\tabcolsep}{4pt}
\caption{\textbf{Training-data audit.} For each evaluated dataset, the methods whose paper or released training code uses it, and our decision. ``Excluded'' means the dataset stays in every per-dataset table but is left out of the aggregate ranking. Methods not listed in a row disclose no use of that dataset, which does not prove that they never trained on it.}
\label{tab:suppl:leakage}
\begin{tabular}{p{0.15\textwidth}p{0.55\textwidth}p{0.22\textwidth}}
\toprule
Dataset & Evidence & Decision \\
\midrule
ScanNet++ & Trained on without a split shown to be disjoint from our 20 test scenes: $\pi^3$-X, OmniVGGT, DepthLM (about 10 held out), SenseNova-Vision, DepthVLM, Fast3R, IGGT, MapAnything and D\'ej\`a View (own held-out scenes, not matched to ours), Pow3R and CUT3R (DUSt3R's training pairs), LoGeR and Scal3R (training split not stated). Clean split: DA3 (230 official training scenes, disjoint test set), DUSt3R and MASt3R (official training split in code). Not used: VGGT and its variants, Co-Me. & Excluded \\
Spring & Our frames come from the training split (the test ground truth is withheld), which DA3, MapAnything, CUT3R, OmniVGGT, D\'ej\`a View, and LoGeR train on. & Excluded \\
TexturePoorSfM & Bags are re-samplings of 17 object captures; a split by bag puts every test frame into the training pool. & Split by object; 30 test bags from 5 held-out objects \\
KITTI Odometry & No released recipe trains on it. Several methods train on Virtual KITTI~2 (synthetic re-renders of five KITTI scenes), and R3 on KITTI-360 (other recordings of the same platform): a domain overlap only. & Kept \\
AerialMetric & Only the test split is used; its training split trains MoGe2-Aerial, which we do not evaluate. D\'ej\`a View and Scal3R train on MatrixCity aerial renders (domain overlap only). & Kept \\
Other datasets & No use found for NYUv2, KITTI (Eigen), ETH3D, DIODE, HAMMER, iBims-1, Synth4K, Sintel, Bonn, DyCheck, 7-Scenes, TUM RGB-D, Oxford Spires, VBR, NRGBD, DTU, Tanks\&Temples, IMC, and HiRoom (DA3's own synthetic test set). The training data of Qwen3-VL and of the three commercial VLMs (proprietary corpora) cannot be verified. & Kept \\
\oursdata & Not released before this work. & Kept \\
\bottomrule
\end{tabular}
\end{table*}
\paragraph{SLAM.}
We use the 18 official test sequences of 7-Scenes (handheld RGB-D in static rooms), the nine \emph{Dynamic Objects} sequences of TUM RGB-D (people sitting, moving, and walking through the view), ten Oxford Spires sequences (outdoor college and landmark sites), seven VBR sequences (large-scale urban recordings in Rome), the eleven KITTI Odometry sequences with public ground truth, and eight \oursdata walks of 794 to 4,145 frames (indoor and outdoor, up to 1\,km long, several of them crowded). For the \emph{short} range track, we use stride 10 on 7-Scenes and TUM and stride 12 on Oxford Spires (inputs of up to 364 frames); for \emph{long}, stride 3 on 7-Scenes and TUM (up to 1,211 frames); and for \emph{large}, every frame of Oxford Spires, VBR, and KITTI Odometry (up to 18,846 frames). We run each \oursdata sequence in the one track for which we selected it.
We associate the TUM RGB-D ground truth with the frames within $0.02$\,s. For Oxford Spires, we use the first 3,840 frames of one camera, rectified from fisheye. For KITTI Odometry, we use sequences 00--10 and the left color camera, and transfer the GPS/INS poses to it through its calibration. For \oursdata, we use one camera at 5\,Hz and feed it at stride 3 (1.7\,Hz) in the short track and at stride 1 in the others. As the \texttt{rgbd} prior, we use the sensor depth of 7-Scenes and TUM and the projected LiDAR depth of the three static or lightly populated \oursdata sequences.
A method enters a range track if its measured single-pass memory budget on one 141\,GB GPU covers the longest input of the track, so we also evaluate a method of a longer track on the shorter ones. We still run the methods whose budget is below the longest short-range input (MapAnything, MASt3R, D\'ej\`a View, DUSt3R, Pow3R, IGGT4D) on every sequence of the short and long tracks, GenCeption (pose), which accepts at most 81 frames, on every sequence of the short track, and LoGeR, R3, ZipMap, and Scal3R on every sequence of the large track. A sequence that fails (out of memory or too long) scores 0, and we record its cause; we mark such methods with $^\circ$ in \cref{tab:exp:slam}. We run long-sequence systems that chunk their input internally (\eg, VGGT-SLAM~\cite{DBLP:conf/nips/MaggioLC25}) with their published settings in all tracks. All methods receive the same input frames (\cref{sec:suppl:preprocess}).

\paragraph{\oursdata pipeline.}
We refine the poses with five rounds of global bundle adjustment with track completion, track merging, and outlier filtering~\cite{DBLP:conf/cvpr/SchonbergerF16}; all cameras of a frame share one rig pose, and each camera keeps its fixed rig-to-camera transform. The final $\mathrm{Sim}(3)$ fit to the scanner poses corrects the scale by $0.99$--$1.16$ and leaves a median camera difference of $2.5$\,cm. For LiDAR depth, we keep the nearest point per pixel without a depth cutoff and mark pixels without a return as invalid. We prompt SAM~3.1 separately with \emph{person}, \emph{vehicle}, \emph{license plate}, and \emph{sky}, and dilate every mask by 8 pixels; we never modify the source images. We initialize the Gaussian Splatting model from the refined sparse points and train it for 30k iterations with LiDAR depth supervision (log-$\ell_1$, weight $0.05$), depth--normal consistency (weight $0.05$), and multi-view geometric consistency (weight $1$). We mark rendered non-finite, non-positive, and sky pixels as invalid.

\paragraph{\oursdata selection.}
We rate every scene by the share of pixels masked as people or vehicles: \emph{static} (below 1\%), \emph{light} (1--3\%), or \emph{crowded} (above 3\%, or people in over 60\% of frames). For the depth tasks, we use the 19 static or lightly populated scenes; for SfM and RGB-only SLAM, which we score on poses only, we also admit crowded scenes. For MonoDepth, we draw images from all three cameras of the 19 depth scenes: among the images whose render PSNR exceeds the scene median, we sample 10\% of the scene's images, spread over time. For VideoDepth, we split seven walks into consecutive, non-overlapping clips with lengths drawn uniformly from 100--500 frames, each on one randomly chosen camera (fixed seed), and keep one 1,021-frame recording whole. For MVRecon, we use six compact scenes (extent at most 36\,m) with every image of the three cameras (5,295 in total). For SfM, we use every image of the three cameras of seven scenes (321 to 1,947 images), from which we draw a fixed-seed subset of up to 300. For SLAM, we use one camera per sequence and size each sequence close to the longest input of its range track. In the depth tasks, we black out people and vehicles in the images and remove them from the ground truth; in SfM and SLAM, we keep the images unmodified. We checked every frame of MonoDepth and VideoDepth by eye next to its LiDAR and dense depth, and every MVRecon, SfM, and SLAM scene in a 3D viewer. This review removed 203 MonoDepth images and the ground truth of 26 VideoDepth frames. In total, \oursdata comprises 32,655 unique images (34,994 over all tasks, since we sample the MonoDepth images from the VideoDepth and MVRecon scenes).

\paragraph{Data leakage audit.}
\label{sec:suppl:leakage}
A benchmark rewards memorization when its test scenes are part of a model's training data. For every evaluated method, we searched its paper's list of training data and, where released, its training configurations and data loaders for each evaluated dataset (\cref{tab:suppl:leakage}). We exclude a dataset from the aggregate ranking when several evaluated methods train on its test scenes, or on scenes not shown to be disjoint from them; we still keep it in every per-dataset table as a diagnostic. We record overlap in domain, such as synthetic re-renders of the same streets, but do not exclude a dataset for it.

We exclude two public datasets. Most methods disclose training on ScanNet++~\cite{DBLP:conf/iccv/YeshwanthLND23} without a split shown to be disjoint from our test scenes, and six methods train on the Spring~\cite{DBLP:conf/cvpr/MehlSJNB23} split that we evaluate. TexturePoorSfM~\cite{DBLP:conf/cvpr/HeSWPHBZ24} leaks within itself: its 1,020 bags are re-samplings of 17 object captures, and bags of one object share identical frames and poses, so a split by bag puts test frames into training. We therefore split by object: 12 objects for training and five held-out objects that supply all 30 test bags; none of the training images is identical to a test image. For the other public datasets, we found no evidence of training overlap, although this does not prove that there is none. Since we have not released \oursdata before, no evaluated method can have been trained on it.

\suppsection{results}

\subsection{Quantitative results}
\label{sec:suppl:res_quant}

We report the headline tables in the format of \cref{tab:exp:mono} and give every metric of \cref{sec:suppl:metrics} in the all-metrics tables.

\begin{table*}[t]
\centering
\scriptsize
\setlength{\tabcolsep}{3pt}
\caption{\textbf{MonoDepth, all metrics, \texttt{rgb} track}, averaged over all datasets. AbsRel and $\delta$ in \%, normal error in degrees. Colors and $^\circ$ as in \cref{tab:exp:mono}.}
\label{tab:suppl:res_mono}
\adjustbox{max width=\linewidth}{
\begin{tabular}{@{}lrrrrrrrrrrrrr@{}}
\toprule
Method & Score$\uparrow$ & AbsRel$\downarrow$ & $\delta_{1.03}$$\uparrow$ & $\delta_{1.25}$$\uparrow$ & $\delta_{1.25^2}$$\uparrow$ & $\delta_{1.25^3}$$\uparrow$ & SqRel$\downarrow$ & RMSE$\downarrow$ & RMSE$_{\log}$$\downarrow$ & log10$\downarrow$ & iRMSE$\downarrow$ & SILog$\downarrow$ & Normal$\downarrow$ \\
\midrule
VGGT-$\Omega$ & \rkA{96.7} & \rkB{10.76} & \rkA{49.85} & \rkB{89.79} & \rkB{95.63} & \rkB{97.58} & 1.041 & 3.515 & \rkB{0.176} & \rkB{0.042} & 0.101 & \rkB{17.37} & 26.49 \\
MoGe-3 & \rkB{92.1} & 11.57 & \rkC{46.21} & 87.34 & 93.89 & 96.44 & \rkA{0.752} & \rkB{3.377} & 0.209 & 0.052 & 0.202 & 20.40 & \rkA{24.39} \\
$\pi^3$-X & \rkC{91.4} & \rkC{11.38} & 45.02 & 87.45 & 94.10 & 96.54 & \rkC{0.799} & 3.872 & 0.216 & 0.053 & 0.447 & 21.11 & 26.16 \\
AMB3R & 91.3\rkn{4} & 11.93 & 44.32 & 87.43 & 94.30 & 96.66 & 0.980 & 3.755 & 0.204 & 0.051 & 1.935 & 19.95 & \rkC{25.68} \\
UniDepthV2 & 91.0\rkn{5} & 12.73 & 45.97 & \rkC{88.26} & \rkC{94.51} & \rkC{96.84} & 1.143 & \rkA{3.352} & \rkC{0.195} & \rkC{0.049} & 0.180 & \rkC{19.12} & 28.07 \\
MoGe & 90.7\rkn{6} & 12.31 & 45.19 & 86.66 & 93.51 & 96.20 & 0.827 & 3.617 & 0.220 & 0.055 & 0.584 & 21.57 & \rkB{24.43} \\
MoGe-2 & 90.5\rkn{7} & 12.01 & 45.30 & 86.93 & 93.73 & 96.28 & \rkB{0.762} & \rkC{3.379} & 0.211 & 0.053 & 0.511 & 20.61 & 25.69 \\
\bottomrule
\end{tabular}}
\end{table*}

\paragraph{MonoDepth.}
VGGT-$\Omega$ leads (\cref{tab:exp:mono,tab:suppl:res_mono}), with the lowest AbsRel of the models scored on every dataset (10.76\%) and the highest $\delta_{1.03}$ (49.85\%), followed by MoGe-3 (11.57\%), the best depth specialist. The commercial VLMs score between 41 and 57, below all depth specialists except Marigold v1.1 (41.9).

\begin{table*}[t]
\centering
\scriptsize
\setlength{\tabcolsep}{3pt}
\caption{\textbf{Headline metrics of VideoDepth, MVRecon, and SfM, \texttt{rgb} track}, averaged over all datasets. (a) AbsRel and $\delta$ in \%. (b) Acc, Comp, and Overall in m, F-score at the threshold of each dataset. (c) RRA, RTA, mAA, and AUC at $30^\circ$, and the ATE after $\mathrm{Sim}(3)$ alignment. Score, colors, and $^\circ$ as in \cref{tab:exp:mono}.}
\begin{subtable}[t]{0.255\textwidth}
\centering
\caption{VideoDepth}\label{tab:exp:video}
\adjustbox{max width=\linewidth}{
\begin{tabular}{@{}lrrrr@{}}
\toprule
Method & Score$\uparrow$ & AbsRel$\downarrow$ & $\delta_{1.03}$$\uparrow$ & $\delta_{1.25}$$\uparrow$ \\
\midrule
VGGT-$\Omega$ & \rkA{96.5} & \rkA{6.49} & \rkA{55.25} & \rkA{94.67} \\
$\pi^3$-X & \rkB{91.9} & \rkB{7.98} & \rkB{52.55} & \rkB{90.97} \\
D\'ej\`a View & \rkC{86.7} & \rkC{9.73} & \rkC{45.09} & \rkC{89.13} \\
DA3-NGL$^\circ$ & 86.3\rkn{4} & 10.36 & 48.43 & 88.64 \\
DA3-G$^\circ$ & 85.6\rkn{5} & 10.30 & 47.68 & 88.34 \\
IGGT4D$^\circ$ & 84.1\rkn{6} & 11.98 & 47.45 & 87.52 \\
DA3-L & 82.2\rkn{7} & 12.12 & 43.46 & 85.84 \\
GenCeption-Depth & 77.1\rkn{8} & 14.04 & 29.65 & 83.98 \\
OmniVGGT$^\circ$ & 75.3\rkn{9} & 13.82 & 42.78 & 84.51 \\
FastVGGT$^\circ$ & 74.6\rkn{10} & 14.35 & 38.67 & 83.86 \\
RobustVGGT$^\circ$ & 74.6\rkn{10} & 14.35 & 38.67 & 83.86 \\
VGGT$^\circ$ & 74.6\rkn{10} & 14.35 & 38.67 & 83.86 \\
DVD & 74.3\rkn{13} & 15.39 & 31.24 & 81.76 \\
AMB3R$^\circ$ & 72.0\rkn{14} & 13.66 & 37.73 & 83.04 \\
IGGT$^\circ$ & 71.8\rkn{15} & 12.48 & 34.51 & 86.25 \\
DepthCrafter & 65.2\rkn{16} & 20.58 & 21.63 & 78.42 \\
MapAnything$^\circ$ & 59.5\rkn{17} & 22.38 & 34.53 & 73.14 \\
CUT3R & 58.3\rkn{18} & 24.92 & 24.63 & 73.08 \\
ChronoDepth & 56.0\rkn{19} & 23.18 & 18.79 & 71.61 \\
Fast3R & 51.9\rkn{20} & 26.91 & 17.73 & 69.55 \\
Co-Me & 50.7\rkn{21} & 25.46 & 20.85 & 70.70 \\
MASt3R$^\circ$ & 50.5\rkn{22} & 24.72 & 16.95 & 74.45 \\
SenseNova-Vision & 47.4\rkn{23} & 27.10 & 16.07 & 66.42 \\
DepthLM & 34.7\rkn{24} & 30.32 & 15.68 & 63.25 \\
DepthVLM & 26.7\rkn{25} & 35.36 & 8.44 & 55.64 \\
Qwen3-VL & 20.9\rkn{26} & 36.37 & 8.53 & 52.39 \\
DUSt3R$^\circ$ & 8.9\rkn{27} & 45.21 & 11.25 & 53.79 \\
Pow3R$^\circ$ & 8.8\rkn{28} & 44.88 & 10.46 & 54.83 \\
\bottomrule
\end{tabular}}
\end{subtable}\hfill
\begin{subtable}[t]{0.335\textwidth}
\centering
\caption{MVRecon}\label{tab:exp:mvrecon}
\adjustbox{max width=\linewidth}{
\begin{tabular}{@{}lrrrrr@{}}
\toprule
Method & Score$\uparrow$ & Acc$\downarrow$ & Comp$\downarrow$ & Overall$\downarrow$ & F-score$\uparrow$ \\
\midrule
DA3-G & \rkA{89.8} & \rkA{0.7276} & \rkA{0.8013} & \rkA{0.7645} & \rkA{0.630} \\
DA3-NGL$^\circ$ & \rkB{88.1} & 0.9317 & 0.7099 & 0.8208 & 0.642 \\
$\pi^3$-X$^\circ$ & \rkC{87.7} & 0.8216 & 0.7871 & 0.8043 & 0.621 \\
VGGT-$\Omega$ & 83.8\rkn{4} & \rkC{1.2077} & \rkC{1.2118} & \rkC{1.2098} & \rkB{0.614} \\
DA3-L$^\circ$ & 82.9\rkn{5} & 0.8297 & 0.7879 & 0.8088 & 0.561 \\
IGGT4D$^\circ$ & 82.4\rkn{6} & 1.1453 & 1.0132 & 1.0793 & 0.606 \\
D\'ej\`a View$^\circ$ & 81.9\rkn{7} & 0.6410 & 0.7248 & 0.6829 & 0.635 \\
OmniVGGT & 74.4\rkn{8} & \rkB{0.9450} & \rkB{0.9606} & \rkB{0.9528} & 0.425 \\
AMB3R$^\circ$ & 70.9\rkn{9} & 1.0238 & 0.8585 & 0.9411 & 0.375 \\
FastVGGT & 70.5\rkn{10} & 1.2902 & 1.4918 & 1.3910 & \rkC{0.461} \\
Co-Me & 64.4\rkn{11} & 1.4029 & 1.6010 & 1.5019 & 0.385 \\
IGGT & 64.3\rkn{12} & 1.5134 & 1.5502 & 1.5318 & 0.400 \\
MapAnything & 61.5\rkn{13} & 9.9175 & 1.3965 & 5.6570 & 0.292 \\
RobustVGGT & 61.0\rkn{14} & 1.4505 & 1.6077 & 1.5291 & 0.351 \\
VGGT & 60.7\rkn{15} & 1.5302 & 1.6239 & 1.5771 & 0.352 \\
CUT3R & 52.5\rkn{16} & 1.8327 & 2.3505 & 2.0916 & 0.289 \\
Pow3R$^\circ$ & 39.9\rkn{17} & 0.7858 & 0.9732 & 0.8795 & 0.500 \\
DUSt3R$^\circ$ & 39.3\rkn{18} & 0.7131 & 0.8176 & 0.7654 & 0.455 \\
MASt3R$^\circ$ & 35.3\rkn{19} & 0.9791 & 2.3037 & 1.6414 & 0.245 \\
Fast3R$^\circ$ & 8.6\rkn{20} & 3.6236 & 2.9091 & 3.2663 & 0.020 \\
\bottomrule
\end{tabular}}
\end{subtable}\hfill
\begin{subtable}[t]{0.385\textwidth}
\centering
\caption{SfM}\label{tab:exp:sfm}
\adjustbox{max width=\linewidth}{
\begin{tabular}{@{}lrrrrrr@{}}
\toprule
Method & Score$\uparrow$ & ATE$\downarrow$ & RRA@30$\uparrow$ & RTA@30$\uparrow$ & mAA@30$\uparrow$ & AUC@30$\uparrow$ \\
\midrule
VGGT-$\Omega$ & \rkA{94.2} & \rkC{0.854} & \rkA{0.995} & \rkA{0.983} & \rkA{0.885} & \rkA{0.873} \\
D\'ej\`a View & \rkB{93.3} & 1.071 & \rkB{0.994} & \rkB{0.979} & \rkB{0.874} & \rkB{0.861} \\
DA3-NGL & \rkC{90.9} & \rkA{0.807} & 0.976 & 0.962 & \rkC{0.848} & \rkC{0.837} \\
DA3-G & 90.6\rkn{4} & \rkB{0.810} & 0.971 & 0.958 & 0.844 & 0.832 \\
RobustVGGT & 88.4\rkn{5} & 1.101 & 0.986 & 0.955 & 0.828 & 0.815 \\
VGGT & 88.4\rkn{5} & 1.101 & 0.986 & 0.955 & 0.828 & 0.815 \\
IGGT4D & 88.4\rkn{7} & 1.271 & 0.970 & 0.954 & 0.828 & 0.815 \\
FastVGGT & 88.3\rkn{8} & 1.099 & 0.985 & 0.955 & 0.826 & 0.814 \\
$\pi^3$-X & 87.9\rkn{9} & 1.276 & \rkC{0.992} & \rkC{0.964} & 0.828 & 0.816 \\
DA3-L & 84.8\rkn{10} & 1.340 & 0.957 & 0.934 & 0.801 & 0.789 \\
Co-Me & 80.7\rkn{11} & 1.344 & 0.957 & 0.921 & 0.751 & 0.740 \\
IGGT & 80.4\rkn{12} & 3.823 & 0.987 & 0.939 & 0.752 & 0.740 \\
OmniVGGT & 80.1\rkn{13} & 2.088 & 0.976 & 0.939 & 0.749 & 0.739 \\
MASt3R$^\circ$ & 77.8\rkn{14} & 2.762 & 0.970 & 0.940 & 0.770 & 0.760 \\
MapAnything & 72.9\rkn{15} & 6.882 & 0.959 & 0.898 & 0.668 & 0.659 \\
CUT3R & 64.8\rkn{16} & 5.835 & 0.897 & 0.824 & 0.575 & 0.568 \\
Pow3R$^\circ$ & 59.6\rkn{17} & 8.463 & 0.988 & 0.894 & 0.685 & 0.678 \\
DUSt3R$^\circ$ & 59.5\rkn{18} & 9.273 & 0.990 & 0.895 & 0.679 & 0.672 \\
Fast3R & 58.8\rkn{19} & 8.216 & 0.847 & 0.750 & 0.526 & 0.519 \\
AMB3R$^\circ$ & 49.6\rkn{20} & 1.043 & 0.580 & 0.676 & 0.433 & 0.427 \\
GenCeption-Pose & 4.6\rkn{21} & 42.702 & 0.376 & 0.202 & 0.027 & 0.023 \\
\bottomrule
\end{tabular}}
\end{subtable}
\end{table*}

\begin{table*}[t]
\centering
\scriptsize
\setlength{\tabcolsep}{3pt}
\caption{\textbf{VideoDepth, all metrics, \texttt{rgb} track}, averaged over all datasets. Units as in \cref{tab:suppl:res_mono}.}
\label{tab:suppl:res_video}
\adjustbox{max width=\linewidth}{
\begin{tabular}{@{}lrrrrrrrrrrrrr@{}}
\toprule
Method & Score$\uparrow$ & AbsRel$\downarrow$ & $\delta_{1.03}$$\uparrow$ & $\delta_{1.25}$$\uparrow$ & $\delta_{1.25^2}$$\uparrow$ & $\delta_{1.25^3}$$\uparrow$ & SqRel$\downarrow$ & RMSE$\downarrow$ & RMSE$_{\log}$$\downarrow$ & log10$\downarrow$ & iRMSE$\downarrow$ & SILog$\downarrow$ & Normal$\downarrow$ \\
\midrule
VGGT-$\Omega$ & \rkA{96.5} & \rkA{6.49} & \rkA{55.25} & \rkA{94.67} & \rkA{98.08} & \rkA{98.98} & \rkA{0.269} & \rkA{1.600} & \rkA{0.126} & \rkA{0.028} & \rkA{0.102} & \rkA{12.52} & -- \\
$\pi^3$-X & \rkB{91.9} & \rkB{7.98} & \rkB{52.55} & \rkB{90.97} & \rkB{96.22} & \rkB{97.93} & \rkB{0.311} & \rkB{1.754} & \rkB{0.158} & \rkB{0.037} & 900.613 & \rkB{15.58} & -- \\
D\'ej\`a View & \rkC{86.7} & \rkC{9.73} & \rkC{45.09} & \rkC{89.13} & \rkC{95.24} & 97.47 & \rkC{0.454} & 2.122 & 0.222 & \rkC{0.052} & 3953.443 & 21.51 & -- \\
DA3-NGL$^\circ$ & 86.3\rkn{4} & 10.36 & 48.43 & 88.64 & 94.79 & 97.27 & 0.494 & 1.958 & 0.168 & 0.045 & 6.494 & 16.68 & -- \\
DA3-G$^\circ$ & 85.6\rkn{5} & 10.30 & 47.68 & 88.34 & 94.70 & 97.04 & 0.442 & 1.994 & 0.171 & 0.045 & 0.130 & 16.92 & -- \\
IGGT4D$^\circ$ & 84.1\rkn{6} & 11.98 & 47.45 & 87.52 & 94.61 & 96.81 & 0.498 & 2.069 & 0.235 & 0.057 & 5354.446 & 22.92 & -- \\
DA3-L & 82.2\rkn{7} & 12.12 & 43.46 & 85.84 & 93.76 & 96.84 & 0.557 & \rkC{2.087} & 0.217 & 0.055 & 2681.432 & 21.33 & -- \\
\bottomrule
\end{tabular}}
\end{table*}

\paragraph{VideoDepth.}
Feed-forward 3D/4D models occupy the first seven places (\cref{tab:exp:video,tab:suppl:res_video}). VGGT-$\Omega$ attains an AbsRel of 6.49\%, followed by $\pi^3$-X (7.98\%) and D\'ej\`a View (9.73\%); the best video depth model, GenCeption-Depth, ranks eighth (14.04\%), and DVD, eighth without \oursdata, falls to 13th (15.39\%).

\begin{table*}[t]
\centering
\scriptsize
\setlength{\tabcolsep}{4pt}
\caption{\textbf{MVRecon, all metrics, \texttt{rgb} track}, averaged over all datasets. Acc, Comp, and Overall in m.}
\label{tab:suppl:res_mvrecon}
\begin{tabular}{@{}lrrrrrrr@{}}
\toprule
Method & Score$\uparrow$ & Acc$\downarrow$ & Comp$\downarrow$ & Overall$\downarrow$ & Prec$\uparrow$ & Rec$\uparrow$ & F-score$\uparrow$ \\
\midrule
DA3-G & \rkA{89.8} & \rkA{0.7276} & \rkA{0.8013} & \rkA{0.7645} & \rkA{0.670} & \rkB{0.649} & \rkA{0.630} \\
DA3-NGL$^\circ$ & \rkB{88.1} & 0.9317 & 0.7099 & 0.8208 & 0.684 & 0.656 & 0.642 \\
$\pi^3$-X$^\circ$ & \rkC{87.7} & 0.8216 & 0.7871 & 0.8043 & 0.667 & 0.632 & 0.621 \\
VGGT-$\Omega$ & 83.8\rkn{4} & \rkC{1.2077} & \rkC{1.2118} & \rkC{1.2098} & \rkB{0.639} & \rkA{0.662} & \rkB{0.614} \\
DA3-L$^\circ$ & 82.9\rkn{5} & 0.8297 & 0.7879 & 0.8088 & 0.599 & 0.581 & 0.561 \\
IGGT4D$^\circ$ & 82.4\rkn{6} & 1.1453 & 1.0132 & 1.0793 & 0.642 & 0.630 & 0.606 \\
D\'ej\`a View$^\circ$ & 81.9\rkn{7} & 0.6410 & 0.7248 & 0.6829 & 0.686 & 0.622 & 0.635 \\
OmniVGGT & 74.4\rkn{8} & \rkB{0.9450} & \rkB{0.9606} & \rkB{0.9528} & 0.461 & 0.443 & 0.425 \\
AMB3R$^\circ$ & 70.9\rkn{9} & 1.0238 & 0.8585 & 0.9411 & 0.400 & 0.403 & 0.375 \\
FastVGGT & 70.5\rkn{10} & 1.2902 & 1.4918 & 1.3910 & \rkC{0.479} & \rkC{0.493} & \rkC{0.461} \\
Co-Me & 64.4\rkn{11} & 1.4029 & 1.6010 & 1.5019 & 0.397 & 0.430 & 0.385 \\
IGGT & 64.3\rkn{12} & 1.5134 & 1.5502 & 1.5318 & 0.406 & 0.441 & 0.400 \\
MapAnything & 61.5\rkn{13} & 9.9175 & 1.3965 & 5.6570 & 0.326 & 0.309 & 0.292 \\
RobustVGGT & 61.0\rkn{14} & 1.4505 & 1.6077 & 1.5291 & 0.369 & 0.368 & 0.351 \\
VGGT & 60.7\rkn{15} & 1.5302 & 1.6239 & 1.5771 & 0.369 & 0.376 & 0.352 \\
CUT3R & 52.5\rkn{16} & 1.8327 & 2.3505 & 2.0916 & 0.288 & 0.328 & 0.289 \\
Pow3R$^\circ$ & 39.9\rkn{17} & 0.7858 & 0.9732 & 0.8795 & 0.496 & 0.510 & 0.500 \\
DUSt3R$^\circ$ & 39.3\rkn{18} & 0.7131 & 0.8176 & 0.7654 & 0.456 & 0.462 & 0.455 \\
MASt3R$^\circ$ & 35.3\rkn{19} & 0.9791 & 2.3037 & 1.6414 & 0.277 & 0.281 & 0.245 \\
Fast3R$^\circ$ & 8.6\rkn{20} & 3.6236 & 2.9091 & 3.2663 & 0.044 & 0.055 & 0.020 \\
\bottomrule
\end{tabular}
\end{table*}

\paragraph{MVRecon.}
DA3-G (89.8) leads with an F-score of 0.630, followed by DA3-NGL (88.1) and $\pi^3$-X (87.7) (\cref{tab:exp:mvrecon,tab:suppl:res_mvrecon}). VGGT-$\Omega$, first overall, ranks fourth: its F-score (0.614) is close to those of the leaders, but its Acc (1.21\,m) is higher than theirs (0.73--0.93\,m).

\begin{table*}[t]
\centering
\setlength{\tabcolsep}{2pt}
\caption{\textbf{SfM, all metrics, \texttt{rgb} track}, averaged over all datasets. Errors in degrees ($\mu$: mean, $1/2$: median; RTAx: translation axis, RPAx: pose axis), ATE after $\mathrm{Sim}(3)$ alignment.}
\label{tab:suppl:res_sfm}
\resizebox{\textwidth}{!}{\input{tables/suppl_sfm_body}}
\end{table*}

\paragraph{SfM.}
VGGT-$\Omega$ attains the highest AUC@$30^\circ$ (0.873), followed by D\'ej\`a View (0.861) and DA3-NGL (0.837), which has the lowest ATE (0.807; \cref{tab:exp:sfm,tab:suppl:res_sfm}).

\paragraph{SLAM.}
VGGT-$\Omega$ leads the short and long ranges (\cref{tab:exp:slam,tab:suppl:res_slam}). The large range, entered only by long-sequence systems, is led by Pi-Long (94.6), which maintains a low ATE and drift on every dataset; Scal3R, R3, and VGGT-SLAM fail on some sequences and score 26--32.

\begin{table*}[t]
\centering
\scriptsize
\setlength{\tabcolsep}{4pt}
\caption{\textbf{SLAM on the short, long, and large range tracks, \texttt{rgb} track.} ATE RMSE in m and drift in \% per dataset. The score of a range is the mean, over every sequence of the range, of the normalized ATE and drift; a failed (\eg, out of memory) or missing sequence scores 0. $^\circ$Some sequences of the range failed; the metric means cover the others. OOM: every sequence of the dataset failed, mostly by running out of memory; TL: every sequence exceeds the model's maximum input length. Colors as in \cref{tab:exp:mono}.}
\label{tab:exp:slam}
\textbf{Short range}\\[1pt]
\begin{tabular}{@{}lrrrrrrrrr@{}}
\toprule
 &  & \multicolumn{2}{c}{7-Scenes} & \multicolumn{2}{c}{Oxford Spires} & \multicolumn{2}{c}{SpatialArena} & \multicolumn{2}{c}{TUM Dynamics} \\
\cmidrule(lr){3-4}\cmidrule(lr){5-6}\cmidrule(lr){7-8}\cmidrule(lr){9-10}
Method & Score$\uparrow$ & ATE$\downarrow$ & Drift$\downarrow$ & ATE$\downarrow$ & Drift$\downarrow$ & ATE$\downarrow$ & Drift$\downarrow$ & ATE$\downarrow$ & Drift$\downarrow$ \\
\midrule
VGGT-$\Omega$ & \rkA{92.7} & \rkA{0.040} & \rkA{11.93} & 13.099 & 129.66 & 1.756 & 66.21 & \rkA{0.010} & \rkA{13.54} \\
D\'ej\`a View & \rkB{92.0} & \rkB{0.056} & \rkB{12.85} & 8.182 & 49.76 & \rkC{1.685} & 43.55 & \rkB{0.013} & \rkC{17.31} \\
Pi-Long & \rkC{88.8} & 0.067 & 14.08 & \rkA{6.069} & \rkB{15.10} & \rkA{1.202} & \rkB{13.82} & 0.035 & 28.80 \\
$\pi^3$-X & 88.1\rkn{4} & \rkC{0.060} & 13.91 & 13.309 & 71.45 & \rkB{1.450} & \rkC{42.12} & 0.019 & 20.91 \\
LingBot-Map & 84.9\rkn{5} & 0.076 & 17.68 & \rkB{6.200} & \rkC{36.57} & 2.346 & 47.69 & 0.029 & 32.96 \\
IGGT4D & 84.2\rkn{6} & 0.075 & 15.29 & 14.483 & 92.36 & 5.008 & 117.63 & 0.018 & 22.14 \\
\bottomrule
\end{tabular}\\[1pt]
\textbf{Long range}\\[1pt]
\begin{tabular}{@{}lrrrrrrr@{}}
\toprule
 &  & \multicolumn{2}{c}{7-Scenes} & \multicolumn{2}{c}{SpatialArena} & \multicolumn{2}{c}{TUM Dynamics} \\
\cmidrule(lr){3-4}\cmidrule(lr){5-6}\cmidrule(lr){7-8}
Method & Score$\uparrow$ & ATE$\downarrow$ & Drift$\downarrow$ & ATE$\downarrow$ & Drift$\downarrow$ & ATE$\downarrow$ & Drift$\downarrow$ \\
\midrule
VGGT-$\Omega$ & \rkA{97.9} & \rkA{0.040} & \rkA{11.69} & \rkC{6.680} & 167.98 & \rkA{0.011} & \rkA{12.27} \\
D\'ej\`a View & \rkB{95.6} & \rkB{0.055} & \rkB{12.47} & 7.790 & 95.71 & \rkB{0.013} & \rkB{16.38} \\
$\pi^3$-X & \rkC{93.6} & \rkC{0.060} & \rkC{13.46} & 7.032 & 144.03 & \rkC{0.019} & \rkC{19.56} \\
LingBot-Map & 89.0\rkn{4} & 0.072 & 16.73 & \rkA{5.047} & 81.33 & 0.031 & 30.07 \\
LoGeR & 88.2\rkn{5} & 0.087 & 15.88 & 8.283 & \rkC{43.63} & 0.035 & 26.65 \\
ZipMap & 87.1\rkn{6} & 0.072 & 15.99 & 22.512 & 352.01 & 0.022 & 24.95 \\
\bottomrule
\end{tabular}\\[1pt]
\textbf{Large range}\\[1pt]
\begin{tabular}{@{}lrrrrrrrrr@{}}
\toprule
 &  & \multicolumn{2}{c}{KITTI Odometry} & \multicolumn{2}{c}{Oxford Spires} & \multicolumn{2}{c}{SpatialArena} & \multicolumn{2}{c}{VBR} \\
\cmidrule(lr){3-4}\cmidrule(lr){5-6}\cmidrule(lr){7-8}\cmidrule(lr){9-10}
Method & Score$\uparrow$ & ATE$\downarrow$ & Drift$\downarrow$ & ATE$\downarrow$ & Drift$\downarrow$ & ATE$\downarrow$ & Drift$\downarrow$ & ATE$\downarrow$ & Drift$\downarrow$ \\
\midrule
Pi-Long & \rkA{94.6} & \rkB{30.567} & \rkB{13.52} & \rkA{5.454} & \rkA{18.31} & \rkA{13.135} & \rkA{27.04} & \rkC{31.416} & \rkB{37.09} \\
ABot-Recon & \rkB{85.6} & 48.669 & \rkA{9.82} & 16.040 & \rkB{20.01} & 36.788 & \rkB{48.12} & \rkA{16.540} & \rkA{13.72} \\
LingBot-Map & \rkC{81.4} & \rkC{32.060} & 44.06 & \rkB{6.048} & 42.10 & \rkB{18.779} & 154.68 & \rkB{30.034} & 164.18 \\
VGGT-Long & 80.3\rkn{4} & \rkA{26.576} & \rkC{15.53} & \rkC{8.047} & \rkC{27.11} & \rkC{31.668} & \rkC{67.11} & 71.031 & \rkC{82.21} \\
LoGeR & 51.6\rkn{5} & 161.055 & 104.06 & 15.097 & 55.05 & 40.550 & 113.77 & 101.434 & 115.97 \\
Scal3R$^\circ$ & 32.0\rkn{6} & 71.791 & 107.91 & 26.748 & 160.11 & {\tiny OOM} & {\tiny OOM} & 109.803 & 213.28 \\
R3$^\circ$ & 26.9\rkn{7} & 196.635 & 153.75 & 23.902 & 161.73 & 43.517 & 125.53 & 82.194 & 249.96 \\
VGGT-SLAM$^\circ$ & 25.7\rkn{8} & 252.294 & 73.33 & 46.675 & 385.24 & 35.760 & 80.05 & 80.820 & 426.94 \\
ZipMap$^\circ$ & 3.1\rkn{9} & 189.152 & 334.57 & {\tiny OOM} & {\tiny OOM} & {\tiny OOM} & {\tiny OOM} & {\tiny OOM} & {\tiny OOM} \\
\bottomrule
\end{tabular}
\end{table*}

\begin{table*}[t]
\centering
\setlength{\tabcolsep}{2pt}
\caption{\textbf{SLAM, all metrics, \texttt{rgb} track}, per range and dataset. ATE and RPE$_t$ (over one second) in m, RPE$_r$ in degrees, drift and completeness in \%. Scores and marks as in \cref{tab:exp:slam}.}
\label{tab:suppl:res_slam}
\textbf{Short range}\\[1pt]
\resizebox{0.94\textwidth}{!}{
\begin{tabular}{@{}lrrrrrrrrrrrrrrrrrrrrrrrrr@{}}
\toprule
 &  & \multicolumn{6}{c}{7-Scenes} & \multicolumn{6}{c}{Oxford Spires} & \multicolumn{6}{c}{SpatialArena} & \multicolumn{6}{c}{TUM Dynamics} \\
\cmidrule(lr){3-8}\cmidrule(lr){9-14}\cmidrule(lr){15-20}\cmidrule(lr){21-26}
Method & Score$\uparrow$ & ATE$\downarrow$ & Drift$\downarrow$ & RPE$_t$$\downarrow$ & RPE$_r$$\downarrow$ & AUC@30$\uparrow$ & Compl.$\uparrow$ & ATE$\downarrow$ & Drift$\downarrow$ & RPE$_t$$\downarrow$ & RPE$_r$$\downarrow$ & AUC@30$\uparrow$ & Compl.$\uparrow$ & ATE$\downarrow$ & Drift$\downarrow$ & RPE$_t$$\downarrow$ & RPE$_r$$\downarrow$ & AUC@30$\uparrow$ & Compl.$\uparrow$ & ATE$\downarrow$ & Drift$\downarrow$ & RPE$_t$$\downarrow$ & RPE$_r$$\downarrow$ & AUC@30$\uparrow$ & Compl.$\uparrow$ \\
\midrule
VGGT-$\Omega$ & \rkA{92.7} & \rkA{0.040} & \rkA{11.93} & \rkA{0.040} & \rkA{1.247} & \rkA{0.859} & \rkA{1.000} & 13.099 & 129.66 & 5.327 & 11.178 & 0.672 & \rkA{1.000} & 1.756 & 66.21 & 1.314 & 7.858 & \rkB{0.776} & \rkA{1.000} & \rkA{0.010} & \rkA{13.54} & \rkA{0.011} & \rkA{0.533} & \rkA{0.844} & \rkA{1.000} \\
D\'ej\`a View & \rkB{92.0} & \rkB{0.056} & \rkB{12.85} & \rkB{0.044} & \rkB{1.451} & \rkB{0.815} & \rkA{1.000} & 8.182 & 49.76 & 2.333 & 15.665 & \rkC{0.745} & \rkA{1.000} & \rkC{1.685} & 43.55 & 0.804 & 9.689 & \rkC{0.773} & \rkA{1.000} & \rkB{0.013} & \rkC{17.31} & \rkB{0.014} & \rkB{0.559} & \rkC{0.792} & \rkA{1.000} \\
Pi-Long & \rkC{88.8} & 0.067 & 14.08 & 0.049 & 1.620 & 0.790 & \rkA{1.000} & \rkA{6.069} & \rkB{15.10} & \rkB{0.618} & \rkB{3.456} & \rkA{0.772} & \rkA{1.000} & \rkA{1.202} & \rkB{13.82} & \rkB{0.211} & \rkA{1.430} & \rkA{0.817} & \rkA{1.000} & 0.035 & 28.80 & 0.024 & 0.622 & 0.594 & \rkA{1.000} \\
$\pi^3$-X & 88.1\rkn{4} & \rkC{0.060} & 13.91 & \rkC{0.047} & 1.647 & \rkC{0.799} & \rkA{1.000} & 13.309 & 71.45 & 2.569 & 12.523 & 0.620 & \rkA{1.000} & \rkB{1.450} & \rkC{42.12} & 0.879 & 8.435 & 0.723 & \rkA{1.000} & 0.019 & 20.91 & 0.020 & 0.601 & 0.736 & \rkA{1.000} \\
\bottomrule
\end{tabular}}\\[4pt]
\textbf{Long range}\\[1pt]
\resizebox{0.94\textwidth}{!}{
\begin{tabular}{@{}lrrrrrrrrrrrrrrrrrrr@{}}
\toprule
 &  & \multicolumn{6}{c}{7-Scenes} & \multicolumn{6}{c}{SpatialArena} & \multicolumn{6}{c}{TUM Dynamics} \\
\cmidrule(lr){3-8}\cmidrule(lr){9-14}\cmidrule(lr){15-20}
Method & Score$\uparrow$ & ATE$\downarrow$ & Drift$\downarrow$ & RPE$_t$$\downarrow$ & RPE$_r$$\downarrow$ & AUC@30$\uparrow$ & Compl.$\uparrow$ & ATE$\downarrow$ & Drift$\downarrow$ & RPE$_t$$\downarrow$ & RPE$_r$$\downarrow$ & AUC@30$\uparrow$ & Compl.$\uparrow$ & ATE$\downarrow$ & Drift$\downarrow$ & RPE$_t$$\downarrow$ & RPE$_r$$\downarrow$ & AUC@30$\uparrow$ & Compl.$\uparrow$ \\
\midrule
VGGT-$\Omega$ & \rkA{97.9} & \rkA{0.040} & \rkA{11.69} & \rkA{0.040} & \rkA{1.255} & \rkA{0.854} & \rkA{1.000} & \rkC{6.680} & 167.98 & 4.656 & 5.461 & \rkC{0.615} & \rkA{1.000} & \rkA{0.011} & \rkA{12.27} & \rkA{0.011} & \rkA{0.555} & \rkA{0.839} & \rkA{1.000} \\
D\'ej\`a View & \rkB{95.6} & \rkB{0.055} & \rkB{12.47} & \rkB{0.044} & \rkB{1.432} & \rkB{0.813} & \rkA{1.000} & 7.790 & 95.71 & 1.734 & 2.661 & \rkB{0.655} & \rkA{1.000} & \rkB{0.013} & \rkB{16.38} & \rkB{0.015} & \rkB{0.565} & \rkB{0.782} & \rkA{1.000} \\
$\pi^3$-X & \rkC{93.6} & \rkC{0.060} & \rkC{13.46} & \rkC{0.047} & 1.655 & \rkC{0.796} & \rkA{1.000} & 7.032 & 144.03 & 2.576 & 4.913 & 0.552 & \rkA{1.000} & \rkC{0.019} & \rkC{19.56} & \rkC{0.020} & \rkC{0.625} & 0.732 & \rkA{1.000} \\
LingBot-Map & 89.0\rkn{4} & 0.072 & 16.73 & 0.056 & 1.728 & 0.765 & \rkA{1.000} & \rkA{5.047} & 81.33 & 1.385 & 2.661 & 0.606 & \rkA{1.000} & 0.031 & 30.07 & 0.031 & 0.807 & 0.640 & \rkA{1.000} \\
LoGeR & 88.2\rkn{5} & 0.087 & 15.88 & 0.055 & \rkC{1.567} & 0.749 & \rkA{1.000} & 8.283 & \rkC{43.63} & 0.827 & 2.923 & 0.517 & \rkA{1.000} & 0.035 & 26.65 & 0.026 & 0.788 & 0.588 & \rkA{1.000} \\
ZipMap & 87.1\rkn{6} & 0.072 & 15.99 & 0.056 & 1.662 & 0.776 & \rkA{1.000} & 22.512 & 352.01 & 6.699 & 7.562 & 0.278 & \rkA{1.000} & 0.022 & 24.95 & 0.026 & 0.639 & 0.715 & \rkA{1.000} \\
\bottomrule
\end{tabular}}\\[4pt]
\textbf{Large range}\\[1pt]
\resizebox{0.94\textwidth}{!}{
\begin{tabular}{@{}lrrrrrrrrrrrrrrrrrrrrrrrrr@{}}
\toprule
 &  & \multicolumn{6}{c}{KITTI Odometry} & \multicolumn{6}{c}{Oxford Spires} & \multicolumn{6}{c}{SpatialArena} & \multicolumn{6}{c}{VBR} \\
\cmidrule(lr){3-8}\cmidrule(lr){9-14}\cmidrule(lr){15-20}\cmidrule(lr){21-26}
Method & Score$\uparrow$ & ATE$\downarrow$ & Drift$\downarrow$ & RPE$_t$$\downarrow$ & RPE$_r$$\downarrow$ & AUC@30$\uparrow$ & Compl.$\uparrow$ & ATE$\downarrow$ & Drift$\downarrow$ & RPE$_t$$\downarrow$ & RPE$_r$$\downarrow$ & AUC@30$\uparrow$ & Compl.$\uparrow$ & ATE$\downarrow$ & Drift$\downarrow$ & RPE$_t$$\downarrow$ & RPE$_r$$\downarrow$ & AUC@30$\uparrow$ & Compl.$\uparrow$ & ATE$\downarrow$ & Drift$\downarrow$ & RPE$_t$$\downarrow$ & RPE$_r$$\downarrow$ & AUC@30$\uparrow$ & Compl.$\uparrow$ \\
\midrule
Pi-Long & \rkA{94.6} & \rkB{30.567} & \rkB{13.52} & \rkC{2.531} & \rkB{0.709} & \rkA{0.845} & \rkA{1.000} & \rkA{5.454} & \rkA{18.31} & \rkB{0.454} & \rkA{0.605} & \rkA{0.806} & \rkA{1.000} & \rkA{13.135} & \rkA{27.04} & \rkA{0.328} & \rkB{2.714} & \rkA{0.540} & \rkA{1.000} & \rkC{31.416} & \rkB{37.09} & \rkB{0.727} & \rkB{1.111} & \rkA{0.627} & \rkA{1.000} \\
ABot-Recon & \rkB{85.6} & 48.669 & \rkA{9.82} & \rkA{1.165} & 1.444 & 0.274 & \rkA{1.000} & 16.040 & \rkB{20.01} & \rkA{0.370} & 2.048 & 0.098 & \rkA{1.000} & 36.788 & \rkB{48.12} & \rkB{0.540} & \rkA{1.466} & 0.047 & \rkA{1.000} & \rkA{16.540} & \rkA{13.72} & \rkA{0.255} & \rkA{1.074} & \rkC{0.310} & \rkA{1.000} \\
LingBot-Map & \rkC{81.4} & \rkC{32.060} & 44.06 & 6.785 & 1.731 & 0.719 & \rkA{1.000} & \rkB{6.048} & 42.10 & 1.137 & 5.854 & \rkB{0.747} & \rkA{1.000} & \rkB{18.779} & 154.68 & 2.168 & 9.702 & \rkB{0.330} & \rkA{1.000} & \rkB{30.034} & 164.18 & 4.286 & 5.876 & \rkB{0.522} & \rkA{1.000} \\
\bottomrule
\end{tabular}}
\end{table*}

\paragraph{Auxiliary-input tracks.}
We rank the models given ground-truth cameras (\texttt{rgbc}) or depth (\texttt{rgbd}) separately and do not include them in \cref{tab:leaderboard} (\cref{tab:exp:aux,tab:suppl:res_aux}). The effect of these priors varies with the model and the dataset (\cref{sec:suppl:disc_priors}).

\begin{table*}[t]
\centering
\scriptsize
\setlength{\tabcolsep}{2pt}
\setlength{\aboverulesep}{0pt}
\setlength{\belowrulesep}{0pt}
\renewcommand{\arraystretch}{1.12}
\caption{\textbf{Leaderboard on the public datasets}, as in \cref{tab:leaderboard}, except that \oursdata does not enter the task scores (ScanNet++ and Spring remain excluded).}
\label{tab:suppl:leaderboard_public}
\adjustbox{max width=\textwidth}{
\begin{tabular}{@{}rlrrrrrrrrr@{\hspace{5pt}}|@{\hspace{5pt}}rlrrrrrrrrr@{}}
\toprule
\# & Method & Score & AR & Cov. & Wins & Mono & Video & MVR & SfM & SLAM & \# & Method & Score & AR & Cov. & Wins & Mono & Video & MVR & SfM & SLAM \\
\midrule
1 & VGGT-$\Omega$ & \rkA{87.5} & \rkA{3.4} & 360 & 970 & \rkB{95.7} & \rkA{96.2} & 83.9\rkn{7} & \rkA{94.7} & 66.8\rkn{6} & 27 & InfiniDepth & 17.9 & 34.0 & 14 & 3 & 89.6\rkn{10} & -- & -- & -- & -- \\
2 & $\pi^3$-X & \rkB{85.7} & \rkB{5.8} & 360 & 593 & 94.1\rkn{6} & \rkB{92.0} & \rkC{90.2} & 89.0\rkn{9} & 63.3\rkn{9} & 28 & Pi-Long & 17.8 & 32.2 & 92 & 25 & -- & -- & -- & -- & \rkA{89.1} \\
3 & D\'ej\`a View & \rkC{84.1} & \rkC{6.2} & 360 & 677 & 86.5\rkn{13} & 87.1\rkn{5} & 88.2\rkn{4} & \rkB{93.1} & 65.6\rkn{7} & 29 & Depth Pro & 17.5 & 34.2 & 14 & 4 & 87.7\rkn{11} & -- & -- & -- & -- \\
4 & IGGT4D & 81.7 & 8.4 & 360 & 650 & 87.7\rkn{12} & 85.7\rkn{6} & 84.0\rkn{6} & 90.5\rkn{8} & 60.8\rkn{10} & 30 & Marigold V2 & 17.3 & 34.8 & 14 & 1 & 86.4\rkn{14} & -- & -- & -- & -- \\
5 & DA3-L & 79.9 & 10.4 & 360 & 469 & 85.2\rkn{17} & 85.0\rkn{7} & 86.1\rkn{5} & 87.9\rkn{10} & 55.3\rkn{13} & 31 & LingBot-Map & 17.2 & 32.4 & 92 & 3 & -- & -- & -- & -- & \rkB{86.2} \\
6 & DA3-G & 79.8 & 6.6 & 360 & 825 & 92.3\rkn{7} & 87.9\rkn{4} & \rkA{92.0} & 92.4\rkn{4} & 34.5\rkn{17} & 32 & GeoNeXt & 16.5 & 35.8 & 14 & 0 & 82.6\rkn{19} & -- & -- & -- & -- \\
7 & DA3-NGL & 79.7 & 7.0 & 360 & 898 & 90.7\rkn{9} & \rkC{88.8} & \rkB{91.9} & \rkC{92.7} & 34.3\rkn{18} & 33 & Lotus-D & 16.3 & 36.0 & 14 & 0 & 81.4\rkn{20} & -- & -- & -- & -- \\
\bottomrule
\end{tabular}}
\end{table*}

\begin{table*}[t]
\centering
\scriptsize
\setlength{\tabcolsep}{3pt}
\caption{\textbf{Auxiliary-input tracks}: models given ground-truth cameras (\texttt{rgbc}) or depth (\texttt{rgbd}), with the headline metrics of \cref{tab:exp:mono,tab:exp:video,tab:exp:mvrecon,tab:exp:sfm,tab:exp:slam}. Ranks are within each table, and these tracks do not enter \cref{tab:leaderboard}. (a) We show VideoDepth \texttt{rgbc} on \oursdata, the only dataset on which we evaluated all its models; it has no task score. Colors, $^\circ$, and OOM as in \cref{tab:exp:mono,tab:exp:slam}.}
\label{tab:exp:aux}
\begin{minipage}[t]{0.38\textwidth}
\centering
(a) VideoDepth, \texttt{rgbc}\\[1pt]
\adjustbox{max width=\linewidth}{
\begin{tabular}{@{}lrrrr@{}}
\toprule
Method & Score$\uparrow$ & AbsRel$\downarrow$ & $\delta_{1.03}$$\uparrow$ & $\delta_{1.25}$$\uparrow$ \\
\midrule
$\pi^3$-X & \rkA{92.7} & \rkA{11.41} & \rkA{34.82} & \rkA{85.85} \\
MapAnything & \rkB{83.7} & \rkB{16.91} & \rkB{25.67} & \rkB{79.80} \\
DA3-NGL & \rkC{79.0} & \rkC{18.48} & \rkC{22.78} & \rkC{77.79} \\
OmniVGGT$^\circ$ & 60.1\rkn{4} & 18.84 & 17.93 & 73.09 \\
\bottomrule
\end{tabular}}
\end{minipage}\hfill
\begin{minipage}[t]{0.6\textwidth}
\centering
(b) MVRecon, \texttt{rgbc}\\[1pt]
\adjustbox{max width=\linewidth}{
\begin{tabular}{@{}lrrrrr@{}}
\toprule
Method & Score$\uparrow$ & Acc$\downarrow$ & Comp$\downarrow$ & Overall$\downarrow$ & F-score$\uparrow$ \\
\midrule
$\pi^3$-X$^\circ$ & \rkA{89.9} & 0.7602 & 0.6751 & 0.7176 & 0.631 \\
OmniVGGT$^\circ$ & \rkB{80.0} & 0.8990 & 0.8900 & 0.8945 & 0.526 \\
MapAnything$^\circ$ & \rkC{77.7} & 0.9711 & 0.9920 & 0.9816 & 0.489 \\
DA3-NGL & 52.7\rkn{4} & \rkA{1.7844} & \rkA{1.7552} & \rkA{1.7698} & \rkA{0.208} \\
\bottomrule
\end{tabular}}
\end{minipage}\\[6pt]
\begin{minipage}[t]{0.42\textwidth}
\centering
(c) MVRecon, \texttt{rgbd}\\[1pt]
\adjustbox{max width=\linewidth}{
\begin{tabular}{@{}lrrrrr@{}}
\toprule
Method & Score$\uparrow$ & Acc$\downarrow$ & Comp$\downarrow$ & Overall$\downarrow$ & F-score$\uparrow$ \\
\midrule
OmniVGGT & \rkA{60.3} & \rkB{7.3889} & \rkA{2.6814} & \rkB{5.0351} & \rkB{0.333} \\
$\pi^3$-X & \rkB{59.0} & \rkA{7.1369} & \rkC{2.7304} & \rkA{4.9336} & \rkA{0.404} \\
MapAnything & \rkC{50.9} & \rkC{338.5147} & \rkB{2.7176} & \rkC{170.6161} & \rkC{0.301} \\
\bottomrule
\end{tabular}}
\end{minipage}\hfill
\begin{minipage}[t]{0.56\textwidth}
\centering
(d) SfM, \texttt{rgbd}\\[1pt]
\adjustbox{max width=\linewidth}{\input{tables/exp_sfm_rgbd_body}}
\end{minipage}\\[6pt]
(e) SLAM short range, \texttt{rgbd}\\[1pt]
\adjustbox{max width=\linewidth}{\input{tables/exp_slam_short_rgbd_body}}\\[6pt]
(f) SLAM long range, \texttt{rgbd}\\[1pt]
\adjustbox{max width=\linewidth}{\input{tables/exp_slam_long_rgbd_body}}
\end{table*}

\begin{table*}[t]
\centering
\scriptsize
\setlength{\tabcolsep}{3pt}
\caption{\textbf{Auxiliary-input tracks, all metrics}, one table per (task, track); units as in \cref{tab:suppl:res_mono,tab:suppl:res_mvrecon,tab:suppl:res_sfm,tab:suppl:res_slam}, and (a) on \oursdata as in \cref{tab:exp:aux}.}
\label{tab:suppl:res_aux}
(a) VideoDepth, \texttt{rgbc}\\[1pt]
\adjustbox{max width=\linewidth}{\input{tables/suppl_videodepth_rgbc_body}}\\[2pt]
\begin{minipage}[t]{0.49\textwidth}
\centering
(b) MVRecon, \texttt{rgbc}\\[1pt]
\adjustbox{max width=\linewidth}{\input{tables/suppl_mvrecon_rgbc_body}}
\end{minipage}\hfill
\begin{minipage}[t]{0.49\textwidth}
\centering
(c) MVRecon, \texttt{rgbd}\\[1pt]
\adjustbox{max width=\linewidth}{\input{tables/suppl_mvrecon_rgbd_body}}
\end{minipage}\\[2pt]
(d) SfM, \texttt{rgbd}\\[1pt]
\begin{adjustbox}{max width=\textwidth}\input{tables/suppl_sfm_rgbd_body}\end{adjustbox}\\[2pt]
(e) SLAM short range, \texttt{rgbd}\\[1pt]
\begin{adjustbox}{max width=\textwidth}\input{tables/suppl_slam_short_rgbd_body}\end{adjustbox}\\[2pt]
(f) SLAM long range, \texttt{rgbd}\\[1pt]
\begin{adjustbox}{max width=\textwidth}\input{tables/suppl_slam_long_rgbd_body}\end{adjustbox}
\end{table*}

\begin{table*}[t]
\centering
\scriptsize
\setlength{\tabcolsep}{3pt}
\setlength{\aboverulesep}{0pt}
\setlength{\belowrulesep}{0pt}
\renewcommand{\arraystretch}{1.12}
\caption{\textbf{MonoDepth, \texttt{rgb} track, on the public datasets}, as in \cref{tab:exp:mono}, except that \oursdata does not enter the scores and averages. Score is the task score of \cref{tab:suppl:leaderboard_public}; colors, ranks, and $^\circ$ as in \cref{tab:exp:mono}.}
\label{tab:suppl:mono_public}
\adjustbox{max width=\textwidth}{\input{tables/appx_monodepth_public_body}}
\end{table*}

\newcommand{\trendwide}[2]{%
  \begin{tabular}[b]{@{}c@{}}\includegraphics[width=0.49\textwidth]{images/exp/#1}\\[-1pt]\footnotesize #2\end{tabular}}
\suppfigpage{%
\centering
\trendwide{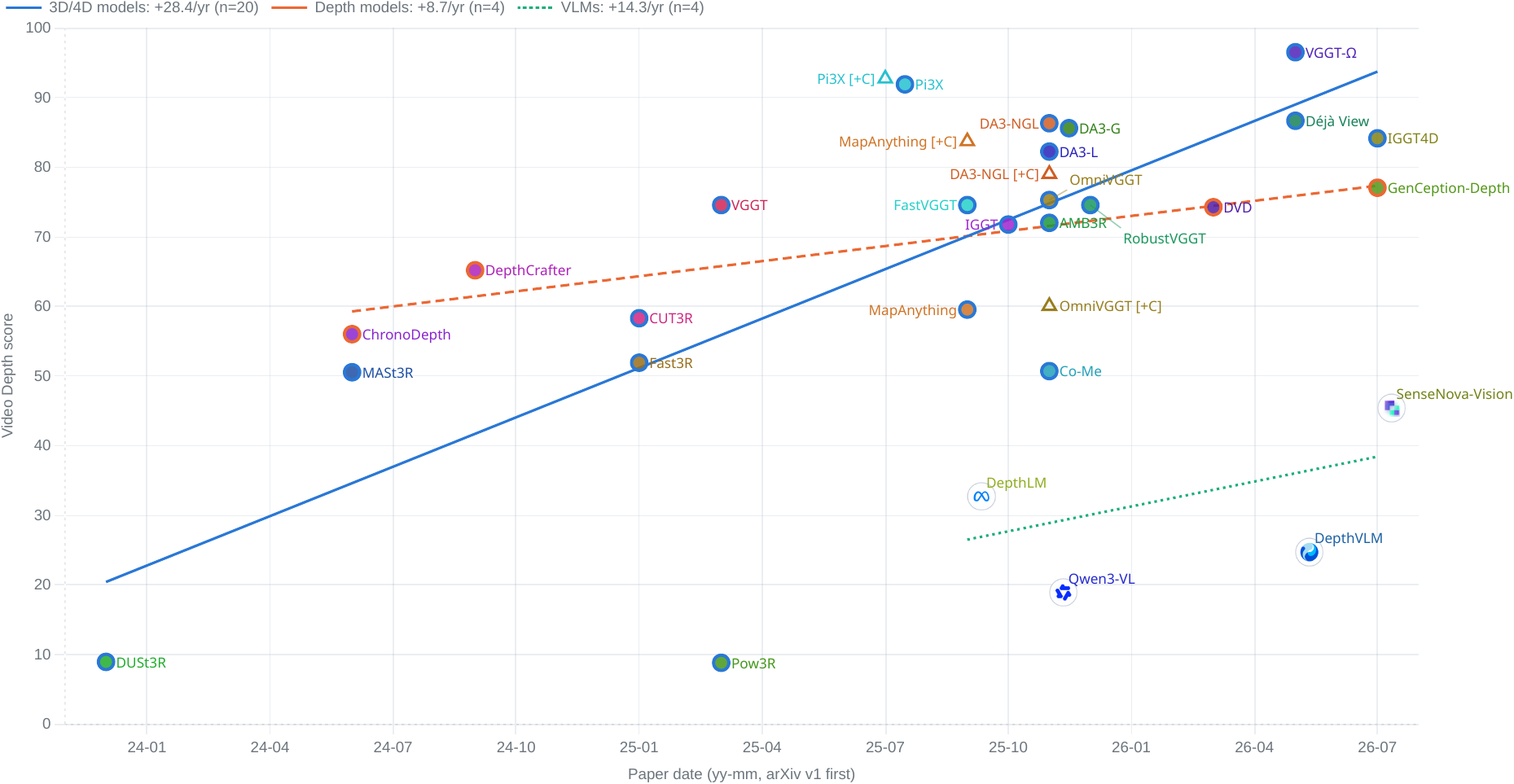}{(a) VideoDepth}\hfill
\trendwide{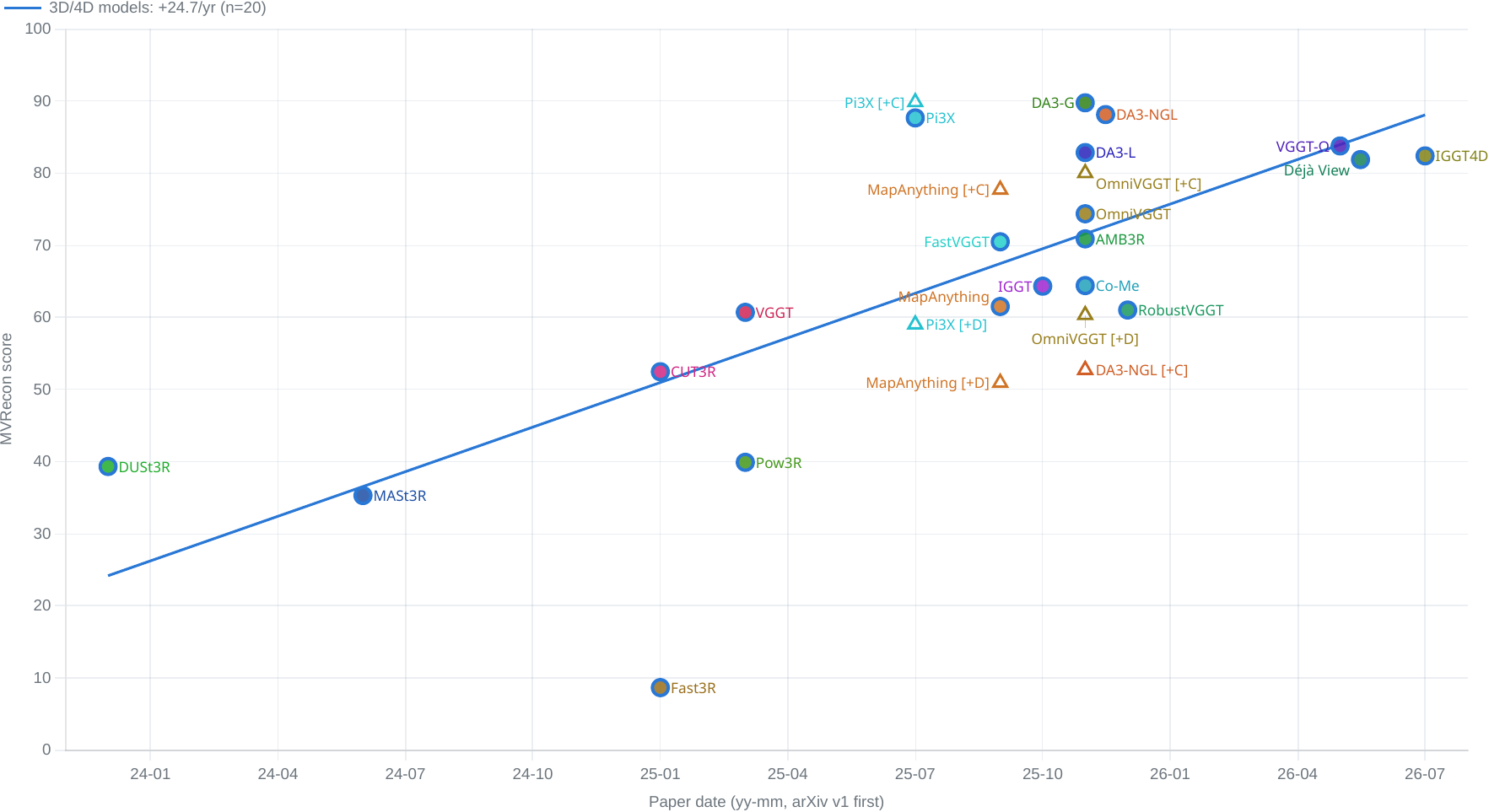}{(b) MVRecon}
\captionof{figure}{\textbf{Task score against paper date} (first arXiv version) for VideoDepth and MVRecon, as in \cref{fig:exp:trends}. Circles are the \texttt{rgb} track; triangles mark the \texttt{rgbc} ([+C]) and \texttt{rgbd} ([+D]) tracks.}
\label{fig:suppl:trends}
}
\suppfigpage{%
\centering
\trendwide{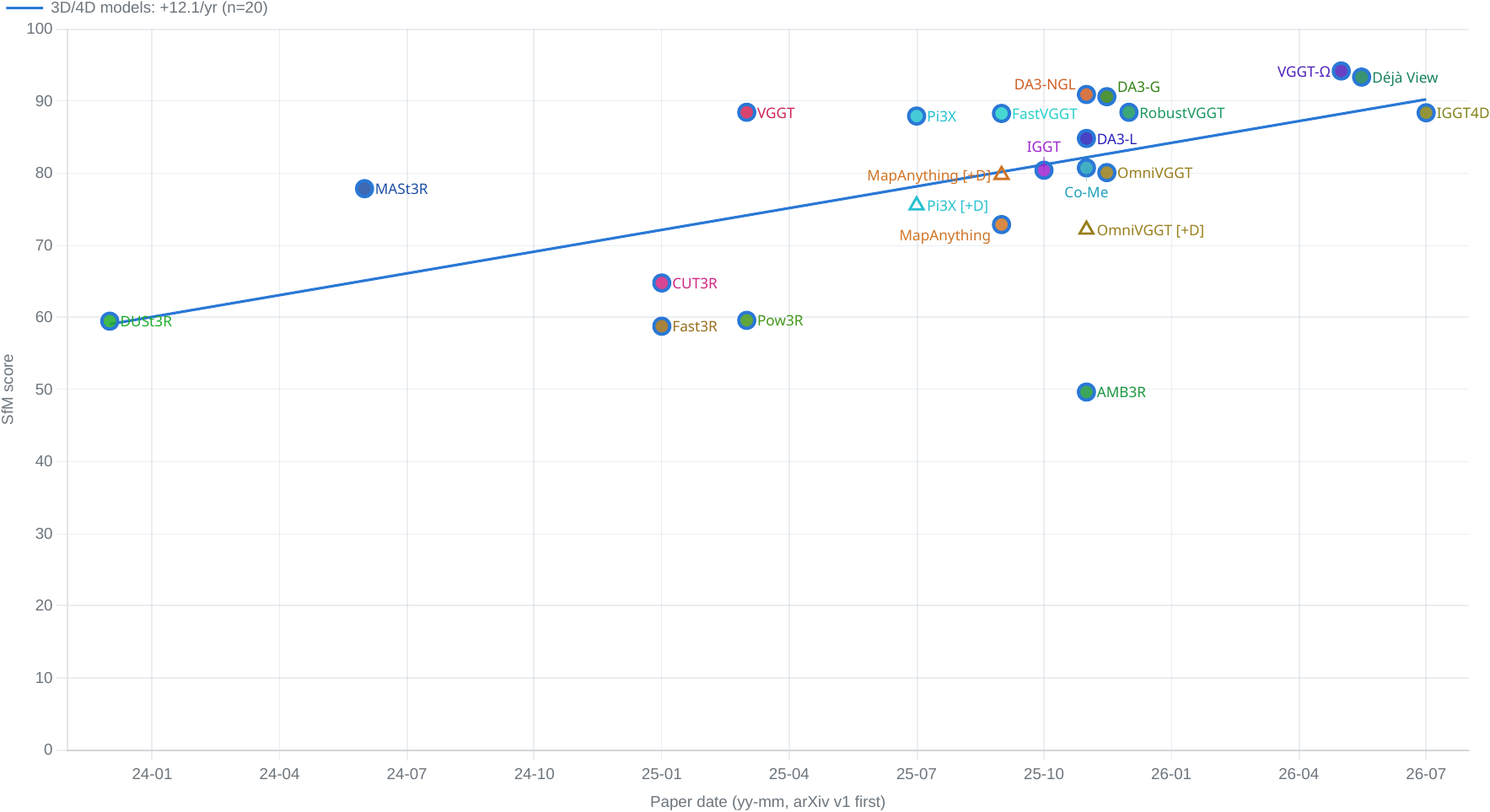}{(a) SfM}\hfill
\trendwide{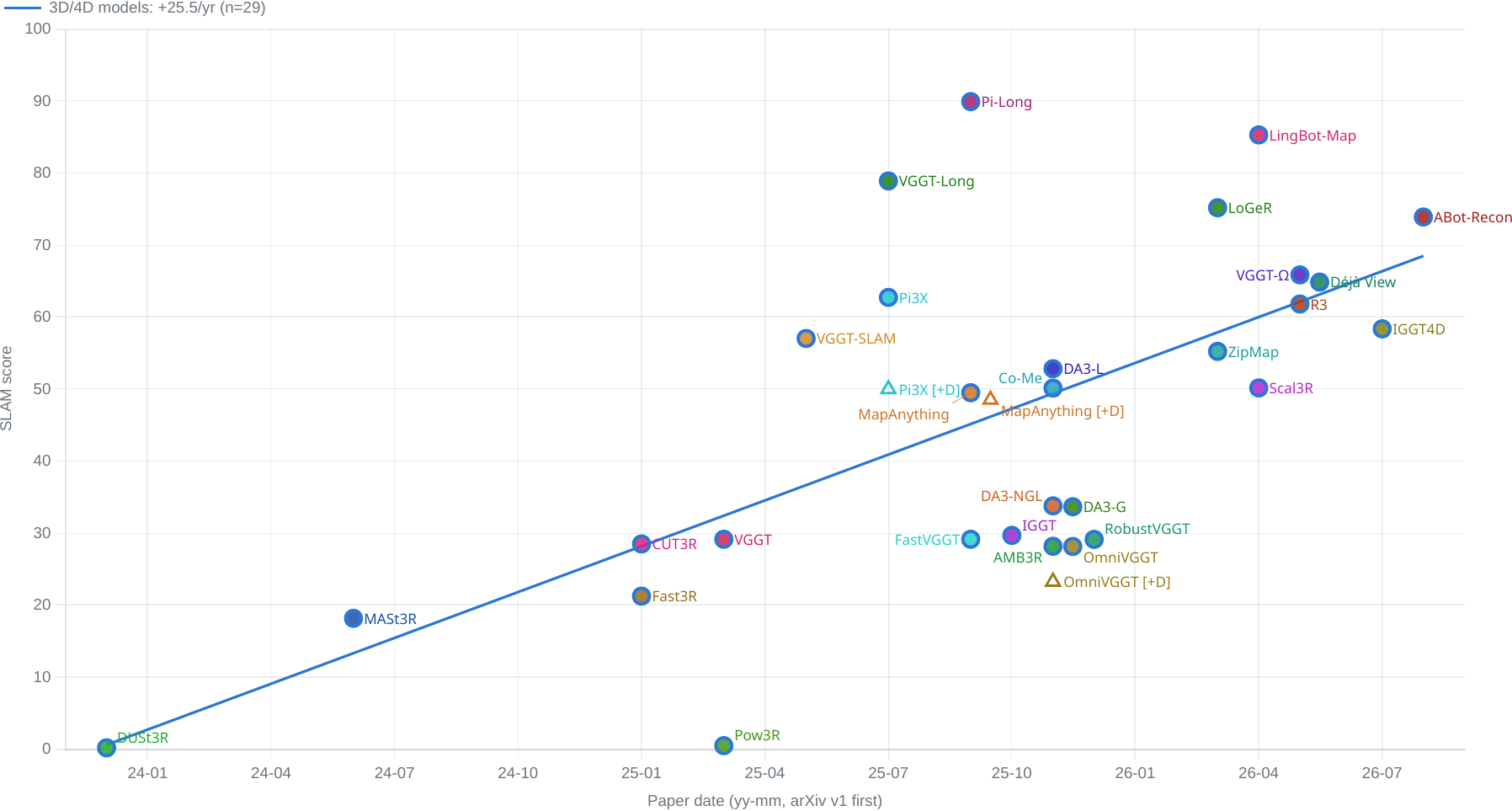}{(b) SLAM}
\captionof{figure}{\textbf{Task score against paper date} for SfM and SLAM, shown as in \cref{fig:suppl:trends}.}
\label{fig:suppl:trends2}
}

\subsection{Progress over time}
\label{sec:suppl:res_trends}

In contrast to MonoDepth, where the feed-forward 3D/4D models improve by 8.4 points per year with \oursdata scored (\cref{fig:exp:trends}; 2.8 on the public datasets alone, \cref{sec:suppl:res_public}), we observe steep progress on the other tasks (\cref{fig:suppl:trends,fig:suppl:trends2}). On VideoDepth, the 3D/4D models improve by 28.4 points per year, compared with 8.7 for the depth models and 14.3 for the VLMs; the earliest 3D/4D models score below the video depth models, and the recent ones above them. On MVRecon and SLAM, the 3D/4D models improve by 24.7 and 25.5 points per year (the SLAM fit includes the long-sequence systems), and on SfM by 12.1 points per year. We exclude the runs given ground-truth cameras or depth (triangles) from the fits.

\subsection{Scores on the public datasets}
\label{sec:suppl:res_public}

All tables and trends of this paper score and average \oursdata like a public dataset. Existing benchmarks contain public data only; for comparison with them, \cref{tab:suppl:leaderboard_public,tab:suppl:mono_public} give the leaderboard and the MonoDepth table with \oursdata left out of the scores and averages. Here we analyze how scoring \oursdata changes the ranking.

The top of the ranking is stable. VGGT-$\Omega$, $\pi^3$-X, D\'ej\`a View, and IGGT4D keep the first four places, and the first 20 places remain feed-forward 3D/4D models. Below them, the models that transfer well to \oursdata move up: DA3-G and DA3-NGL pass DA3-L, IGGT rises from 16th to 15th (score 53.1 to 59.2), and FastVGGT and MapAnything gain one place each, whereas AMB3R falls from 11th to 13th and CUT3R from 15th to 16th. GeoNeXt, which has no \oursdata MonoDepth result, falls from 32nd to 50th, since its missing cells count as 0.

MonoDepth changes most, because \oursdata is the only MonoDepth dataset on which the 3D/4D models are clearly more accurate than the depth specialists (\cref{sec:suppl:disc_general}). VGGT-$\Omega$ replaces MoGe-3 as the leader (score 96.7 against 92.1; mean AbsRel 10.76\% against 11.57\%), $\pi^3$-X and AMB3R rise from 6th and 8th to 3rd and 4th, and MoGe-2 falls from 3rd to 7th. The mean AbsRel more than doubles for the leading models (MoGe-3: 4.94\% to 11.57\%), since \oursdata adds errors three to four times those of the public datasets. The trends change accordingly (\cref{fig:exp:trends}): with \oursdata, the MonoDepth score of the depth models decreases by 2.8 points per year instead of increasing by 1.9, whereas that of the 3D/4D models increases by 8.4 instead of 2.8 points per year (VLMs: 31.5 instead of 33.5). The newest specialists degrade most on \oursdata (MoGe-3 and InfiniDepth reach $4.2\times$ their public AbsRel, Marigold $2.9\times$), so their gains on the public datasets do not carry over to large, uncontrolled scenes.

\newcommand{\supqualpanel}[3][0.9\textwidth]{%
  \begin{tabular}[b]{@{}c@{}}\includegraphics[width=#1]{images/artifacts/#2}\\[-1pt]\footnotesize #3\end{tabular}}

\suppfigpage{%
\centering
\supqualpanel[0.9\textwidth]{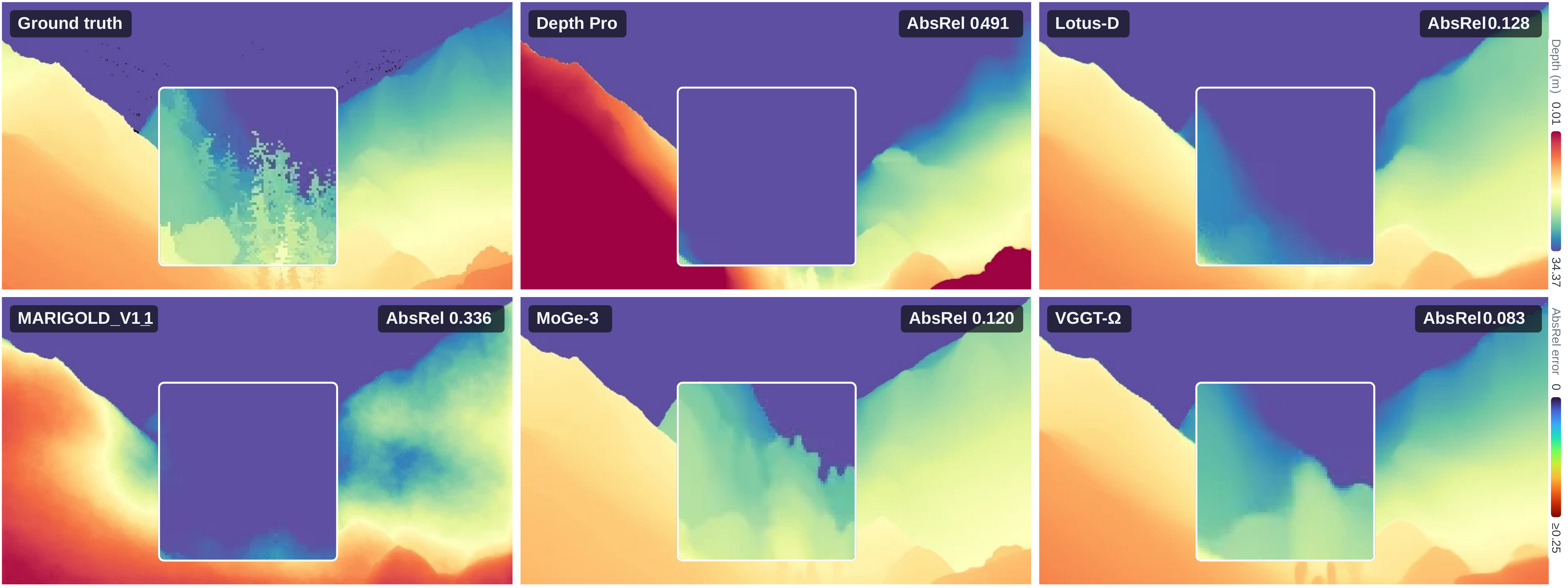}{(a) Spring$^\dagger$}\\[4pt]
\supqualpanel[0.9\textwidth]{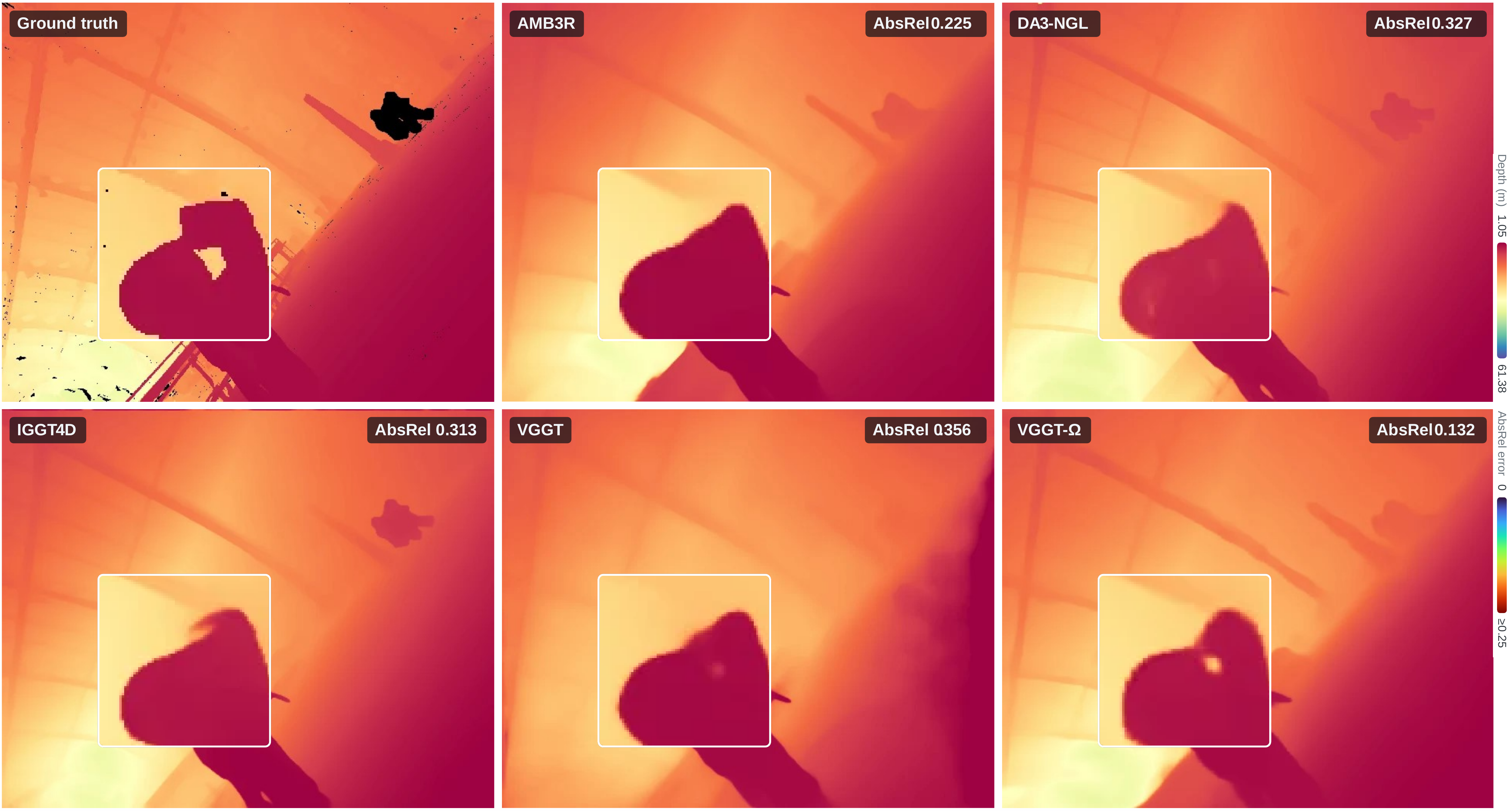}{(b) \oursdata, StPancrasStatue1}
\captionof{figure}{\textbf{MonoDepth predictions} on one frame of Spring and of \oursdata, shown as in \cref{fig:exp:depth}. $^\dagger$Excluded from the ranking.}
\label{fig:suppl:depth}
}

\subsection{Qualitative results}
\label{sec:suppl:res_qual}

\paragraph{MonoDepth.}
On the Spring and \oursdata frames of \cref{fig:suppl:depth}, VGGT-$\Omega$, first in MonoDepth, attains the lowest error, and most models smooth away thin structures such as the trees in Spring. VGGT, ranked 15th, has the highest error on the \oursdata frame.

\suppfigpage{%
\centering
\supqualpanel[0.9\textwidth]{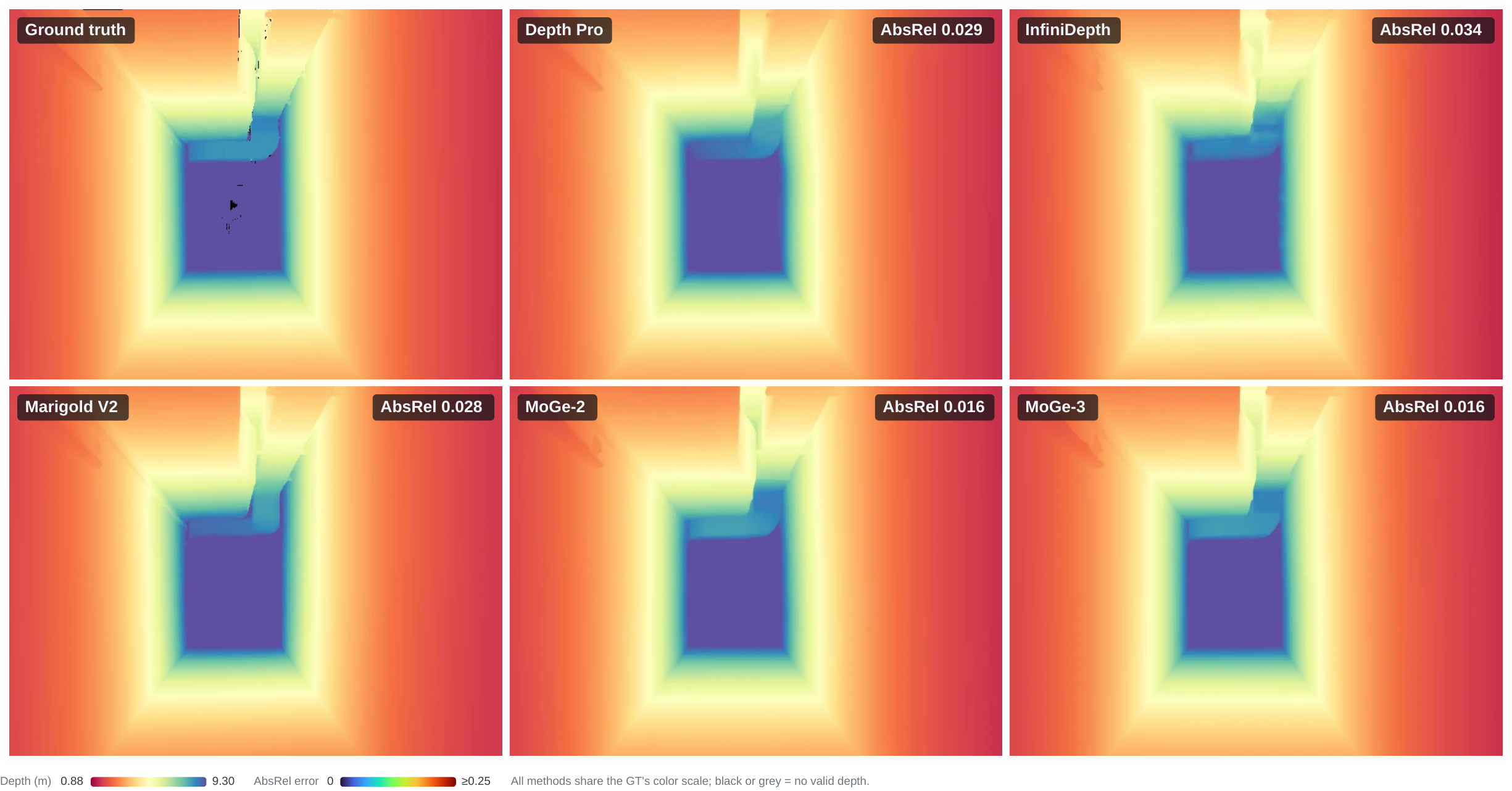}{(a) iBims-1, room}\\[4pt]
\supqualpanel[0.9\textwidth]{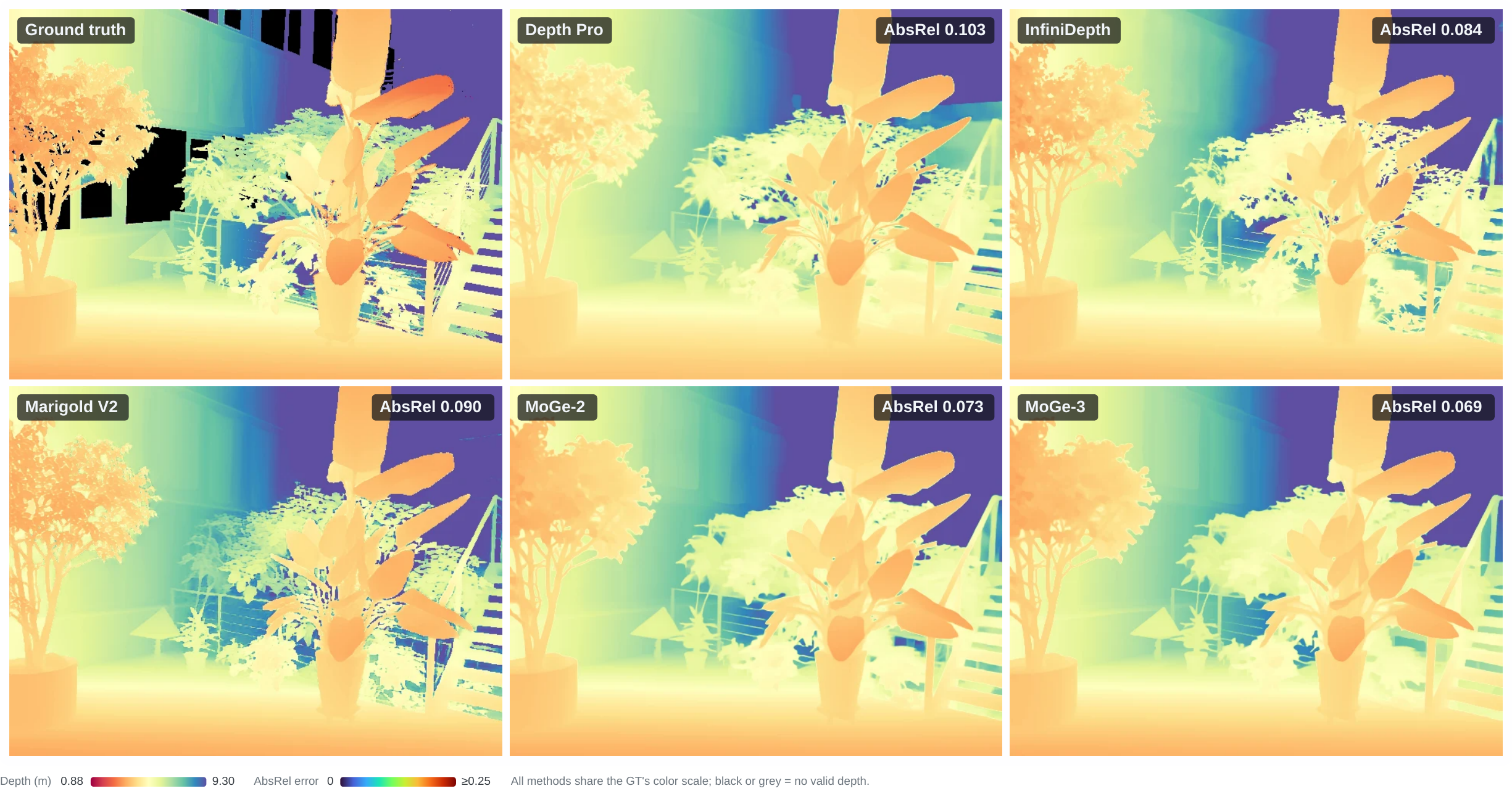}{(b) iBims-1, plants}
\captionof{figure}{\textbf{MonoDepth predictions of the depth specialists} on two iBims-1 frames, shown as in \cref{fig:exp:depth}. On the simple room (a), all models are within 0.016--0.034 AbsRel, and MoGe-2 and MoGe-3 are best. On the cluttered frame (b), the thin leaves and railings raise every error by a factor of 2.5--4.6; MoGe-3 is again best (0.069) and Depth Pro worst (0.103).}
\label{fig:suppl:depth_ibims}
}

\begin{figure*}[!tp]
\centering
\supqualpanel[0.44\textwidth]{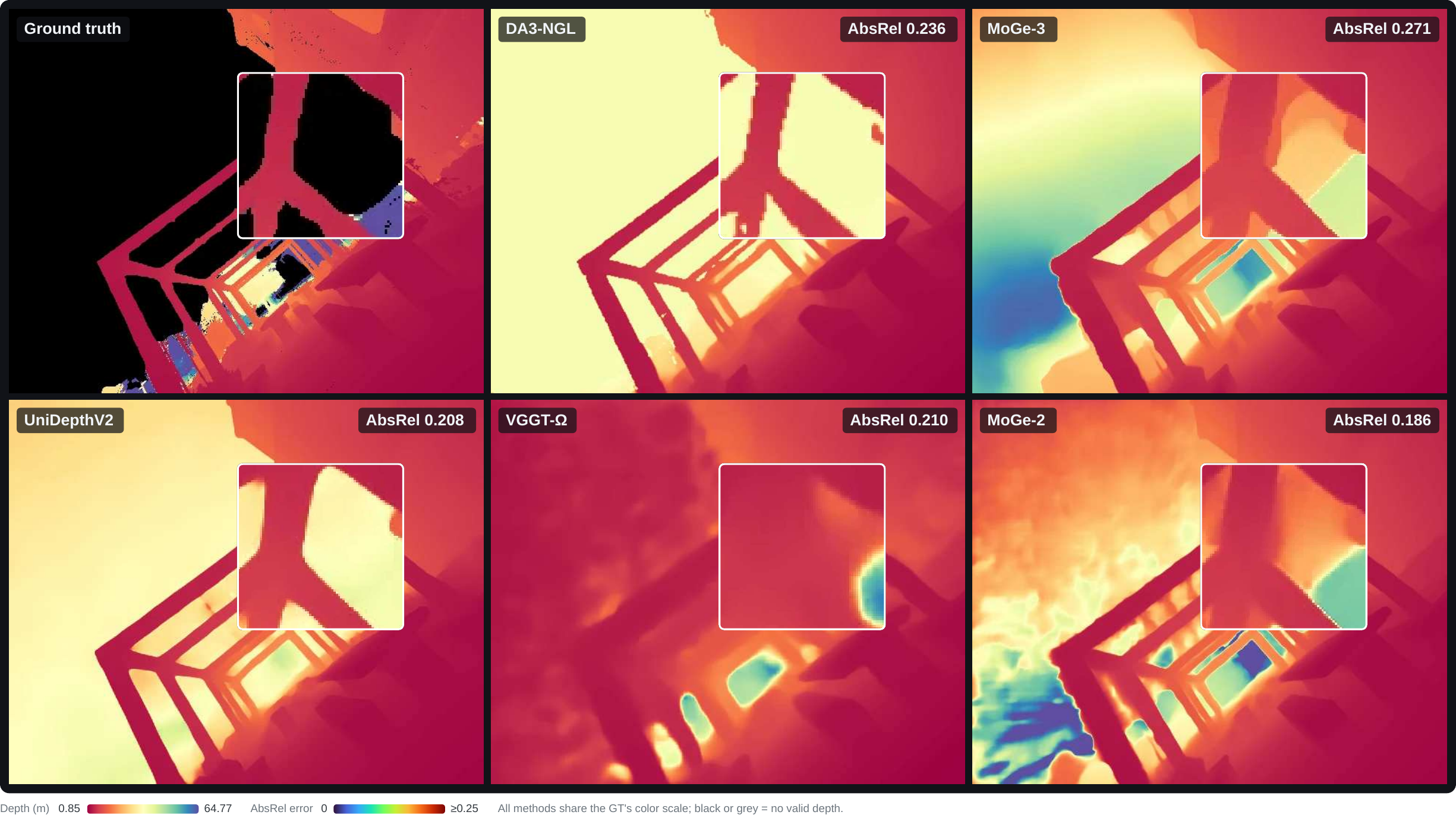}{(a) CambridgeSmallGarden1, depth}\hspace{0.02\textwidth}
\supqualpanel[0.44\textwidth]{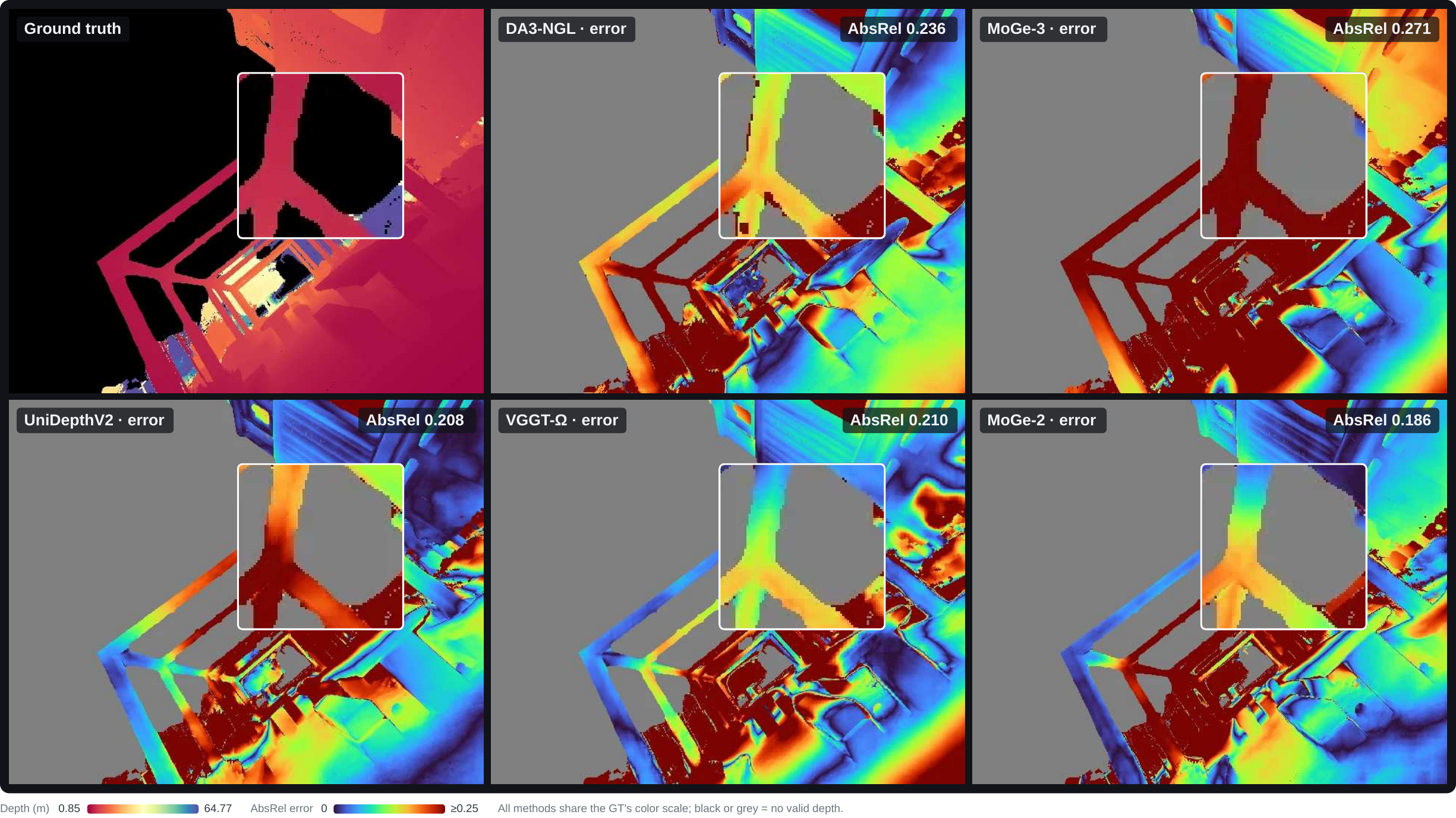}{(b) CambridgeSmallGarden1, error}\\[4pt]
\supqualpanel[0.44\textwidth]{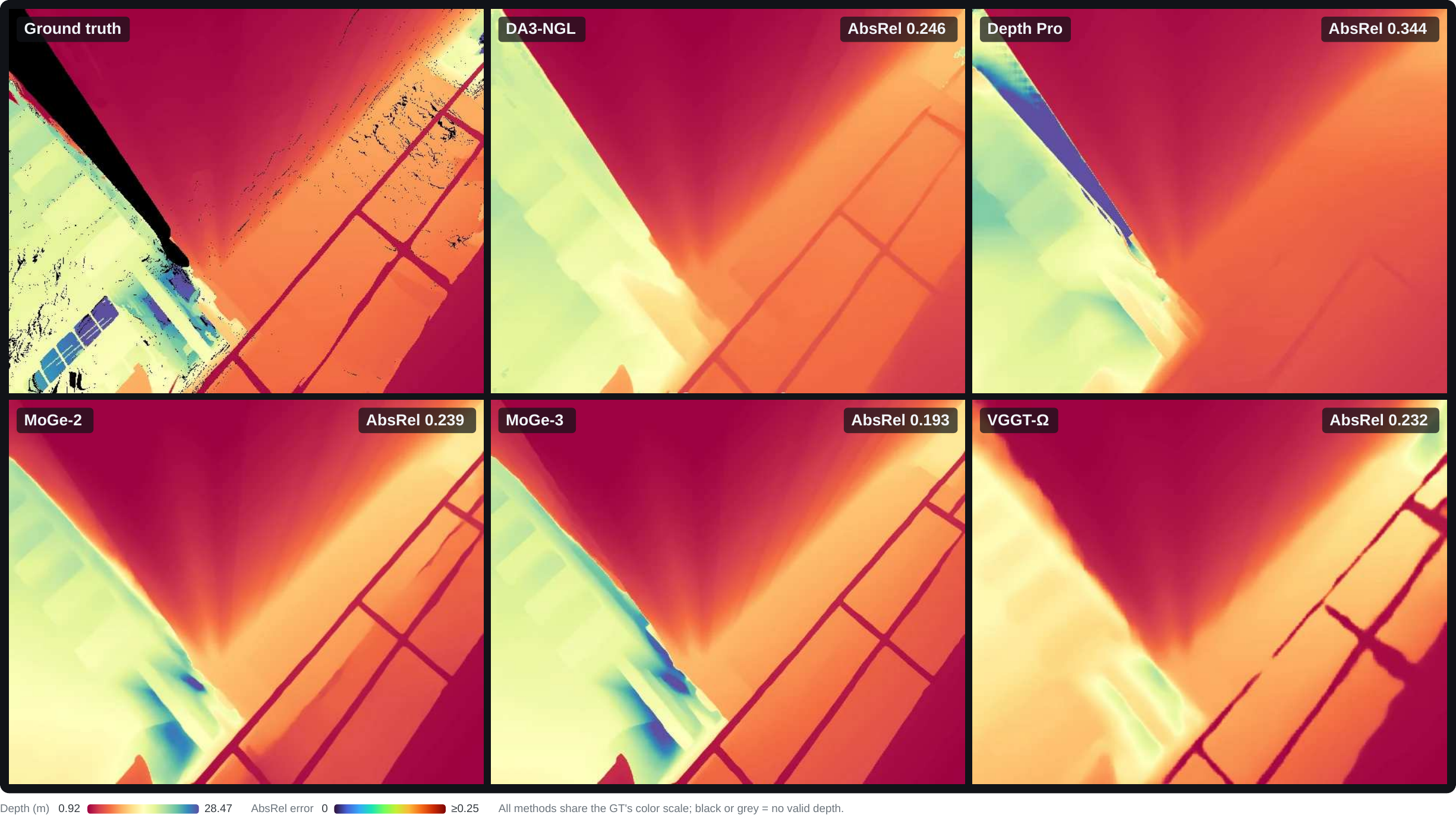}{(c) KingsCrossCanal2, depth}\hspace{0.02\textwidth}
\supqualpanel[0.44\textwidth]{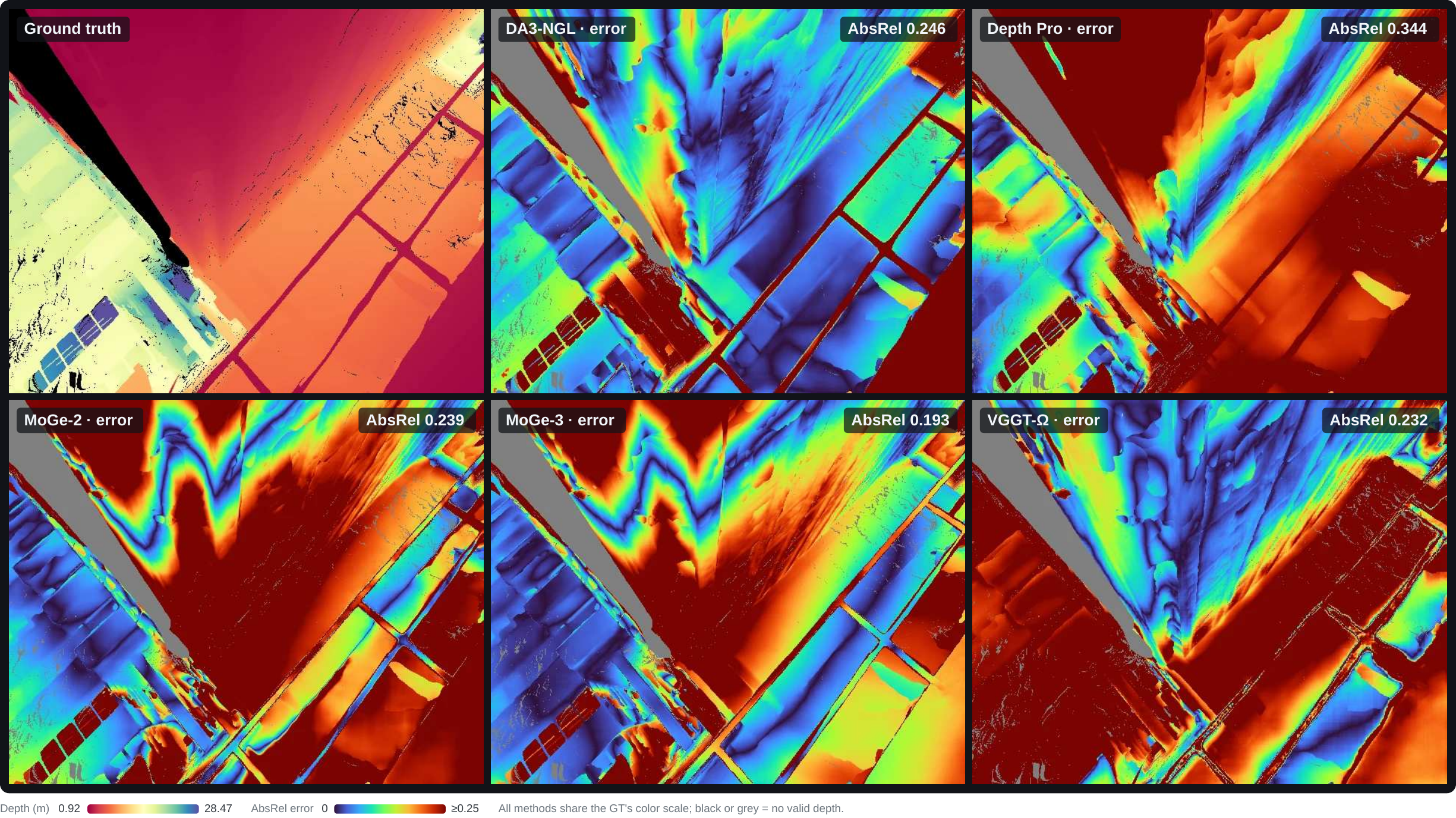}{(d) KingsCrossCanal2, error}
\captionof{figure}{\textbf{MonoDepth predictions and AbsRel error maps on two \oursdata frames.} Depth maps share the color scale of the ground truth; error maps run from 0 (blue) to $\geq$0.25 (red), and grey marks pixels without valid depth. In the garden (a, b), MoGe-2 is most accurate (0.186) and MoGe-3 least (0.271), with its largest errors on the pergola beams; VGGT-$\Omega$ does not separate the beams from the background (inset). On the canal (c, d), MoGe-3 is most accurate (0.193) and Depth Pro least (0.344), with large errors over most of the frame; MoGe-2 and MoGe-3 err in bands on the near surface at the top.}
\label{fig:suppl:depth_sa}
\end{figure*}

On iBims-1 (\cref{fig:suppl:depth_ibims}), the depth specialists are within 0.016--0.034 AbsRel of one another on a simple room, whereas foliage and railings increase every error by a factor of 2.5--4.6; MoGe-3 is among the best on both frames.

On the two \oursdata frames of \cref{fig:suppl:depth_sa}, no model is consistently best: MoGe-2 leads in the garden (0.186) and MoGe-3 on the canal (0.193), although MoGe-3 is least accurate in the garden (0.271). In the garden, the largest errors lie on the thin pergola beams near the camera, which VGGT-$\Omega$ merges with the background.

\begin{figure*}[!tp]
\centering
\includegraphics[width=0.78\textwidth]{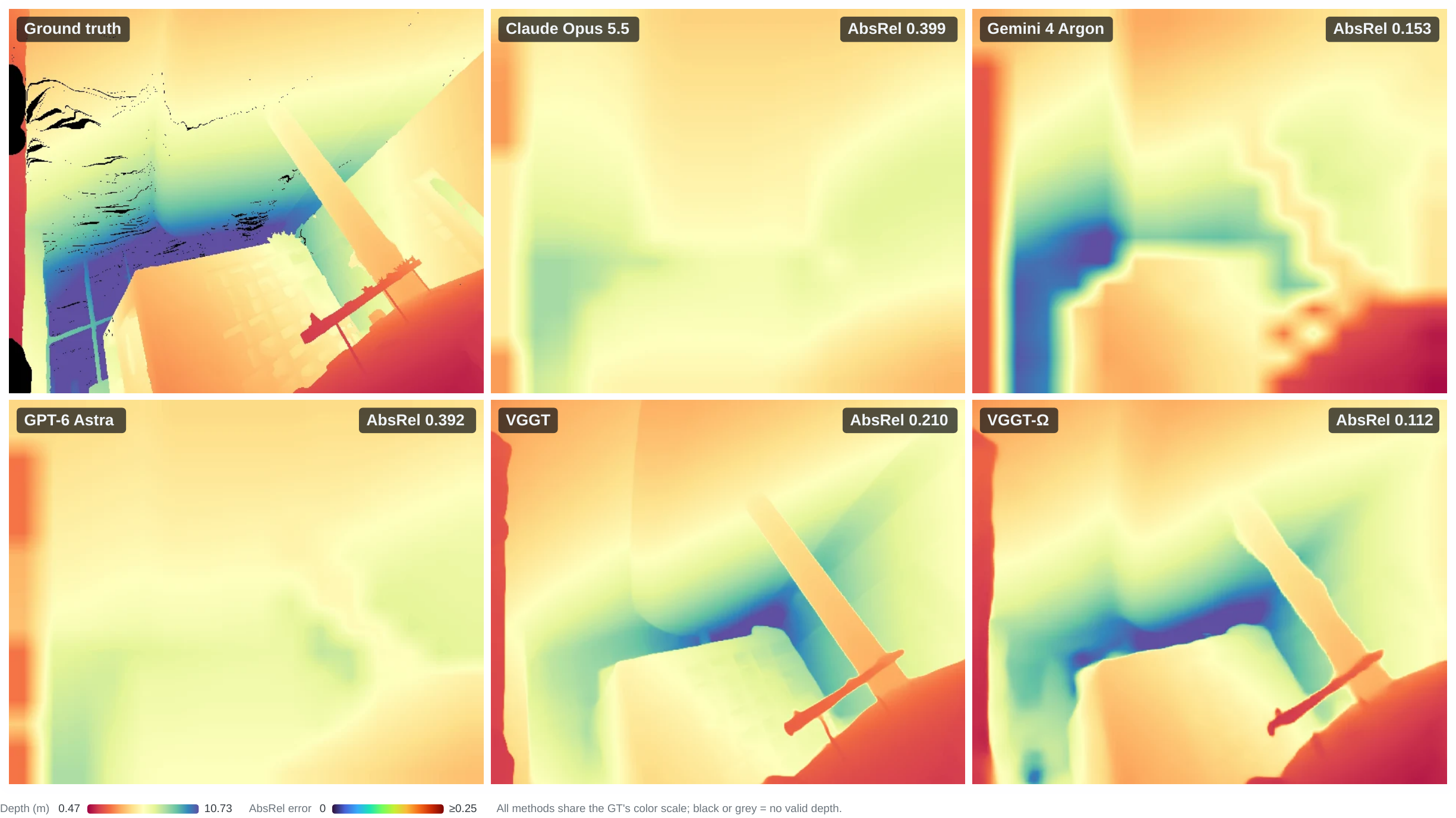}
\captionof{figure}{\textbf{MonoDepth predictions of the commercial VLMs} on a \oursdata frame (CambridgeRalishLibrary), next to VGGT and VGGT-$\Omega$, shown as in \cref{fig:exp:depth}. Claude Opus 5.5 and GPT-6 Astra return nearly flat depth (AbsRel 0.399 and 0.392); Gemini 4 Argon recovers the layout (0.153) and stays behind VGGT-$\Omega$ (0.112). The VLM maps are interpolated from a $16{\times}16$ query grid (\cref{sec:suppl:vlm}) and cannot represent fine structure.}
\label{fig:suppl:depth_vlm}
\end{figure*}

On a \oursdata frame (\cref{fig:suppl:depth_vlm}), Claude Opus 5.5 and GPT-6 Astra produce nearly planar depth, and only Gemini 4 Argon recovers the layout of the corridor (AbsRel 0.153), still behind VGGT-$\Omega$ (0.112).

\suppfigpage{%
\centering
\supqualpanel[0.9\textwidth]{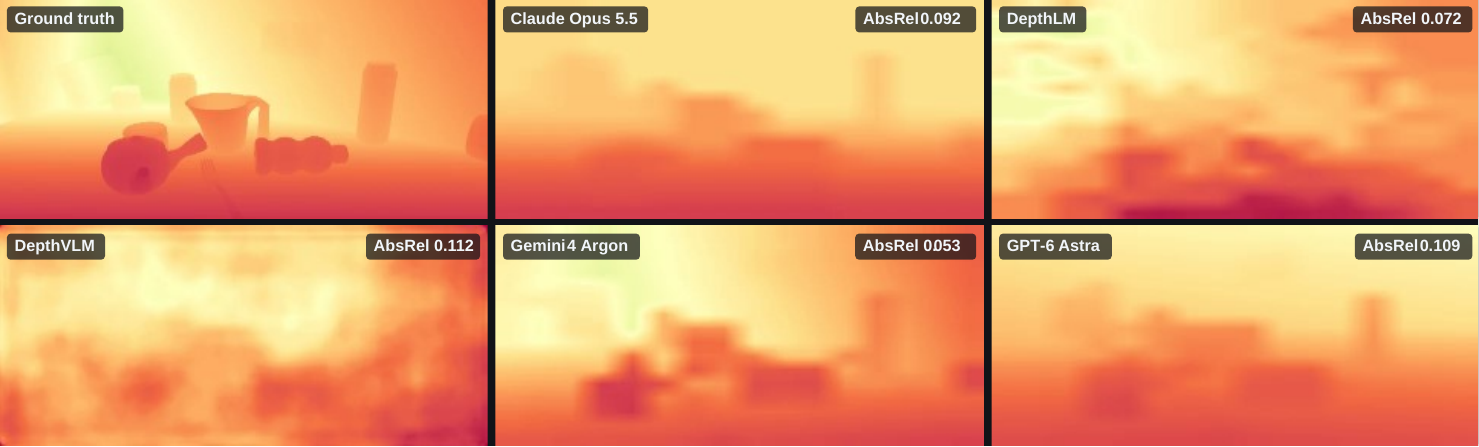}{(a) HAMMER}\\[4pt]
\supqualpanel[0.9\textwidth]{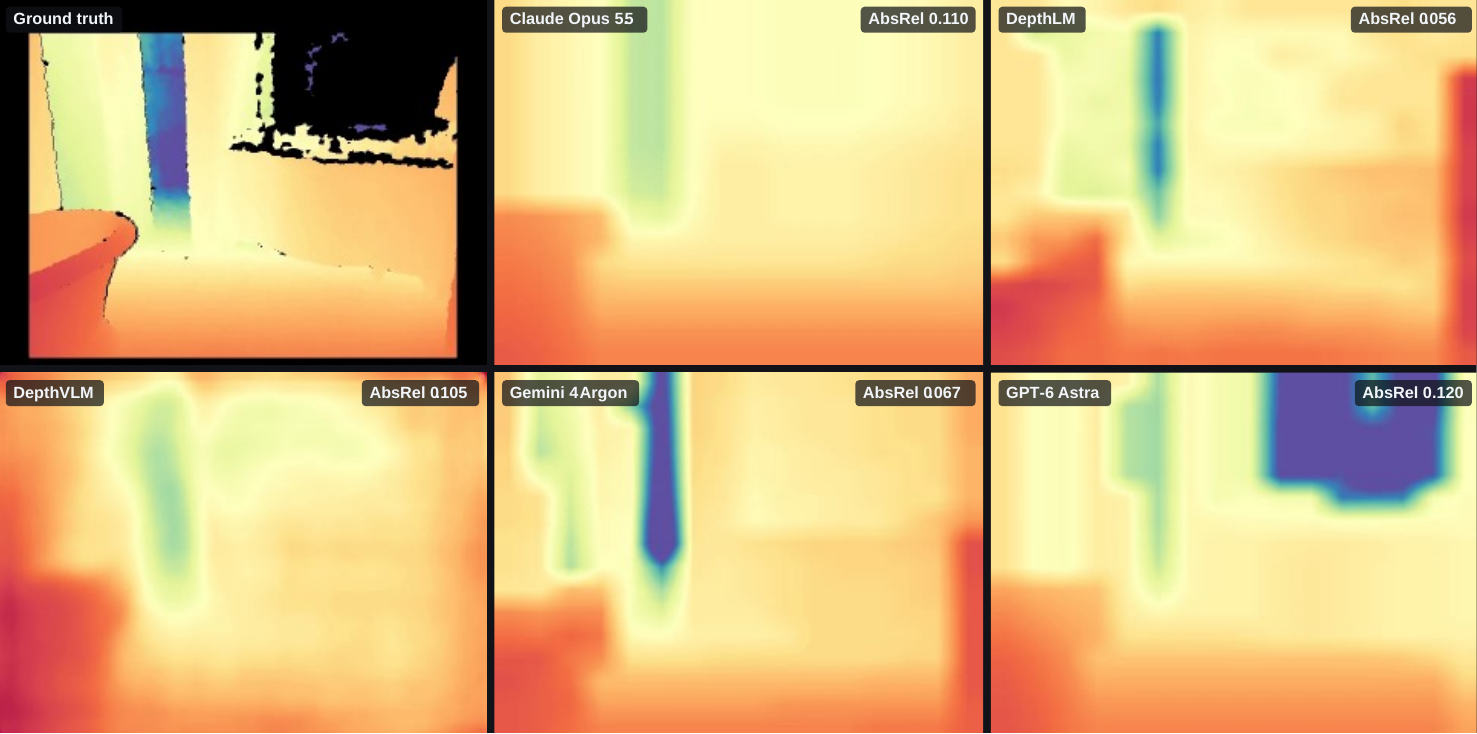}{(b) NYUv2}
\captionof{figure}{\textbf{MonoDepth predictions of the VLMs} on a HAMMER and an NYUv2 frame, shown as in \cref{fig:exp:depth}. All VLMs recover the layout of the scene. The commercial VLMs answer a $16{\times}16$ grid of points (\cref{sec:suppl:vlm}), so their interpolated maps cannot resolve object boundaries by construction. Gemini 4 Argon is best on HAMMER (0.053) and DepthLM on NYUv2 (0.056); Claude Opus 5.5, GPT-6 Astra, and DepthVLM have errors of 0.092--0.120 on both frames.}
\label{fig:suppl:depth_vlm_public}
}

On HAMMER and NYUv2 (\cref{fig:suppl:depth_vlm_public}), all VLMs recover the layout of the scene. The maps of the commercial VLMs are interpolated from a $16{\times}16$ query grid (\cref{sec:suppl:vlm}) and therefore cannot represent object boundaries, so we compare them only on their layout and AbsRel: Gemini 4 Argon is the most accurate VLM on HAMMER and DepthLM on NYUv2.

\suppfigpage{%
\centering
\supqualpanel[0.9\textwidth]{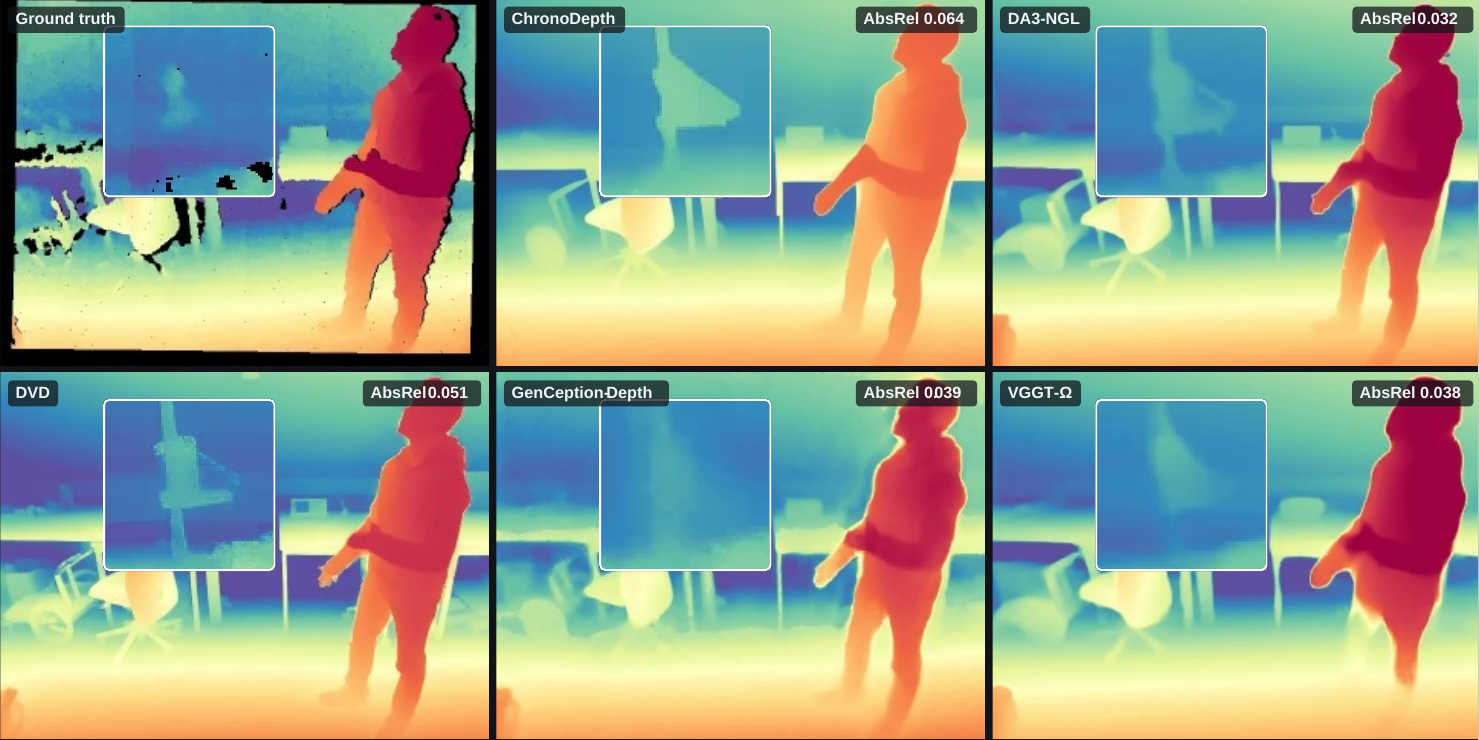}{(a) Bonn, balloon}\\[4pt]
\includegraphics[width=0.55\textwidth]{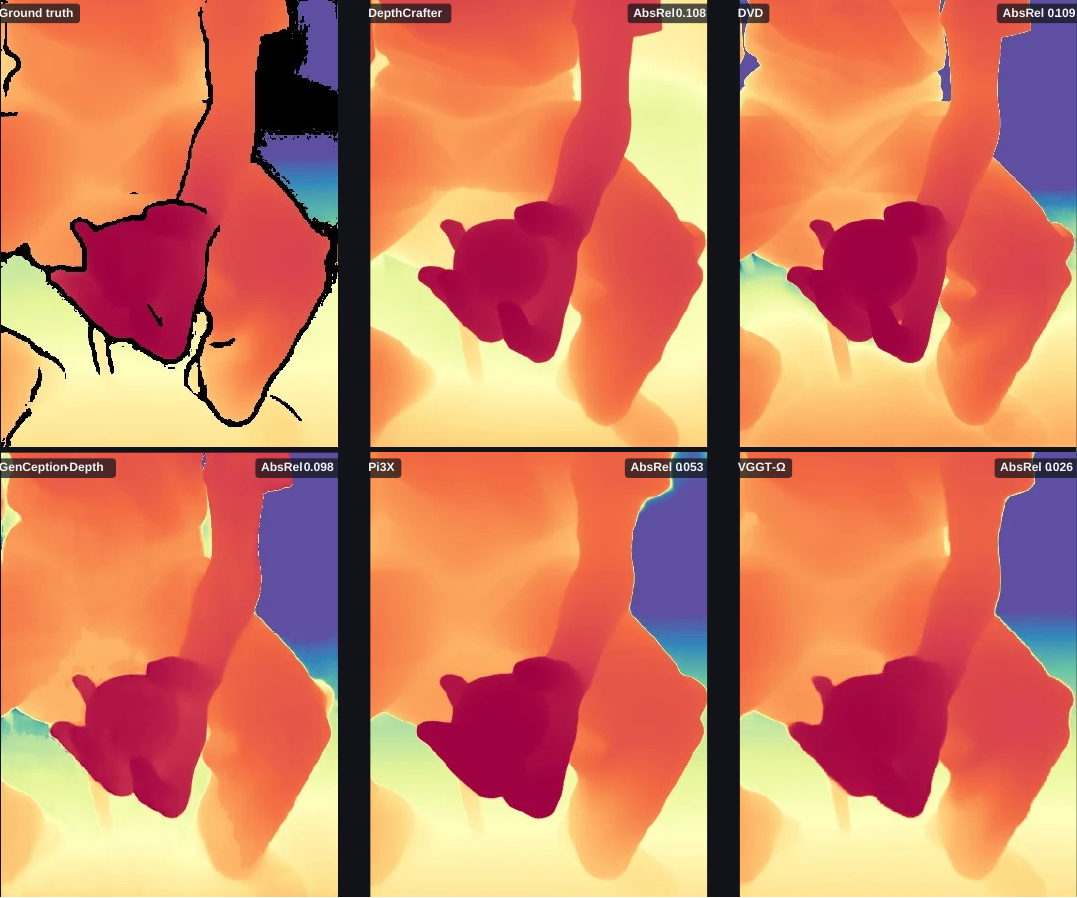}\\[-1pt]{\footnotesize (b) DyCheck, apple}
\captionof{figure}{\textbf{VideoDepth predictions on public sequences}, one frame each, shown as in \cref{fig:exp:depth}. On the dynamic scenes of Bonn (a) and DyCheck (b), the feed-forward 3D/4D models have the lowest errors (DA3-NGL 0.032 and VGGT-$\Omega$ 0.038 on Bonn; VGGT-$\Omega$ 0.026 and $\pi^3$-X 0.053 on DyCheck), below those of the video depth models DVD, ChronoDepth, and DepthCrafter (0.051--0.109).}
\label{fig:suppl:video_depth}
}
\suppfigpage{%
\centering
\supqualpanel[0.9\textwidth]{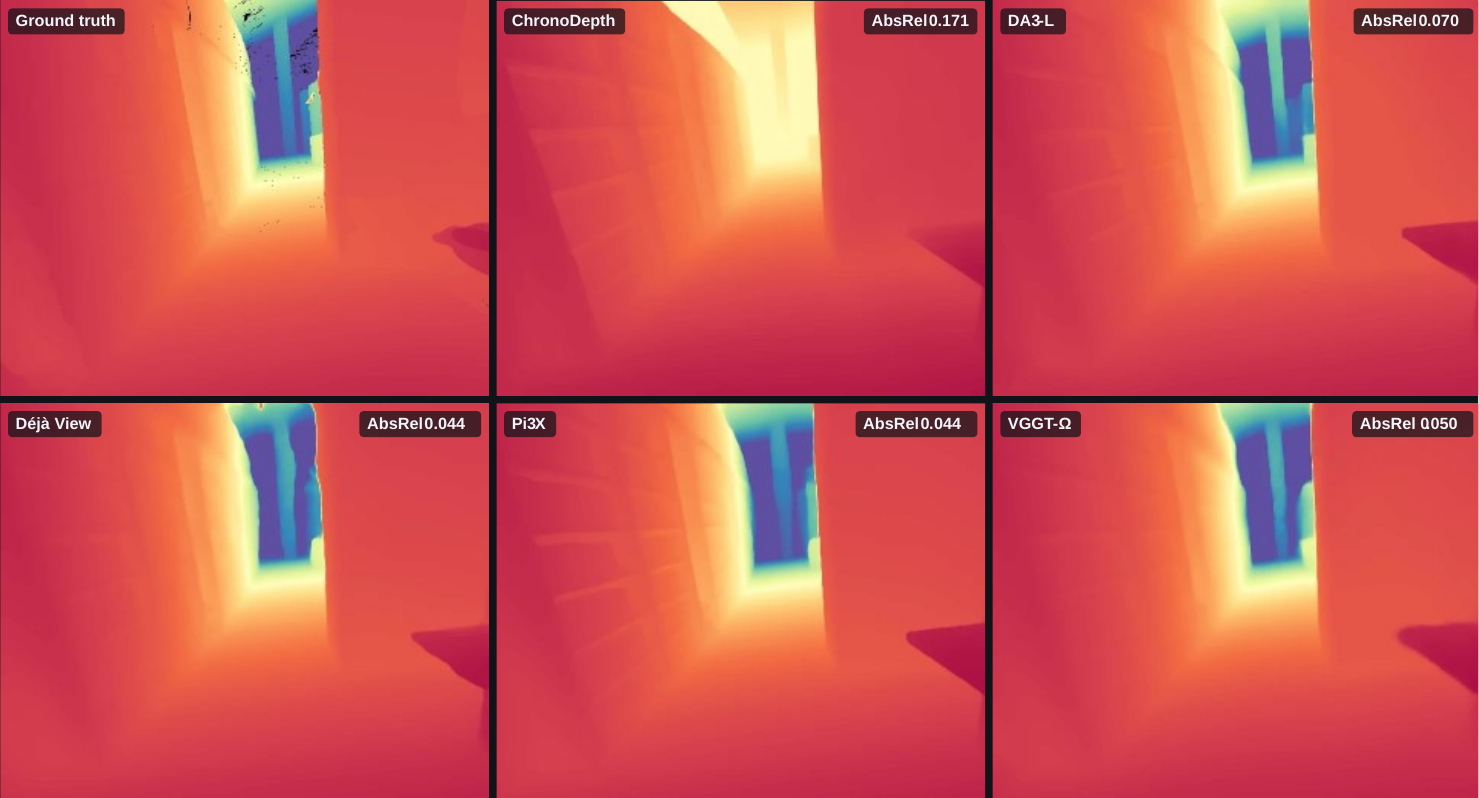}{(a) CambridgeRalishLibrary}\\[4pt]
\supqualpanel[0.9\textwidth]{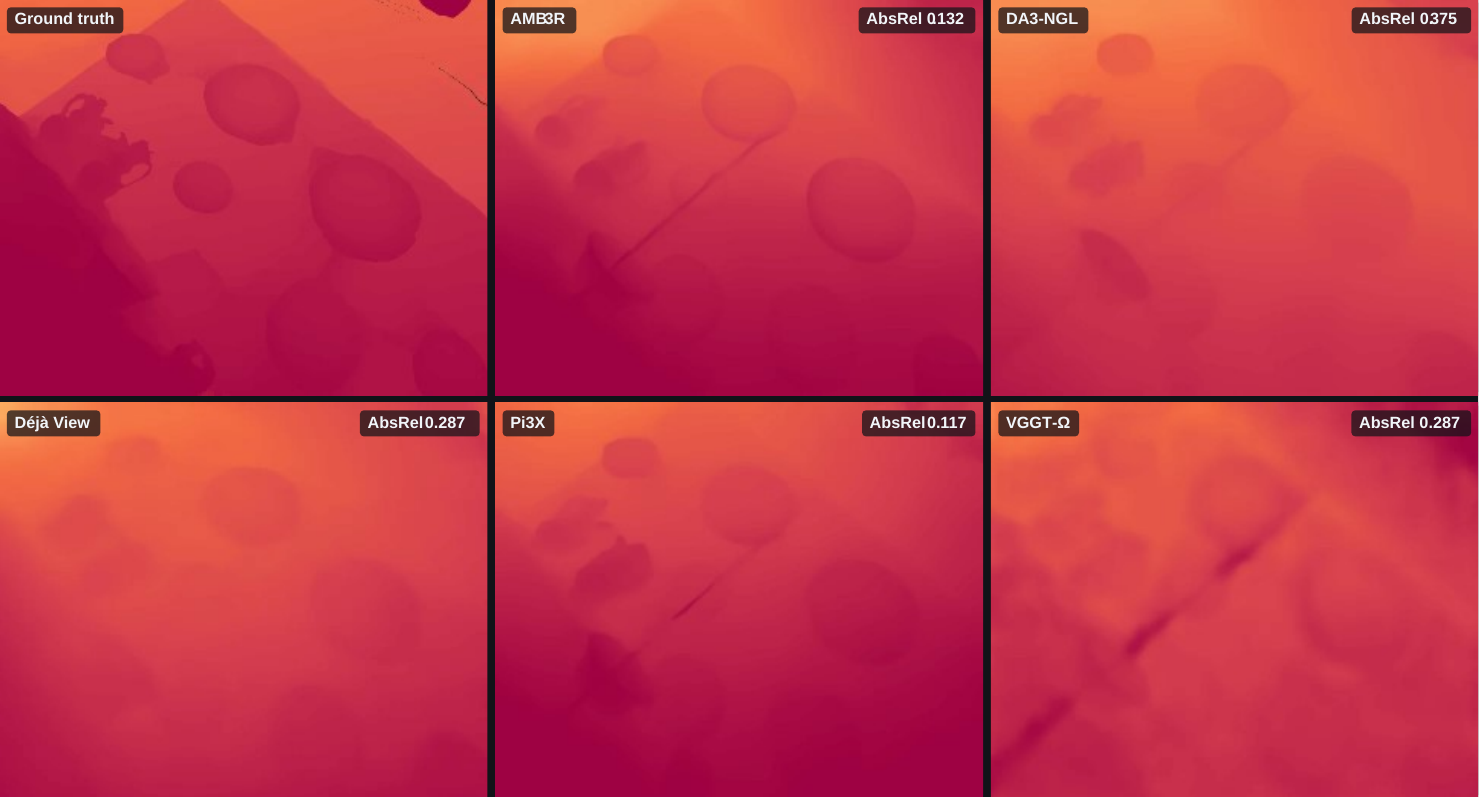}{(b) NationalScotlandMuseum6}
\captionof{figure}{\textbf{VideoDepth predictions on \oursdata sequences}, shown as in \cref{fig:suppl:video_depth}. In the library corridor (a), Déjà View and $\pi^3$-X are best (0.044), while ChronoDepth misses the depth of the far opening (0.171). On the close-up of a nearly planar surface in the museum (b), every model fails to some degree (0.117--0.375): $\pi^3$-X and AMB3R are best, and VGGT-$\Omega$ (0.287) and DA3-NGL (0.375) are worst.}
\label{fig:suppl:video_depth_sa}
}

\paragraph{VideoDepth.}
On the dynamic Bonn and DyCheck sequences (\cref{fig:suppl:video_depth}), the feed-forward 3D/4D models are more accurate than the video depth models DVD, ChronoDepth, and DepthCrafter, consistent with \cref{tab:exp:video}. On \oursdata (\cref{fig:suppl:video_depth_sa}), ChronoDepth misses the far opening of the library corridor, and a close-up of a nearly planar museum surface degrades every model; VGGT-$\Omega$ and DA3-NGL, among the strongest models on public data, are among the weakest on this frame.

\suppfigpage{%
\centering
\supqualpanel[0.9\textwidth]{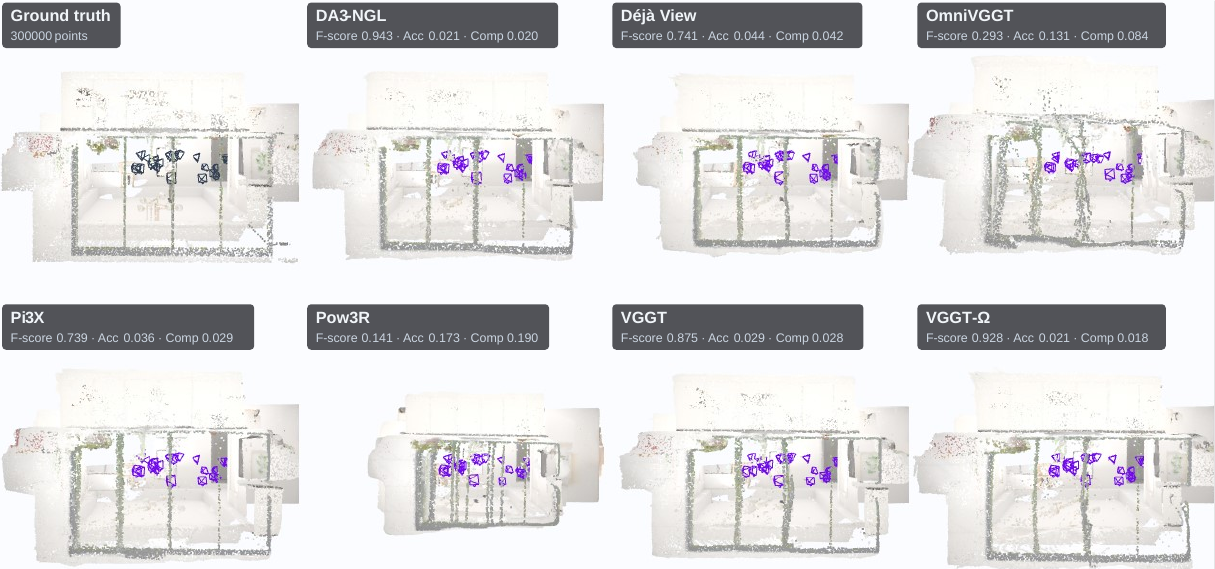}{(a) Reconstructions}\\[4pt]
\supqualpanel[0.9\textwidth]{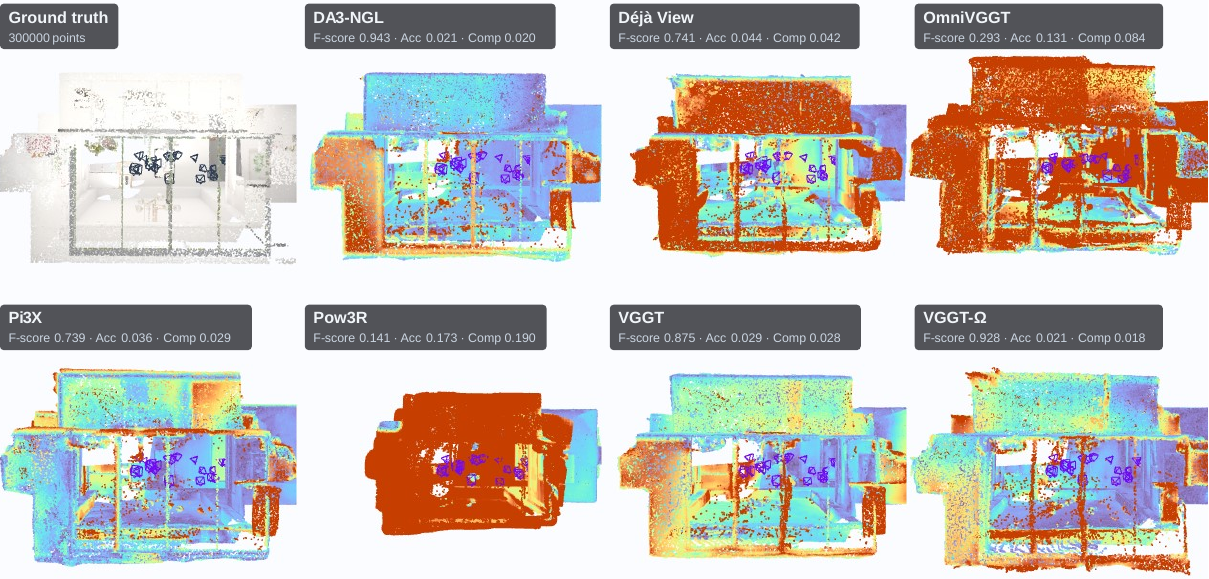}{(b) Point error, blue (small) to red (large)}
\captionof{figure}{\textbf{MVRecon reconstructions on a HiRoom scene}, shown as in \cref{fig:exp:recon}, in color (a) and colored by each point's distance to the ground truth (b). DA3-NGL (F-score 0.943), VGGT-$\Omega$ (0.928), and VGGT (0.875) reconstruct the room almost exactly, whereas the errors of OmniVGGT (0.293) and Pow3R (0.141) spread over whole walls.}
\label{fig:suppl:recon_hiroom}
}

\paragraph{MVRecon.}
On HiRoom (\cref{fig:suppl:recon_hiroom}), DA3-NGL, VGGT-$\Omega$, and VGGT reconstruct the room almost exactly (F-score 0.875--0.943), whereas the errors of OmniVGGT and Pow3R extend over entire walls rather than being confined to object boundaries.

\suppfigpage{%
\centering
\supqualpanel[0.9\textwidth]{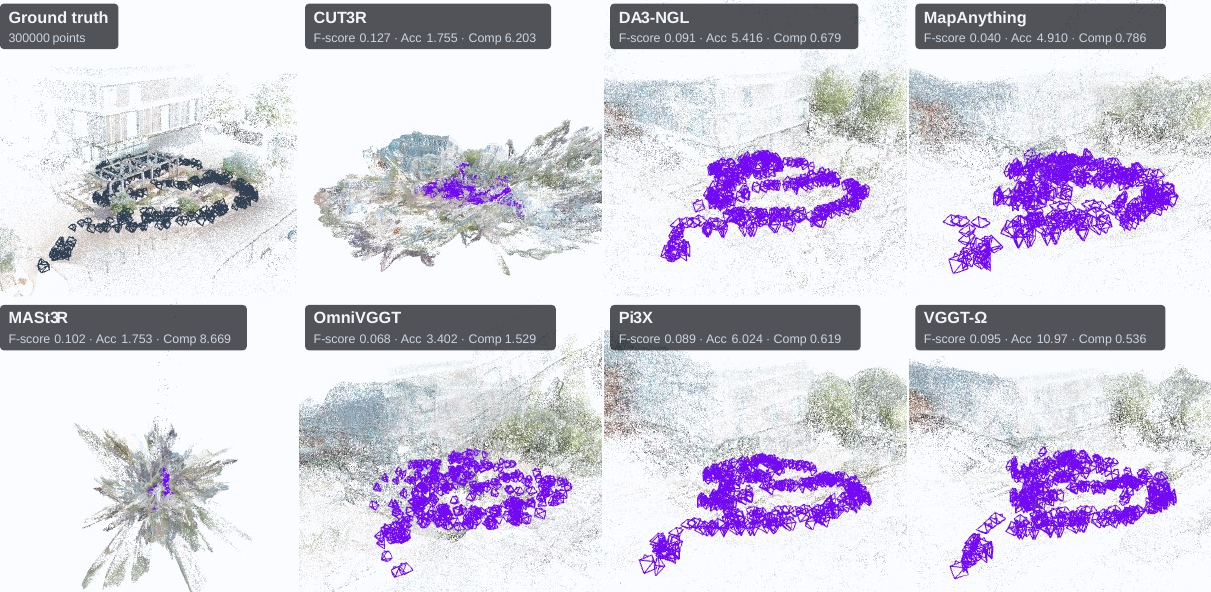}{(a) CambridgeSmallGarden1}\\[4pt]
\supqualpanel[0.9\textwidth]{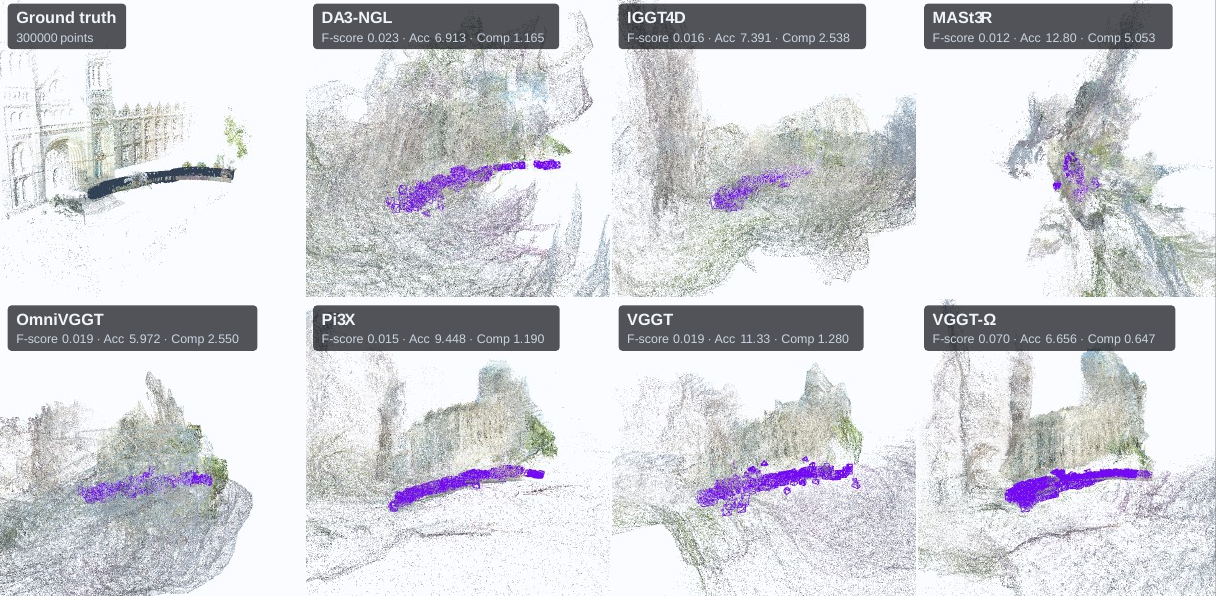}{(b) NationalHistoryMuseum2}
\captionof{figure}{\textbf{MVRecon failure cases on \oursdata}, shown as in \cref{fig:exp:recon}. On these large scenes, no model exceeds an F-score of 0.35 at $\tau=25$\,cm: the best are $\pi^3$-X in the garden (a, 0.349) and VGGT-$\Omega$ in the museum (b, 0.159).}
\label{fig:suppl:recon_fail}
}
\suppfigpage{%
\centering
\supqualpanel[0.9\textwidth]{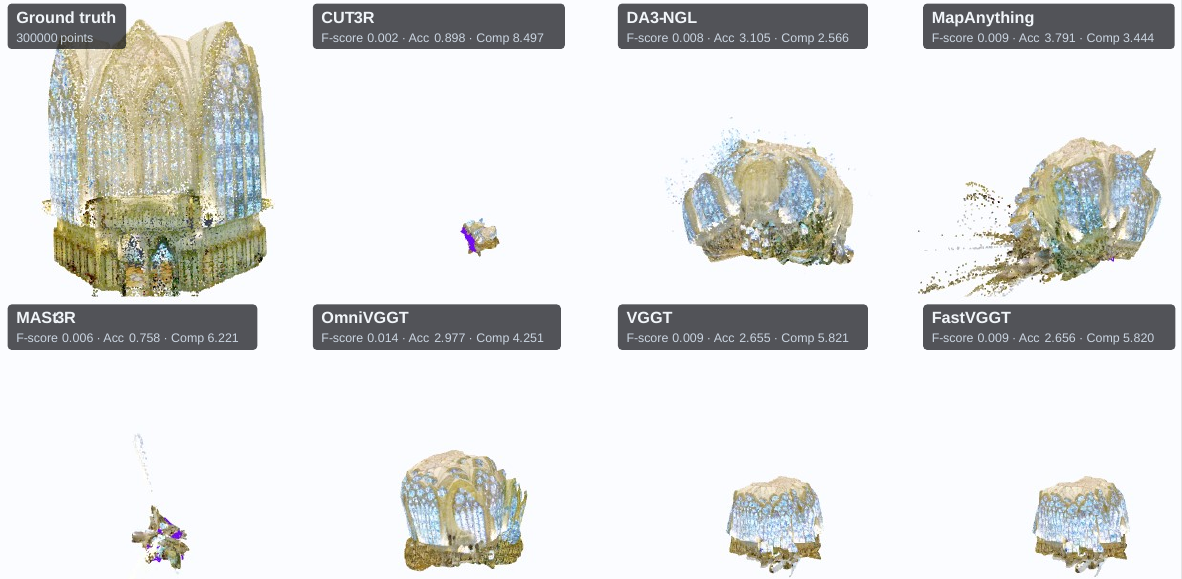}{YorminsterTheChapteraofRoof}
\captionof{figure}{\textbf{A third MVRecon failure case on \oursdata}, shown as in \cref{fig:suppl:recon_fail}. Most models recover a recognizable shape of the roof, but at the wrong scale and position, so few points fall within $\tau=5$\,cm; the best F-score is 0.124 (VGGT-$\Omega$).}
\label{fig:suppl:recon_fail_roof}
}

On three \oursdata scenes (\cref{fig:suppl:recon_fail,fig:suppl:recon_fail_roof}), no model exceeds an F-score of 0.35. On the roof, most models recover a plausible shape at the wrong scale and position, so few points fall within the 5\,cm threshold.

\suppfigpage{%
\centering
\begin{subfigure}[b]{0.42\textwidth}
\centering
\includegraphics[width=\linewidth]{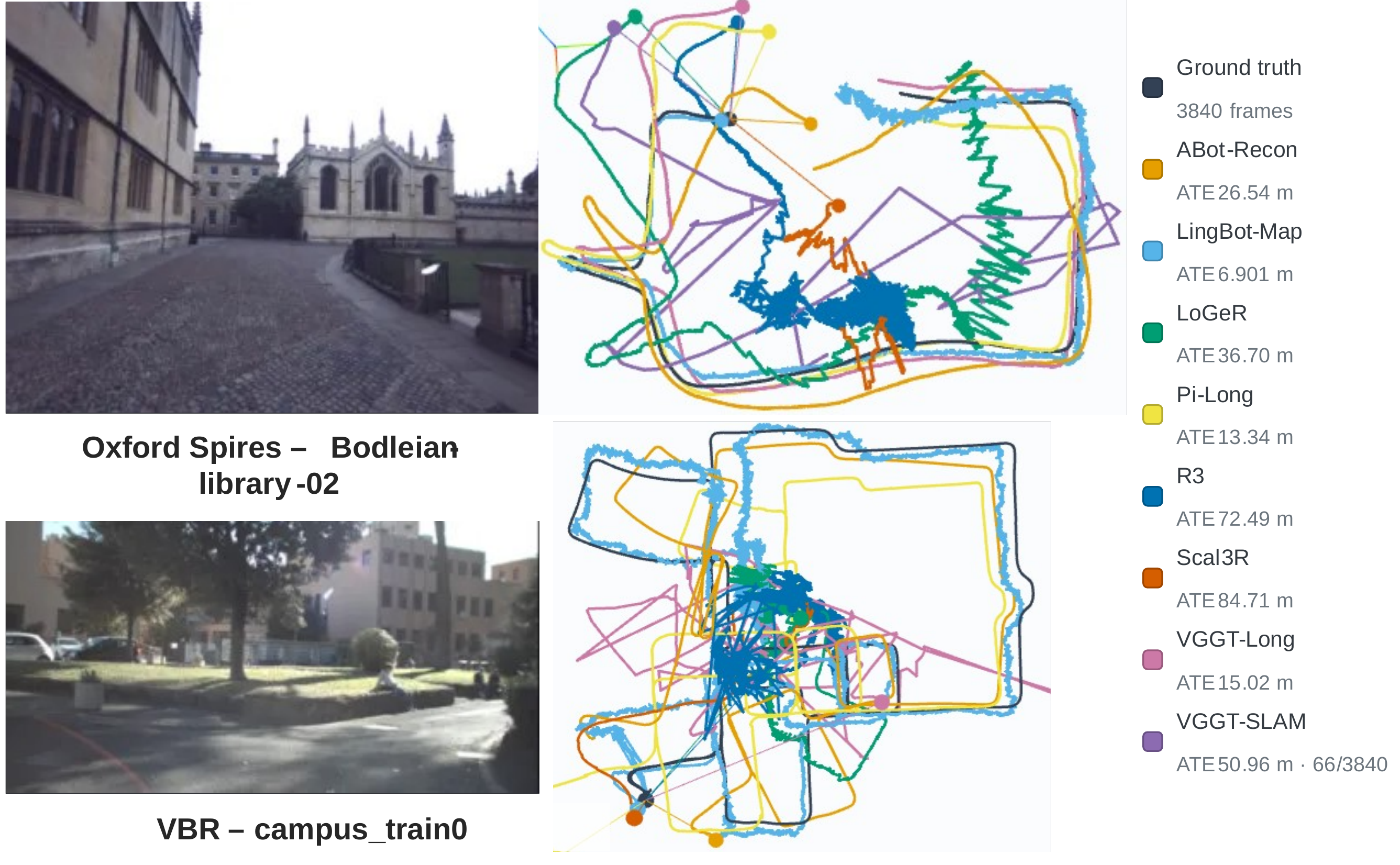}
\caption{Oxford Spires and VBR}\label{fig:exp:slam_traj}
\end{subfigure}\\[4pt]
\begin{subfigure}[b]{0.75\textwidth}
\centering
\includegraphics[width=\linewidth]{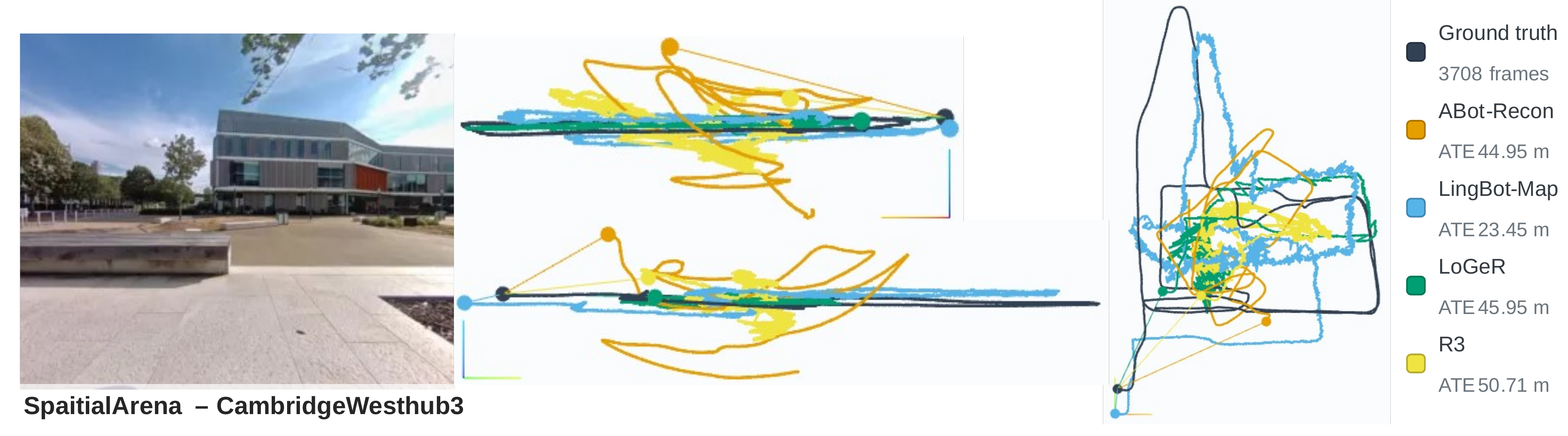}
\caption{\oursdata}\label{fig:suppl:slam_traj}
\end{subfigure}
\captionof{figure}{\textbf{Estimated trajectories on the large SLAM range}, aligned to the ground truth (black) as for ATE. The legend gives each model's ATE; in (a), VGGT-SLAM tracks only 66 of the 3,840 frames.}
}

\paragraph{SLAM.}
On the large range (\cref{fig:exp:slam_traj,fig:suppl:slam_traj}), LingBot-Map stays closest to the ground truth on the sequences shown, whereas other systems drift or collapse onto a small region; VGGT-SLAM tracks only 66 of the 3,840 frames of the Oxford Spires sequence.

\suppfigpage{%
\centering
\supqualpanel[\textwidth]{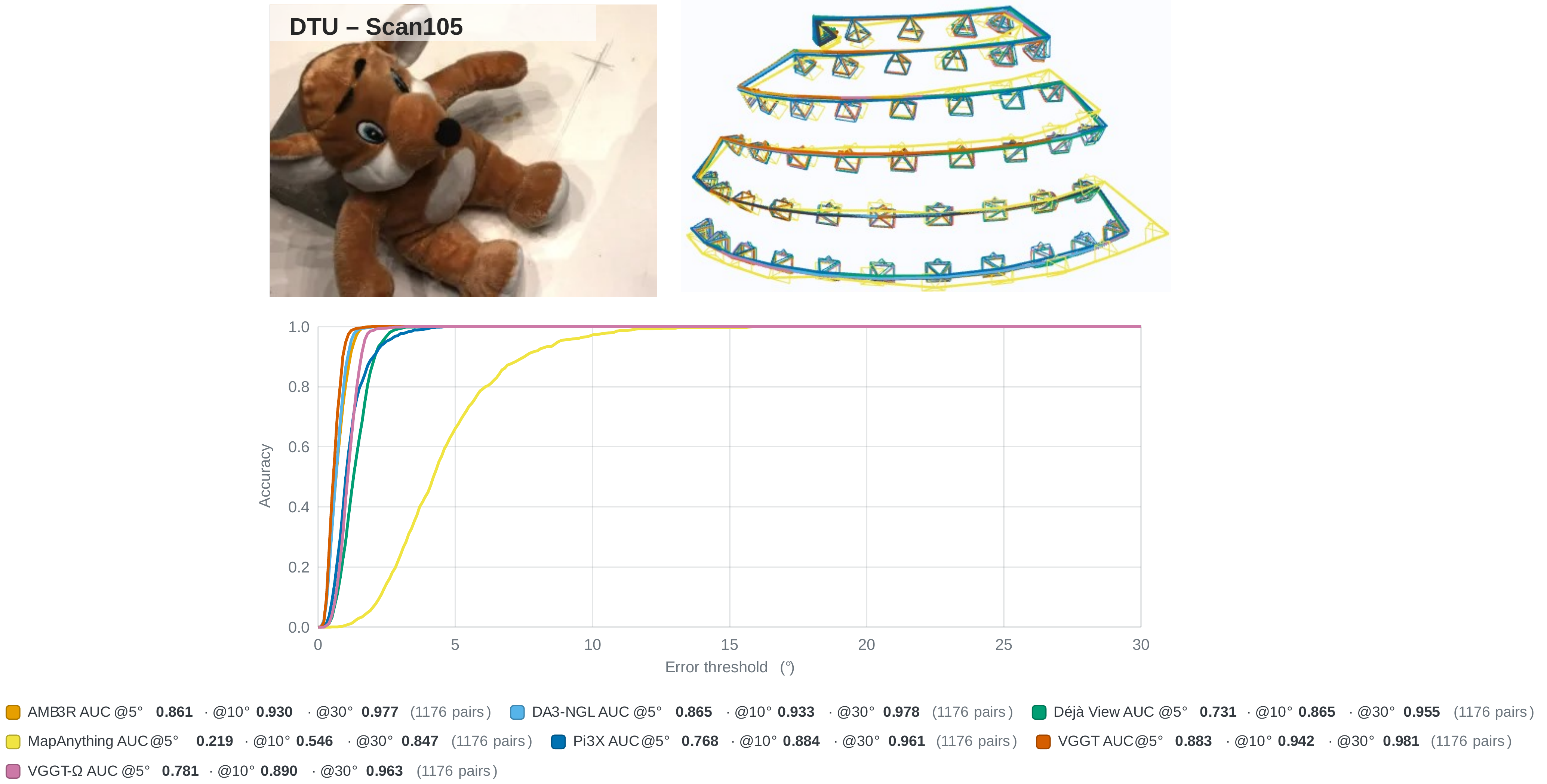}{(a) DTU, scan105 (49 images)}\\[4pt]
\supqualpanel[\textwidth]{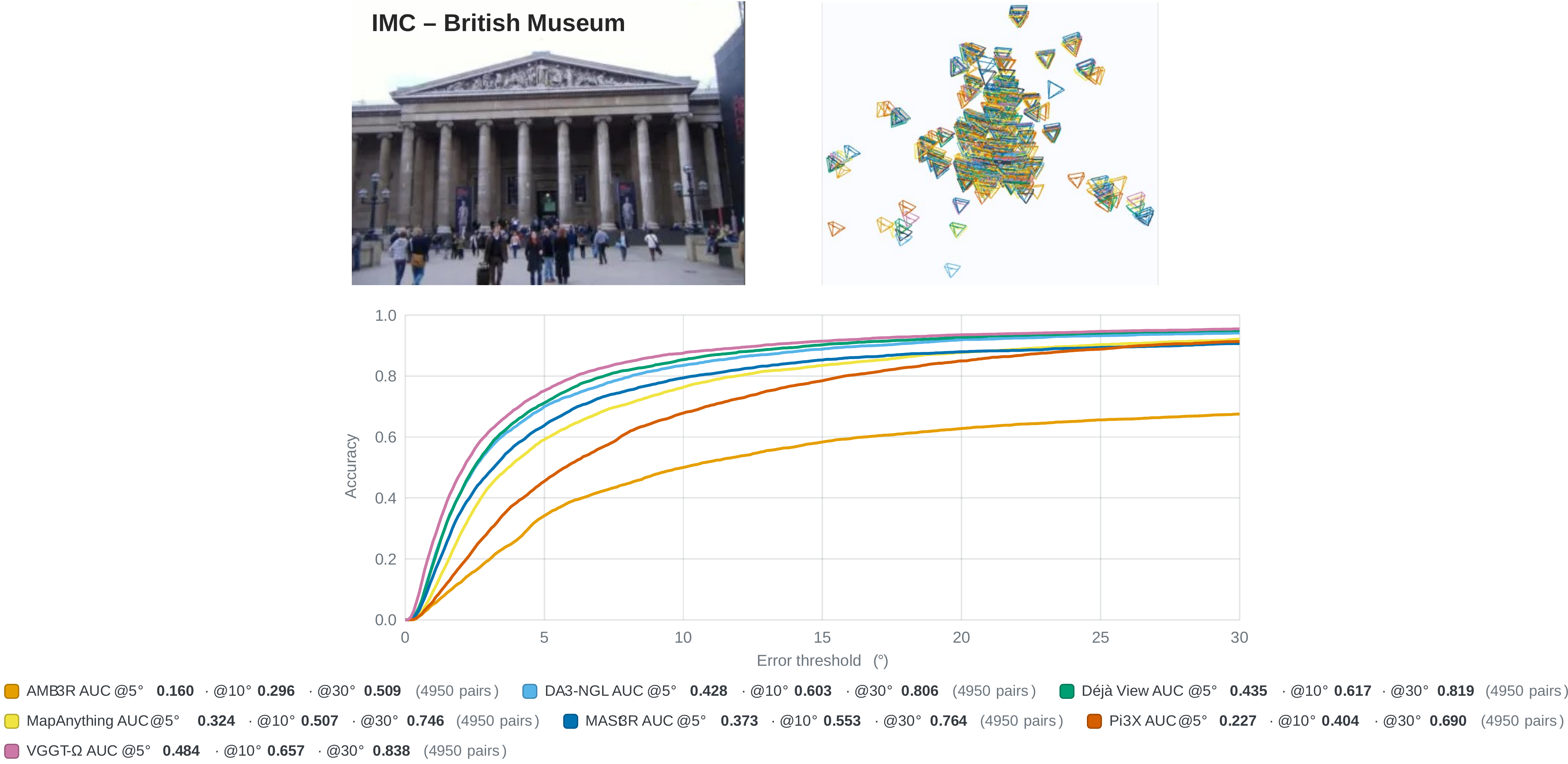}{(b) IMC Phototourism, British Museum (100 images)}
\captionof{figure}{\textbf{SfM on two public scenes.} Each panel shows an input image, the predicted cameras after $\mathrm{Sim}(3)$ alignment (one color per model, ground truth in dark gray), and the fraction of image pairs whose pose error $\max(e^R,e^t)$ is below each threshold, with the AUC at $5^\circ$, $10^\circ$, and $30^\circ$ in the legend. On the object-centric DTU scene (a), all models but MapAnything reach an AUC@$5^\circ$ of 0.73--0.88 (VGGT best, 0.883); MapAnything (0.219) shifts the cameras of the outer rings. On the unordered internet photos of IMC (b), VGGT-$\Omega$ is best (0.484), followed by Déjà View (0.435) and DA3-NGL (0.428), whereas AMB3R (0.160) leaves about a third of the pairs above $30^\circ$.}
\label{fig:suppl:sfm_public}
}

\paragraph{SfM.}
On the object-centric DTU scene (\cref{fig:suppl:sfm_public}), all models except MapAnything reach an AUC@$5^\circ$ of 0.73--0.88. On the unordered internet photos of IMC, the accuracy is lower and more variable: VGGT-$\Omega$ is best (0.484), whereas AMB3R leaves about a third of the image pairs above $30^\circ$.

\suppfigpage{%
\centering
\includegraphics[width=0.78\textwidth]{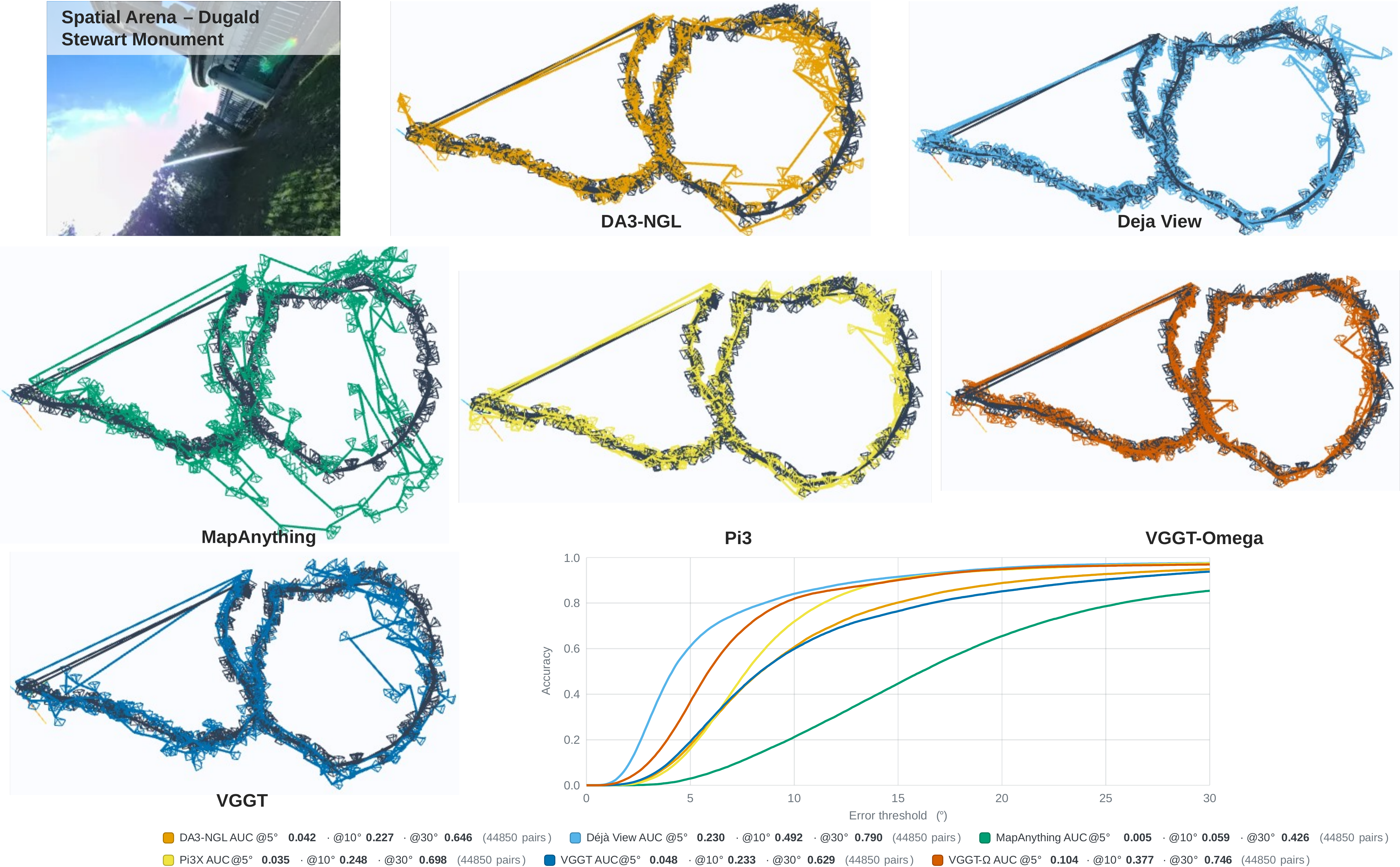}
\captionof{figure}{\textbf{SfM on a \oursdata scene} (Dugald Stewart Monument, 300 images), shown as in \cref{fig:suppl:sfm_public}, with the cameras of each model in a separate panel. All models recover the shape of the loop, but their cameras scatter around it, and the AUC@$5^\circ$ is far below that on the public scenes: Déjà View is best (0.230), followed by VGGT-$\Omega$ (0.104); VGGT, DA3-NGL, and $\pi^3$-X stay below 0.05, and the cameras of MapAnything (0.005) leave the loop.}
\label{fig:suppl:sfm_monument}
}

On the \oursdata walk around the Dugald Stewart Monument (\cref{fig:suppl:sfm_monument}), all models recover the loop, but their cameras scatter around it, and the AUC@$5^\circ$ falls far below its values on the public scenes. D\'ej\`a View is best (0.230), and the cameras of MapAnything leave the loop.

\suppfigpage{%
\centering
\includegraphics[width=0.72\textwidth]{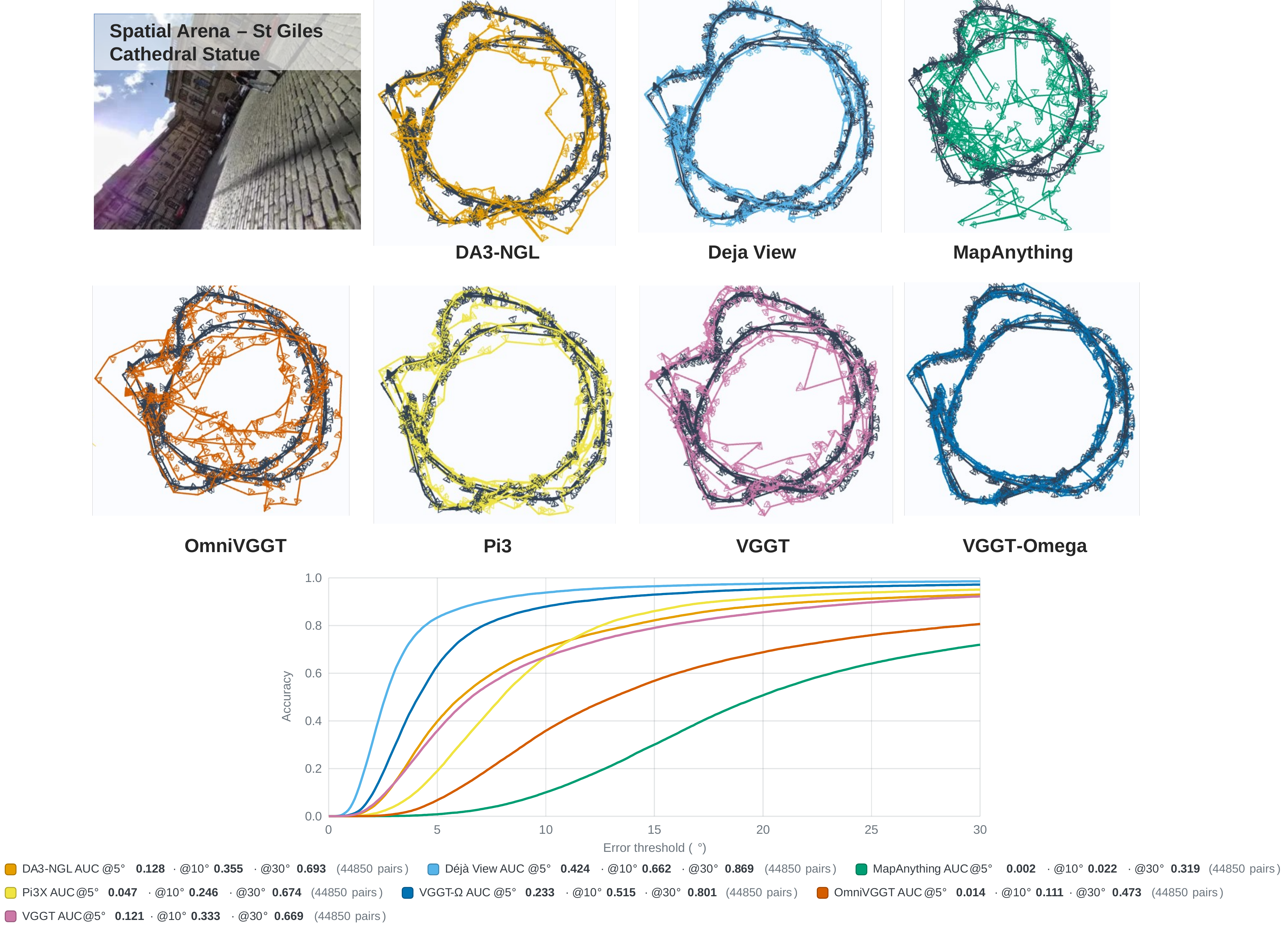}
\captionof{figure}{\textbf{SfM on a second \oursdata scene} (St Giles' Cathedral statue, 300 images), shown as in \cref{fig:suppl:sfm_monument}. Déjà View (AUC@$5^\circ$ 0.424) and VGGT-$\Omega$ (0.233) follow the repeated loops around the statue; DA3-NGL and VGGT (0.128 and 0.121) add spurious jumps across the loop, and the cameras of OmniVGGT (0.014) and MapAnything (0.002) scatter over its inside.}
\label{fig:suppl:sfm_stgiles}
}

The repeated loops around the St Giles' Cathedral statue (\cref{fig:suppl:sfm_stgiles}) separate the models further. D\'ej\`a View (AUC@$5^\circ$ 0.424) and VGGT-$\Omega$ (0.233) follow the loops, DA3-NGL and VGGT introduce spurious jumps across them, and the cameras of OmniVGGT and MapAnything scatter over the interior.

\suppsection{metrics}

\paragraph{Valid depth ranges.}
In MonoDepth and VideoDepth, we resize the prediction bilinearly to ground-truth resolution, align it as in \cref{sec:method:depthalign} on the set $\Omega$ of valid ground-truth pixels, and clip the aligned depth $\tilde{D}$ to the valid range of each dataset. In MonoDepth, these ranges are $[10^{-3},10]$\,m for NYUv2, $[10^{-5},80]$\,m for KITTI, $(3,150)$\,m for Synth4K, $(10^{-3},400)$\,m for AerialMetric, and $[0.1,100]$\,m for \oursdata; HAMMER, iBims-1, and Spring have only a lower bound. DIODE, whose indoor and outdoor scenes span very different depth ranges, has no range filter on the ground truth; instead, we clip each aligned prediction to $[10^{-3},\max_{p\in\Omega}D(p)]$ of its own image. In VideoDepth, we cap valid ground truth at $70$\,m (Sintel), $10$\,m (Bonn), $300$\,m (KITTI), $100$\,m (\oursdata, with a lower bound of $0.1$\,m), and $10$ units (DyCheck, in the release's depth units), and clip aligned predictions to the same range.

\paragraph{Depth metrics.}
Let $\langle x\rangle=\frac{1}{|\Omega|}\sum_{p\in\Omega}x(p)$ denote the mean over the valid pixels and $d=\log\tilde{D}-\log D$ the log error. We report
\begin{align}
  \mathrm{AbsRel} &= \big\langle |\tilde{D}-D|/D \big\rangle, \\
  \mathrm{SqRel} &= \big\langle (\tilde{D}-D)^2/D \big\rangle, \\
  \mathrm{RMSE} &= \big\langle (\tilde{D}-D)^2 \big\rangle^{1/2}, \\
  \mathrm{RMSE}_{\log} &= \big\langle d^2 \big\rangle^{1/2}, \\
  \log_{10} &= \big\langle |\log_{10}\tilde{D}-\log_{10}D| \big\rangle, \\
  \mathrm{iRMSE} &= \big\langle (1/\tilde{D}-1/D)^2 \big\rangle^{1/2}, \\
  \mathrm{SILog} &= 100\,\big(\langle d^2\rangle-\langle d\rangle^2\big)^{1/2}, \\
  \delta_{\tau} &= \big\langle \mathbb{1}\big[\max(\tilde{D}/D,\,D/\tilde{D})<\tau\big] \big\rangle.
\end{align}
RMSE on Spring is in its relative units. The normal error is the mean angle between surface normals computed from the back-projected aligned and ground-truth depth. We never use model confidence maps to discard pixels. In MonoDepth, we average the metrics uniformly over images; an image with fewer than two valid pixels cannot determine a scale and shift, and we exclude it for all methods alike. In VideoDepth, we compute the metrics over all valid pixels of a sequence and average them uniformly over sequences.

\paragraph{MVRecon.}
After aligning the predicted poses with $\mathrm{Sim}(3)$ and rescaling the depths by its scale $s$, we fuse all views into a TSDF volume~\cite{DBLP:conf/siggraph/CurlessL96}. We extract a mesh, sample $10^6$ points $\hat{\mathcal{P}}$ from it, and compare them with points $\mathcal{P}$ sampled from the reference surface after voxel down-sampling:
\begin{align}
  \mathrm{Acc} &= \tfrac{1}{|\hat{\mathcal{P}}|}\textstyle\sum_{\hat{\mathbf{x}}\in\hat{\mathcal{P}}}\min_{\mathbf{x}\in\mathcal{P}}\|\hat{\mathbf{x}}-\mathbf{x}\|, \\
  \mathrm{Comp} &= \tfrac{1}{|\mathcal{P}|}\textstyle\sum_{\mathbf{x}\in\mathcal{P}}\min_{\hat{\mathbf{x}}\in\hat{\mathcal{P}}}\|\mathbf{x}-\hat{\mathbf{x}}\|,
\end{align}
and $\mathrm{Overall}=(\mathrm{Acc}+\mathrm{Comp})/2$. Precision and recall are the fractions of predicted and reference points within a distance $\tau$ of the other set, and $F_\tau$ is their harmonic mean. We follow DA3-BENCH~\cite{DBLP:conf/iclr/LinCLCLZPGZSFK26} for the TSDF and evaluation parameters. For indoor scenes, we use a voxel size of $4/512$\,m, truncation $0.04$\,m, and $\tau=5$\,cm. For ScanNet++, we use a voxel size of $0.02$\,m, truncation $0.15$\,m, a maximum depth of $5$\,m, and $\tau=5$\,cm, and crop the prediction to the bounding box of the ground truth. For ETH3D and Tanks\&Temples, we scale all lengths by $5\times$ ($\tau=25$\,cm). For \oursdata, we do not fuse a TSDF volume: we back-project the rescaled depths of all views, up to a maximum depth of $100$\,m, directly into one point cloud, voxel down-sample it (keeping the first valid point per voxel), and use it as $\hat{\mathcal{P}}$, with $\tau=5$\,cm and a voxel size of $4/512$\,m indoors and $\tau=25$\,cm and $20/512$\,m outdoors. On Tanks\&Temples, we first map the fused points into the laser-scan frame with the official per-scene registration and crop them, together with the reference, to the official bounding volume; our scores are thus in laser-scan meters at one threshold for all scenes and are not comparable to the per-scene thresholds of the official leaderboard. On Tanks\&Temples, DTU, and NRGBD, we further refine the pose-based alignment with a scale-aware point-to-point ICP~\cite{DBLP:journals/pami/BeslM92} on the fused points. For DTU, we use the official MVSNet-style evaluation~\cite{DBLP:conf/eccv/YaoLLFQ18} with observation and plane masks and report Acc, Comp, and Overall. We average all metrics over scenes. The toolbox also offers a \emph{posed} variant that fuses the views with ground-truth poses, which separates depth quality from pose error.

\paragraph{SfM.}
For every pair of matched images $(i,j)$, we compare the predicted relative motion $(\hat{R}_{ij},\hat{\mathbf{t}}_{ij})$ with the ground truth $(R_{ij},\mathbf{t}_{ij})$. We compute the relative rotation error $e^{R}_{ij}=\angle(\hat{R}_{ij}^{\top}R_{ij})$, the translation direction error $e^{t}_{ij}=\angle(\hat{\mathbf{t}}_{ij},\mathbf{t}_{ij})$, and the sign-agnostic translation axis error $e^{\bar{t}}_{ij}=\min(e^{t}_{ij},180^\circ-e^{t}_{ij})$. RRA@$\tau$ and RTA@$\tau$ are the fractions of pairs with $e^R<\tau$ and $e^{\bar{t}}<\tau$ for $\tau\in\{5^\circ,15^\circ,30^\circ\}$. AUC@$\tau$ is the normalized area under the recall curve of $\max(e^R,e^t)$ up to $\tau\in\{1^\circ,3^\circ,5^\circ,10^\circ,20^\circ,30^\circ\}$~\cite{DBLP:conf/cvpr/WangCKV0N25}, and mAA@$30^\circ$ averages the accuracy over integer thresholds of $\max(e^R,e^{\bar{t}})$ up to $30^\circ$. The ATE is the RMSE of camera centers after the $\mathrm{Sim}(3)$ alignment of \cref{sec:method:align}. We use all pairs without subsampling, and average the relative metrics with pair weights and the ATE with image weights.

\paragraph{SLAM.}
We compute the ATE (RMSE, mean, and median) over every returned frame, including frames that the alignment down-weights. We measure the RPE in translation and rotation between consecutive inputs and over one second of recording. We compute the pose AUC of the SfM task over pairs of returned frames, using 1,000 frames evenly spaced in time on longer sequences; both of its angles are invariant to a similarity transform, so it does not depend on the alignment. Drift is the sum of the one-second translational RPE divided by the ground-truth path length over the same windows; as a ratio of sums, near-static segments cannot inflate it. Completeness is the fraction of input frames with a returned pose. Even with metric depth priors, most methods return poses whose scale differs from the ground truth, so a fixed-scale fit would mostly measure this offset; we therefore score after the $\mathrm{Sim}(3)$ alignment. To expose metric scale, we also store for each case the ATE after a fixed-scale ($\mathrm{SE}(3)$) alignment and the fitted $\mathrm{Sim}(3)$ scale. We rank the methods within each track and dataset by the mean of their ATE rank and drift rank; we report a method that fails on any sequence with its coverage but do not rank it.

\suppsection{impl}

\subsection{Depth alignment}
\label{sec:suppl:depth}

\subsubsection{Implementation}
\label{sec:suppl:impl:depth}

\Cref{alg:depth_align} details the depth alignment of \cref{sec:method:depthalign}. We apply it once per image in MonoDepth ($N=1$) and once per sequence in VideoDepth.

\paragraph{Non-positive predictions.}
We raise a non-positive predicted depth to $10^{-5}$ before the fit, so that its inverse is defined, and use raw disparities as they are.

\paragraph{Solver.}
We start the iteratively reweighted least squares from the least-squares solution over all pixels of $\Omega$. Each iteration assigns pixel weights $w=\min\!\big(1,\delta/|r|\big)$, with $r$ the normalized residual, and solves the weighted affine fit in closed form. We stop when $a$ and $b$ change by less than $10^{-7}$ relative to their magnitude, or after 100 iterations. A constant prediction, \eg, a VLM that answers one depth everywhere, has no defined scale; instead of failing the case, we score it with the best single value, a Huber location fit with $a=0$. We standardize the predictions internally for numerical conditioning only, and record for every fit its iterations, stopping reason, objective, and the number of non-positive predictions raised to $10^{-5}$, so that convergence can be audited per case.

\paragraph{Back to depth.}
For depth outputs, we clip the aligned depth $a^*\hat{D}+b^*$ to the dataset's valid range. For disparity outputs, the floor $\epsilon$ in $\tilde{D}=1/\max\!\big(a^*\hat{q}+b^*,\,\epsilon\big)$ maps near-zero or negative aligned disparity to a far but finite depth instead of producing invalid values, so every pixel of $\Omega$ remains scored. Following MoGe~\cite{DBLP:conf/cvpr/WangXDX00Y25}, we set $\epsilon=1/\max_{p\in\Omega}D(p)$ in MonoDepth, so no aligned pixel lies beyond the farthest ground truth of its image. In VideoDepth, we use a fixed $\epsilon=10^{-3}$, after which the dataset's depth cap applies. We apply each fit to the full depth map, which also yields the aligned normals used in MonoDepth (\cref{sec:evaltasks}).

\begin{algorithm}[t]
\caption{Ground-truth-only native-space depth alignment}\label{alg:depth_align}
\small
\KwIn{Predictions $\{\hat{D}_i\}_{i=1}^{N}$ or disparities $\{\hat{q}_i\}_{i=1}^{N}$, ground truth $\{D_i\}$, masks $\{M_i\}$, valid range $(d_{\min},d_{\max})$; $\delta{=}0.05$, $\tau{=}10^{-7}$, $K{=}100$}
\KwOut{Aligned depths $\{\tilde{D}_i\}$, scored pixels $\{\Omega_i\}$}
\For{$i=1,\dots,N$}{
  $\Omega_i\leftarrow\{p : D_i(p)\ \text{finite},\ d_{\min}<D_i(p)<d_{\max},\ M_i(p)\}$\;
  \lIf{$\hat{D}_i$ is non-finite on $\Omega_i$}{\textbf{fail} the case}
  $\hat{D}_i\leftarrow\max(\hat{D}_i,10^{-5})$ \tcp*{depth output only}
}
\leIf{depth output}{$\mathbf{x}\leftarrow[\hat{D}_i(p)]$;\ $\mathbf{y}\leftarrow[D_i(p)]$}{$\mathbf{x}\leftarrow[\hat{q}_i(p)]$;\ $\mathbf{y}\leftarrow[1/D_i(p)]$}\tcp*{$i,\ p\in\Omega_i$}
$m\leftarrow\mathrm{median}(\mathbf{y})$;\quad $\mathbf{y}\leftarrow\mathbf{y}/m$;\quad $\mathbf{z}\leftarrow(\mathbf{x}-\mu_{\mathbf{x}})/\sigma_{\mathbf{x}}$ \tcp*{scale-free}
$(\alpha,\beta)\leftarrow\arg\min_{\alpha,\beta}\|\alpha\mathbf{z}+\beta-\mathbf{y}\|_2^2$ \tcp*{L2 initialization}
\For{$k=1,\dots,K$}{
  $\mathbf{r}\leftarrow\alpha\mathbf{z}+\beta-\mathbf{y}$;\quad $w_j\leftarrow\min(1,\delta/|r_j|)$ \tcp*{Huber weights}
  $(\alpha',\beta')\leftarrow\arg\min_{\alpha,\beta}\sum_j w_j(\alpha z_j+\beta-y_j)^2$ \tcp*{closed form}
  $\Delta\leftarrow\max(|\alpha'-\alpha|,|\beta'-\beta|)$;\quad $(\alpha,\beta)\leftarrow(\alpha',\beta')$\;
  \lIf{$\Delta\le\tau\,(1+\max(|\alpha|,|\beta|))$}{\textbf{break}}
}
$a\leftarrow m\,\alpha/\sigma_{\mathbf{x}}$;\quad $b\leftarrow m\,\beta-a\,\mu_{\mathbf{x}}$ \tcp*{undo normalization}
$\epsilon\leftarrow 1/\max_{i,\,p\in\Omega_i}D_i(p)$ (MonoDepth) or $10^{-3}$ (VideoDepth)\;
\For{$i=1,\dots,N$}{
  \leIf{depth output}{$\tilde{D}_i\leftarrow\mathrm{clip}\big(a\,\hat{D}_i+b,\ d_{\min},\ d_{\max}\big)$}{$\tilde{D}_i\leftarrow\mathrm{clip}\big(1/\max(a\,\hat{q}_i+b,\ \epsilon),\ d_{\min},\ d_{\max}\big)$}
}
\Return{$\{\tilde{D}_i\}$, $\{\Omega_i\}$}\;
\end{algorithm}

\begin{table}[t]
\centering
\footnotesize
\setlength{\tabcolsep}{2.5pt}
\caption{\textbf{Depth alignment choices on MonoDepth.} We score the same predictions after six per-image affine alignments: fit space (depth or inverse depth) $\times$ estimator ($\ell_1$ with fixed Adam steps, least squares, or Huber IRLS). All use the ground-truth-only pixel set, and we compute the metrics in depth. AbsRel ($\times 100$) and $\delta_{1.25}$ (\%) are equal-weighted means over 15 subsets (NYUv2, KITTI, ETH3D, DIODE, HAMMER, iBims-1, Spring, five Synth4K games, three AerialMetric scenes; 5,205 images, at most 100 per Synth4K and AerialMetric scene). Bold: best per model. Gray: our default, Huber IRLS in each model's native output space (depth for VGGT-$\Omega$, DA3-G, and Marigold; inverse depth for Lotus-D).}
\label{tab:depth_align}
\adjustbox{max width=\linewidth}{%
\begin{tabular}{llcccccccc}
\toprule
 & & \multicolumn{2}{c}{VGGT-$\Omega$} & \multicolumn{2}{c}{DA3-G} & \multicolumn{2}{c}{Lotus-D} & \multicolumn{2}{c}{Marigold v1.1} \\
\cmidrule(lr){3-4}\cmidrule(lr){5-6}\cmidrule(lr){7-8}\cmidrule(lr){9-10}
Space & Estimator & AbsRel$\downarrow$ & $\delta_{1.25}\uparrow$ & AbsRel$\downarrow$ & $\delta_{1.25}\uparrow$ & AbsRel$\downarrow$ & $\delta_{1.25}\uparrow$ & AbsRel$\downarrow$ & $\delta_{1.25}\uparrow$ \\
\midrule
\multirow{3}{*}{Depth} & $\ell_1$ + Adam & 7.15 & 94.0 & 8.69 & 91.1 & 81.58 & 56.9 & 31.61 & 55.0 \\
 & Least squares & 10.40 & 90.2 & 12.49 & 87.8 & 45.67 & 51.8 & 19.28 & 78.4 \\
 & Huber IRLS & \cellcolor{gray!15}6.56 & \cellcolor{gray!15}\textbf{94.9} & \cellcolor{gray!15}8.04 & \cellcolor{gray!15}92.6 & 26.44 & 64.4 & \cellcolor{gray!15}\textbf{12.67} & \cellcolor{gray!15}\textbf{86.8} \\
\midrule
\multirow{3}{*}{Inv.\ depth} & $\ell_1$ + Adam & 6.45 & 94.8 & 7.79 & 92.0 & 19.29 & 75.7 & 34.10 & 62.2 \\
 & Least squares & 6.81 & 93.5 & 8.21 & 91.3 & 10.32 & 88.8 & 27.62 & 56.5 \\
 & Huber IRLS & \textbf{6.42} & 94.8 & \textbf{7.51} & \textbf{92.6} & \cellcolor{gray!15}\textbf{9.55} & \cellcolor{gray!15}\textbf{90.0} & 25.82 & 66.5 \\
\bottomrule
\end{tabular}}
\end{table}

\paragraph{Comparison of alignment choices.}
To measure how much the alignment alone changes a score, we re-score identical MonoDepth predictions under six alignments: we fit in depth or in inverse depth and estimate the affine parameters by $\ell_1$ with a fixed number of Adam steps (as in CUT3R and $\pi^3$~\cite{DBLP:conf/cvpr/WangZHEK25,DBLP:conf/iclr/WangZZCZLCPSH26}), by least squares (as in Marigold~\cite{DBLP:conf/cvpr/KeOHMDS24}), or by our Huber IRLS. All six share the ground-truth-only pixel set and the metric computation, which is always in depth. We evaluate four models with different output parameterizations: two multi-view geometry models, VGGT-$\Omega$~\cite{DBLP:conf/cvpr/WangCZKSLBNVR26} and DA3-G~\cite{DBLP:conf/iclr/LinCLCLZPGZSFK26}, which predict depth up to scale; Lotus-D~\cite{DBLP:conf/iclr/HeL0LLZZLC25}, which predicts affine-invariant disparity; and Marigold v1.1~\cite{DBLP:conf/cvpr/KeOHMDS24}, which predicts affine-invariant depth. The study covers 15 MonoDepth subsets and 5,205 images (\cref{tab:depth_align}); we give per-subset results, including RMSE, in \cref{sec:suppl:depth_align}.
We make three observations. First, Huber IRLS is the best estimator in every one of the eight (model, space) combinations, for both AbsRel and $\delta_{1.25}$. Least squares is pulled by outlier pixels; in depth space it raises the AbsRel of VGGT-$\Omega$ and DA3-G by 55--59\% over Huber IRLS. The fixed-step $\ell_1$ fit is unreliable when the prediction is far from the ground-truth scale, as for the normalized outputs of Lotus-D and Marigold: its AbsRel reaches $0.82$ for Lotus-D in depth space. Second, for the multi-view models the space of the fit hardly matters once the estimator is robust (AbsRel within $0.006$, $\delta_{1.25}$ within $0.1$ points). Third, for the relative-depth models the space decides the result: the AbsRel of Lotus-D nearly triples when we fit its disparity in depth space ($0.264$ against $0.096$; $\delta_{1.25}$ of $64.4\%$ against $90.0\%$), and Marigold v1.1 behaves in the opposite way, with an AbsRel twice as high in inverse depth ($0.258$ against $0.127$; $\delta_{1.25}$ of $66.5\%$ against $86.8\%$). An affine fit is therefore best suited to the space in which a model's output is affine-invariant. This motivates our default: Huber IRLS in the native output space that each model declares, which is the best alignment for Lotus-D and Marigold and, for VGGT-$\Omega$ and DA3-G, matches the best $\delta_{1.25}$ and is within $0.006$ of the best AbsRel. We treat the space as a property of the model's output, declared once in the registry, not as a per-dataset tuning choice.

\subsubsection{Alignment choices per subset}
\label{sec:suppl:depth_align}

\Cref{tab:suppl:depth_align_a,tab:suppl:depth_align_b,tab:suppl:depth_align_c} break down the comparison of \cref{tab:depth_align} by subset and add RMSE. We compute each model's predictions once and re-score them under six per-image affine alignments: we fit in depth (Depth) or inverse depth (Inv.), with $\ell_1$ and a fixed number of Adam steps, least squares (Least sq.), or Huber IRLS (\cref{sec:method:depthalign}). All six use the ground-truth-only pixel set, and we compute every metric in depth. The 15 subsets contain 5,205 scored images per configuration, with at most 100 images per Synth4K game and AerialMetric scene. $^\dagger$Spring has relative ground truth, so its RMSE is in relative units; we exclude it from the aggregate ranking (\cref{tab:leaderboard}). $^\ast$One Cyberpunk image has fewer than two valid ground-truth pixels, and we skip it for every configuration.

The per-subset results confirm our conclusions of \cref{sec:suppl:impl:depth}. Fitting in the space in which the output is affine-invariant wins on every subset: Huber IRLS in depth space gives Marigold v1.1 a lower AbsRel than in inverse depth on all 15 subsets, and Huber IRLS in inverse depth beats it in depth space for Lotus-D on all 15 subsets. Within a fit space, Huber IRLS has the lowest AbsRel in 92 of the 120 (subset, model, space) cases and a $\delta_{1.25}$ at least as high as $\ell_1$ with Adam in 105 of them. Where the $\ell_1$ fit has the lower AbsRel, the margin is usually negligible (median $0.0007$); its few large margins all come from Lotus-D fitted in depth space on Synth4K, a space in which both estimators fail. In return, the $\ell_1$ fit fails badly in some cases where Huber IRLS does not, \eg, Lotus-D in depth space on HAMMER, iBims-1, and ETH3D (AbsRel higher by $3.20$, $2.20$, and $1.18$). Least squares is never the best estimator by AbsRel in depth space, although, as expected from its objective, it has the lowest RMSE there in 52 of the 60 (subset, model) cases. For VGGT-$\Omega$ and DA3-G, depth and inverse-depth Huber IRLS differ by less than $0.008$ AbsRel on all subsets except Synth4K Cyberpunk ($0.018$ for DA3-G) and Spring ($0.051$ for DA3-G).

\input{tables/suppl_depth_align_a}

\input{tables/suppl_depth_align_b}

\input{tables/suppl_depth_align_c}

\subsection{Trajectory alignment}
\label{sec:suppl:traj}
\input{tables/align_synthetic}
\input{tables/align_worst}
\input{tables/align_collinear}
\input{tables/align_real}

\subsubsection{Implementation}
\label{sec:suppl:impl:traj}

\Cref{alg:traj_align} details the trajectory alignment of \cref{sec:method:align}. All three stages minimize \cref{eq:pose_cost} with Levenberg--Marquardt. The alignment only determines how we weight the frames in the fit; we compute all metrics on every matched frame.

\paragraph{RANSAC initialization.}
We draw minimal samples of two frames, fit $S$ on each by minimizing \cref{eq:pose_cost} over the sample with $w_i=\sigma_R=\sigma_t=1$, and keep the hypothesis with the smallest median squared center residual (LMedS~\cite{rousseeuw1984least}; 25 iterations, fixed seed). We then refit it on the frames whose center residual lies within $2.5\hat{\sigma}$, where $\hat{\sigma}=1.4826\,(1+\tfrac{5}{n-2})\sqrt{\mathrm{med}\,\|\mathbf{r}^{t}_i\|^2}$. This stage escapes jumps and gross outliers.

\paragraph{Robust refit on all frames.}
A minimal-sample fit still depends on which sample wins. This matters under smooth drift, where no subset of frames is exactly right. Starting from the RANSAC estimate, we therefore run iteratively reweighted least squares on all frames. Each iteration sets the robust scales $\sigma_R$ and $\sigma_t$ of \cref{eq:pose_cost} to $1.4826\cdot$MAD of the residual norms, floored at $1.4826$ times their median. It combines the normalized residuals into $e_i=\big((\|\mathbf{r}^R_i\|/\sigma_R)^2+(\|\mathbf{r}^t_i\|/\sigma_t)^2\big)^{1/2}$ and sets the Cauchy weight $w_i=1/\big(1+(e_i/c)^2\big)$, $c=2.3849$. We then solve the weighted problem until the parameters stop changing (at most 20 iterations). Frames with gross errors keep a near-zero weight. The result no longer depends on the RANSAC seed, and on clean data it is as precise as a full least-squares fit.

\paragraph{Scale-collapse guard.}
If a prediction's centers disagree with the ground truth under the rotation implied by its orientations, the cost of \cref{eq:pose_cost} keeps decreasing as $s\rightarrow0$. In this degenerate optimum, every frame maps onto the ground-truth centroid. If the fitted scale falls below $1\%$ of the scale from a center-only fit (Umeyama inside the same RANSAC harness, minimal sample of three), we use the center-only fit and flag the case. This happens only for trajectories that fail outright.

\begin{algorithm}[t]
\caption{Robust pose-aware $\mathrm{Sim}(3)$ trajectory alignment}\label{alg:traj_align}
\small
\KwIn{Predicted and ground-truth world-to-camera rotations $\{\hat{R}_i\},\{R_i\}$ and camera centers $\{\hat{\mathbf{c}}_i\},\{\mathbf{c}_i\}$, $i=1,\dots,n$; $T{=}25$, $c{=}2.3849$, $K{=}20$, $\gamma{=}10^{-2}$}
\KwOut{$S=(s,R,\mathbf{t})$, inlier flags, collapse flag}
Define $E_{\mathcal{A}}(S)$ as \cref{eq:pose_cost} summed over $i\in\mathcal{A}$ with $w_i=\sigma_R=\sigma_t=1$\;
\lIf{$n<3$}{\Return{Umeyama fit on all centers~\cite{DBLP:journals/pami/Umeyama91}}}
\tcp{(1) LMedS RANSAC over minimal samples}
$\mu^\star\leftarrow\infty$\;
\For{$k=1,\dots,T$ \textnormal{(fixed seed)}}{
  draw $\mathcal{A}$ with $|\mathcal{A}|=2$;\quad $S_k\leftarrow\mathrm{LM}\big(\arg\min_S E_{\mathcal{A}}(S)\big)$ from a closed-form initialization\;
  $\mu\leftarrow\mathrm{median}_i\|\mathbf{r}^t_i(S_k)\|^2$\;
  \lIf{$\mu<\mu^\star$}{$(S,\mu^\star)\leftarrow(S_k,\mu)$}
}
$\hat{\sigma}\leftarrow1.4826\,(1+\tfrac{5}{n-2})\sqrt{\mu^\star}$;\quad $\mathcal{I}\leftarrow\{i:\|\mathbf{r}^t_i(S)\|\le2.5\,\hat{\sigma}\}$\;
\lIf{$|\mathcal{I}|\ge2$}{$S\leftarrow\mathrm{LM}\big(\arg\min_S E_{\mathcal{I}}(S)\big)$ from $S$}
\tcp{(2) Cauchy IRLS refit on all frames}
\For{$k=1,\dots,K$}{
  $\sigma_R\leftarrow\varsigma(\{\|\mathbf{r}^R_i\|\})$;\quad $\sigma_t\leftarrow\varsigma(\{\|\mathbf{r}^t_i\|\})$, with $\varsigma(u)=\max\big(1.4826\,\mathrm{MAD}(u),\ 1.4826\,\mathrm{med}(u)\big)$\;
  $e_i\leftarrow\big((\|\mathbf{r}^R_i\|/\sigma_R)^2+(\|\mathbf{r}^t_i\|/\sigma_t)^2\big)^{1/2}$;\quad $w_i\leftarrow1/\big(1+(e_i/c)^2\big)$\;
  $S'\leftarrow\mathrm{LM}\big(\arg\min_S\sum_i w_i\big(\|\mathbf{r}^R_i\|^2/\sigma_R^2+\|\mathbf{r}^t_i\|^2/\sigma_t^2\big)\big)$ from $S$\;
  $\Delta\leftarrow\|S'-S\|$;\quad $S\leftarrow S'$\;
  \lIf{$\Delta<10^{-9}$}{\textbf{break}}
}
\tcp{(3) Scale-collapse guard}
$S_c=(s_c,R_c,\mathbf{t}_c)\leftarrow$ stage (1) with Umeyama on centers only, $|\mathcal{A}|=3$\;
\lIf{$s<\gamma\,s_c$}{\Return{$S_c$, flagged as collapsed}}
\Return{$S$, inliers $\{i: e_i\le c\}$}\;
\end{algorithm}

\subsubsection{Choice of the estimator}
\label{sec:suppl:traj_align}

On real predictions, the correct alignment is unknown, so we cannot validate an alignment estimator on benchmark outputs. We therefore simulate predictions whose true alignment is known and compare our estimator (\cref{sec:method:align}) with five alternatives:
\textbf{evo}, the Umeyama $\mathrm{Sim}(3)$ on all camera centers used by most published SLAM tables~\cite{grupp2017evo,DBLP:journals/pami/Umeyama91};
\textbf{R-DAGSfM}, LMedS RANSAC over the DAGSfM pose fit~\cite{DBLP:journals/pr/ChenSCW20} with an inlier refit, the estimator of an earlier version of the benchmark;
\textbf{GTSAM}, the pose-aware fit of \cref{eq:pose_residual} on all frames without RANSAC;
\textbf{R-GTSAM}, the same fit inside the RANSAC harness;
\textbf{GTSAM+refit}, GTSAM followed by the robust refit; and
\textbf{R-GTSAM+refit} (ours), R-GTSAM followed by the robust refit.
All RANSAC variants share minimal samples of two frames, 25 iterations, and the same seed.

\paragraph{Simulation setup.}
We generate ground-truth trajectories of 300 frames in one of four shapes: handheld, near-static, driving, and loop. We map each through a random $\mathrm{Sim}(3)$ with a scale in $[0.05,20]$ and then corrupt it in one of six ways:
\emph{noise} (Gaussian noise on the centers, as a fraction of the trajectory extent, plus $1^\circ$ orientation noise);
\emph{drift} (an odometry random walk, in which every relative motion is perturbed and re-integrated);
\emph{scale drift} (relative translations scaled from 1 to $1+\lambda$ along the sequence);
\emph{jump} (tracking loss: the last fraction of frames is offset by a rigid $20^\circ$, 0.3-extent jump);
\emph{outliers} (a fraction of frames replaced by random poses); and
\emph{rotation noise} (orientation-only noise).
A further family covers a 100\,m straight path with lateral deviation $\epsilon$ times its length, $\epsilon\in[0,0.05]$, on which the roll about the path is barely observable from centers. In total, we run 3,450 trials, with 30 trials per shape and setting.
We report four metrics:
\emph{rotation error}, the angle between the estimated and the true alignment rotation;
\emph{ATE ratio}, the ATE after the estimated alignment divided by the ATE after the true one (1 is exact, values below 1 hide error, and values above 1 inflate it);
\emph{worst case}, the maximum of these over all trials; and
\emph{seed CV}, the coefficient of variation of the ATE over 10 RANSAC seeds, measured on the first five trials of each setting.

\begin{figure}[t]
\centering
\includegraphics[width=0.6\linewidth]{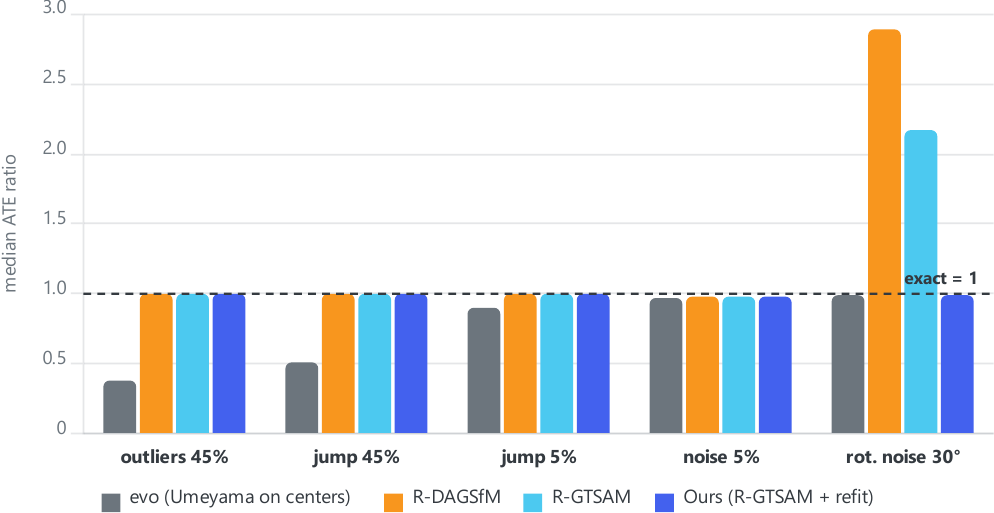}
   \caption{\textbf{Trajectory alignment matters.} Median ATE ratio, \ie, the ATE after the estimated alignment divided by the ATE after the true one, on the real TUM RGB-D and 7-Scenes ground-truth trajectories under five corruptions (\cref{tab:align_real}); 1 is exact, values below 1 hide error, and values above 1 inflate it. Under 45\% outliers or a 45\% tracking-loss jump, the Umeyama fit of evo understates the ATE by up to $2.6\times$. RANSAC (R) fixes this, but R-DAGSfM and R-GTSAM inflate the ATE by $2.9\times$ and $2.2\times$ under $30^\circ$ orientation noise. Our estimator, R-GTSAM followed by the robust refit, stays exact under outliers and jumps, is unaffected by orientation noise, and matches evo under position noise.}
\label{fig:spatialarena_traj_align}
\end{figure}

\paragraph{Results on simulated trajectories.}
From \cref{tab:align_synthetic,tab:align_worst,tab:align_collinear}, we draw four conclusions.
\emph{evo is not robust.} One tracking-loss jump or a few wrong frames drag the whole alignment, so the reported ATE understates the true error (ATE ratio 0.44--0.87). The method then looks better than it is. On exactly straight paths evo fails outright, and on nearly straight ones it returns rotations that are off by tens of degrees.
\emph{The previous estimator is robust but noisy.} R-DAGSfM averages rotations from minimal samples, so orientation noise inflates its ATE ($2.6\times$ at $30^\circ$). Under drift, where no subset of frames is exactly right, its answer depends on the RANSAC seed: the ATE varies by 10--35\% across seeds. R-GTSAM removes this bias in the median but not in the worst case ($5.3\times$ at $30^\circ$), and it keeps the seed dependence.
\emph{The robust refit removes the seed dependence} (seed CV $\le 2{\times}10^{-5}$) and is as precise as evo on clean data. At the same time it stays exact under jumps and outliers and correct on (near-)collinear paths.
\emph{RANSAC is still needed before the refit.} Started from the plain all-frame fit, the refit reaches the same answer everywhere except under the heaviest jumps (45\%), where it stays in the wrong basin ($8$--$11^\circ$ rotation error).
Under smooth drift, any $\mathrm{Sim}(3)$ fitted to all frames absorbs part of the error. Our estimator absorbs slightly more than evo (ATE ratio 0.35--0.40 \vs 0.34--0.36), which keeps our ATE close to evo-based published numbers on well-behaved runs. Finally, 25 RANSAC iterations suffice: with 200 iterations R-DAGSfM differs only in its seed CV, and the refit makes the iteration count irrelevant as long as one sample lands in the right basin.

\paragraph{Results on real trajectories.}
To rule out artifacts of the synthetic shapes, we apply the same corruptions to the real ground-truth trajectories of TUM RGB-D and 7-Scenes, sampled at the short and long SLAM strides (\cref{tab:align_real}). The real trajectories confirm our simulation results. Our estimator is exact under outliers and jumps, where evo understates the error by up to $2.6\times$. It matches evo under noise. It is the only RANSAC variant without orientation-noise inflation or seed dependence: its largest seed CV over all 10,800 trials is 0.004, against 0.37 for R-DAGSfM. Under smooth drift, no $\mathrm{Sim}(3)$ alignment recovers the true one: rotation errors of about $12^\circ$ are common to all estimators. Among the RANSAC variants, ours absorbs the least extra error (ATE ratio 0.45, against 0.43 for evo).

\input{tables/suppl_preprocess}

\subsection{Model inputs and commercial VLM queries}
\label{sec:suppl:models}

\subsubsection{Input preprocessing}
\label{sec:suppl:preprocess}

\paragraph{Per-model preprocessing.}
We run each method at its own native input resolution and record this resolution profile with the results. Our adapters record the exact resize, padding, and crop applied to each image and invert them on the output: they return depth and confidence maps on the source pixel grid, mark border pixels that the model never saw as invalid, and export intrinsics that describe the source image. Where a model's own loader would crop image content, our adapter pads instead, wherever the model allows it. We list the result for every model in \cref{tab:suppl:preprocess}. We resize predictions that are not on the source grid bilinearly to the ground truth before scoring. In MonoDepth, we give each image to multi-view models as a single-view input and to pairwise models as the pair (image, image).

\paragraph{SLAM inputs.}
In SLAM, all methods see identical pixels. We write every sequence once at a width of 504 pixels (36 patches of 14\,px), close to the dataset's own aspect ratio: $504{\times}378$ for 7-Scenes, TUM RGB-D, and Oxford Spires, $504{\times}252$ for VBR, $504{\times}154$ for KITTI Odometry, and $504{\times}406$ for \oursdata. We resize each frame to fit and zero-pad it by at most a few pixels, never crop it, and map the ground-truth intrinsics and the \texttt{rgbd} priors in the same way. The methods keep these frames unrescaled: models that require a square canvas pad them, models with 16-pixel patches crop a border of at most 8\,px, and for methods whose code fixes another size, we set it to the input size (\cref{tab:suppl:preprocess}).

\subsubsection{Querying commercial VLMs}
\label{sec:suppl:vlm}

Claude Opus 5.5~\cite{anthropic2026opus55}, GPT-6 Astra~\cite{openai2026gpt6astra}, and Gemini 4 Argon~\cite{google2026gemini4argon} return depth only as text, and every request re-sends the whole image, so asking for one pixel per request would multiply the cost by the number of queried points. We therefore ask for a whole grid of points in a single request per image, through the OpenRouter API, on MonoDepth only.

\paragraph{Query.}
We resize each image to a long side of 1,024 pixels and send it twice as JPEG: the clean photo, and a copy with a $16{\times}16$ grid of numbered red dots. The dots are spaced uniformly, with a margin of half a cell from the image border, and each carries its number, from 1 to 256 in row-major order, on a white tag. We ask the following user message (no system prompt):
\begin{prompttext}
The first image is a photo. The second image is the same photo with 256 red dots, each labelled with a number from 1 to 256; use the first image to see the scene the markers cover. For every dot, estimate the metric depth of the scene point under the dot centre: its distance from the camera along the camera's viewing (optical) axis, in meters. If a dot lies on the sky or on anything effectively infinitely far away, answer "sky" instead of a number. Give a numeric estimate for every other dot even when unsure.\par
Return the final answer as a single JSON object mapping each dot number (as a string) to its depth in meters (or "sky"), wrapped in <answer> </answer> tags, e.g. <answer>\{"1": 2.4, "2": "sky", "3": 3.1, ...\}</answer>.
\end{prompttext}
We query all three models with temperature 0, reasoning effort \emph{low} (the reasoning text is not returned), and at most 8,192 output tokens, and only through providers that do not store or train on the prompts.

\paragraph{Incomplete answers.}
Some answers stop after the first rows of dots. In this case, we send up to two follow-up messages in the same conversation that ask for exactly the missing dots, without re-sending the images:
\begin{prompttext}
Your answer did not include dots <missing ranges, e.g. 97-256>. Give the metric depth (or "sky") for exactly these dots, estimated the same way and consistent with your previous estimates. Return one JSON object mapping each of these dot numbers (as a string) to its depth, wrapped in <answer> </answer> tags.
\end{prompttext}
The follow-ups ask for the rest of the answer rather than re-sampling it, and we stop as soon as one adds no new dot.

\paragraph{From answers to a dense depth map.}
We clamp each answered depth to the depth range declared for the scene, so that, \eg, a sky point answered at thousands of meters does not spread into its neighbors. We fill grid cells answered ``sky'' or left unparsed from their answered 8-neighbors with the mean log depth, from the nearest answers inward; a single scene-median value per gap would leave flat blocks. We then upsample the $16{\times}16$ grid bilinearly to the resized image and invert the resize, so that we score the map at ground-truth resolution with the per-image alignment of \cref{sec:method:depthalign}. An image without any parseable answer receives the scene's median depth, and we log it as an imputation; we score a constant answer with the offset-only fit of \cref{sec:suppl:impl:depth}.

\paragraph{Protocol selection.}
We chose this protocol on a pilot of 14 images (two per MonoDepth dataset). Adding the clean photo and the ``sky'' answer lowered the aligned AbsRel by about 9\% compared to a single marked image with an $8{\times}8$ grid, and the $16{\times}16$ grid lowered it a little further. Neither an RGB-guided joint bilateral fill, a second request for extra dots around depth edges, nor a system prompt that frames the task for a depth expert improved the aligned scores, so we use none of them.

\paragraph{Cost.}
We record the billed cost of every request with its answer. Running the full MonoDepth suite, including the pilot and the follow-up requests, cost \$510 for Claude Opus 5.5 (11,820 requests), \$1,122 for GPT-6 Astra (11,804), and \$364 for Gemini 4 Argon (11,706), \$1,996 in total (OpenRouter prices of November 2026), \ie, about 4.3, 9.5, and 3.1 US cents per request. We cache the answers by model, prompt, settings, and image, so re-scoring a model costs nothing.

\FloatBarrier

\end{document}